\documentclass{article}

\usepackage{arxiv}

\usepackage[utf8]{inputenc} 
\usepackage[T1]{fontenc}    
\usepackage{hyperref}       
\usepackage{url}            
\usepackage{booktabs}       
\usepackage{amsfonts}       
\usepackage{nicefrac}       
\usepackage{microtype}      
\usepackage{lipsum}
\usepackage{graphicx}
\graphicspath{ {./images/} }
\usepackage{amsmath,amssymb}
\usepackage{multirow}
\usepackage[ruled,vlined]{algorithm2e} 
\usepackage{algorithmic}
\usepackage{caption}
\usepackage{subcaption}
\SetKwComment{Comment}{$\triangleright$\ }{}

\title{Tom: Leveraging trend of the observed gradients for faster convergence}

\author{
 Anirudh Maiya \\
  Department of Computer Science and Engineering\\
  PES University\\
  Bangalore, India\\
  \texttt{maiyaanirudh@gmail.com}\\
   \And
 Inumella Sricharan \\
  Department of Computer Science and Engineering\\
  PES University\\
  Bangalore, India\\
  \texttt{charan.sricharan.sri793@gmail.com} \\
  \And
 Anshuman Pandey \\
  Department of Computer Science and Engineering\\
  PES University\\
  Bangalore, India\\
  \texttt{anshumanpandey30799@gmail.com}  \\
  \And
  Srinivas K. S. \\
  Department of Computer Science and Engineering\\
  PES University\\
  Bangalore, India\\
  \texttt{srinivasks@pes.edu}
}

\begin{document}
\maketitle
\begin{abstract}
The success of deep learning can be attributed to various factors such as increase in computational power, large datasets, deep convolutional neural networks, optimizers etc. Particularly, the choice of optimizer affects the generalization, convergence rate, and training stability. Stochastic Gradient Descent (SGD) is a first order iterative optimizer that updates the gradient uniformly for all parameters. This uniform update may not be suitable across the entire training phase. A rudimentary solution for this is to employ a fine-tuned learning rate scheduler which decreases learning rate as a function of iteration. To eliminate the dependency of learning rate schedulers, adaptive gradient optimizers such as AdaGrad, AdaDelta, RMSProp, Adam employ a parameter-wise scaling term for learning rate which is a function of the gradient itself. We propose \emph{Tom} (\textbf{T}rend \textbf{o}ver \textbf{M}omentum) optimizer, which is a novel variant of Adam that takes into account of the trend which is observed for the gradients in the loss landscape traversed by the neural network. In the proposed \emph{Tom} optimizer, an additional smoothing equation is introduced to address the trend observed during the process of optimization. The smoothing parameter introduced for the trend requires no tuning and can be used with default values. Experimental results for classification datasets such as CIFAR-10, CIFAR-100 and CINIC-10 image datasets show that \emph{Tom} outperforms Adagrad, Adadelta, RMSProp and Adam in terms of both accuracy and has a faster convergence. The source code is publicly made available at \href{https://github.com/AnirudhMaiya/Tom}{this URL}
\end{abstract}

\section{Introduction}
Over the recent years, Deep learning (DL) techniques have greatly simplified the task of solving classical computer vision, machine vision and image processing problems. Deep Neural Networks, Graph Neural Networks, Recurrent Neural Networks, and Convolutional Neural Networks (ConvNets) have found their use-case in solving a wide array of problems across different domains. As shown in~\cite{multidomain}, these networks can be used to solve problems from different domains all while maintaining the same multi-domain model which works for different visual domains. LeNet~\cite{lenet} was one of the first convolutional neural networks which was successfully used to recognize handwritten zip code numbers. AlexNet~\cite{alexnet} created a breakthrough in the field of deep learning and showed that utilization of GPU's and training ConvNets can be done in unison. Due to their increasing popularity, ConvNets such as ZFNet~\cite{zfnet}, Inception Network~\cite{inception}, ResNet~\cite{resnet}, ResNeXt~\cite{resnext}, SENet~\cite{senet} have been used to win ImageNet Large Scale Visual Recognition Challenge~\cite{imagenet}. In the recent years, ConvNets such as EfficientNet~\cite{efficientnet}, MobileNets~\cite{mobilenet}, ShuffleNet~\cite{shufflenet}, MnasNet~\cite{mnasnet} have gained traction due to their increased accuracy even with a reduced number of parameters and FLOPS cost. U-Net~\cite{unet}, DeepLab~\cite{deeplab} and Kernel-Sharing Atrous Convolution (KSAC)~\cite{ksac} which use ConvNets have been used to solve the problem of semantic and instance segmentation for images. Additionaly Spatio-Temporal Fully Convolutional Network (STFCN)~\cite{stfcn} and Clockwork ConvNets~\cite{clockwork} have been used to address the problem of video segmentation. R-CNN~\cite{rcnn}, SPPNet~\cite{sppnet}, Fast R-CNN~\cite{fastrcnn}, Faster R-CNN~\cite{fasterrcnn}, Mask R-CNN~\cite{maskrcnn}, Feature Pyramid Networks~\cite{fpn}, You Only Look Once (YOLO)~\cite{yolo}, Single Shot MultiBox Detector (SSD)~\cite{ssd}, RetinaNet~\cite{retinanet} utilize ConvNets to solve the task of object detection. Furthermore, ConvNets have also found their use case in detecting brain tumours~\cite{braintumour}, reducing traffic congestion~\cite{trafficcnn}, classifying horticultural plantations for hyper-spectral images~\cite{remotesensing}, etc.

Most of the real world problems which can be seen from the perspective of a mathematical model involving the optimization of a loss function, deep learning has outperformed alternative machine learning solutions by a significant margin. Particularly Stochastic Gradient Descent (SGD) has emerged as one of the most powerful optimization techniques for training deep learning models. Its simplicity and strong theoretical guarantees has made it a default optimization technique for training both machine learning and deep learning models. Consider a minimization problem of an objective function $\mathcal{L}$,
\begin{equation}
    \label{eqn:01}
    \underset{\theta}{\mathrm{min}}\hspace{1mm} \mathcal{L}(y,\hat{y}) \qquad constraint \hspace{1mm} to \hspace{1mm} \mathcal{F}(y,X; \theta)
\end{equation}
where $X$ is the input data, $y$ is the observed variable and $\hat{y}$ is the output of the neural network that is represented by $\mathcal{F(\cdot)}$ with parameter $\theta$. The parameters of the network are iteratively estimated with the following updated rule, 
\begin{equation}
    \label{eqn:02}
    \theta_{t} = \theta_{t-1} - \alpha \cdot g_{t}
\end{equation}
where $\theta_{t}$ is the parameter of the network at iteration $t$, $\alpha$ is the step-size and $g_{t}$ is the gradient of the objective function $\mathcal{L}$ with respect to parameter $\theta_{t}$. A modification to accelerate the optimization process of vanilla SGD is to exponentially accumulate the past gradients. This method of accumulating exponential decaying average of past gradients is called Stochastic Gradient Descent with Momentum (SGD-M)~\cite{ilyaSGD}. SGD-M is particularly useful in accelerating the optimization process when the gradients are small or noisy. The update rule of SGD-M is as follows,
\begin{align}
    \label{eqn:03}
    m_{t} & = \beta \cdot m_{t-1} - \alpha \cdot g_{t} \\ 
    \theta_{t} & = \theta_{t-1} + m_{t}
\end{align}
where $\beta$ is the momentum parameter. However, both vanilla SGD and SGD-M have a fundamental problem of the gradient updating uniformly across all directions. Too large of a gradient update leads to unstable training and too small of an update leads to slow convergence along with the risk of getting the model stuck in a local minimum. A rudimentary solution to this problem is to employ a learning rate scheduler which decreases the learning rate as a function of the number of iterations. Some of the common schedulers include  decaying the learning rate step-based, time-based etc. Although the solution works, choosing the right learning rate and learning rate scheduler is time consuming and laborious. Hence to eliminate the dependency of learning rate schedulers, adaptive gradient methods such as AdaGrad~\cite{adagrad}, AdaDelta~\cite{adadelta}, RMSProp~\cite{rmsprop}, Adam~\cite{adam} have been introduced. These adaptive optimizers scale the gradients diagonally and are scale invariant. Every parameter of the network is multiplied with a learning rate that is adaptive. Therefore no tuning of learning rate is required for adaptive optimizers and can be often used with default learning rates without learning rate schedulers.

AdaGrad~\cite{adagrad} is an adaptive optimizer which is based on the intuition that parameters with frequent updates must be scaled with a lower learning rate and vice-versa. Hence parameters which rarely receive a larger gradient should be updated with a larger learning rate and similarly parameters which rarely receive a smaller gradient should be updated with a lower learning rate. This property of dynamically adapting learning rate is achieved by keeping a running sum of current and past squared gradients for every parameter in the network. Square root of this running sum is then calculated and inverted. The inverse value is then multiplied with a global learning rate, hence making the optimizer adaptive and less sensitive to the global learning rate set during optimization process. Mathematically the update rule can be represented as,
\begin{align}
    \label{eqn:04}
    \theta_{t} & = \theta_{t-1} - \frac{\alpha}{\sqrt{G_{t} + \epsilon}}g_{t}
\end{align}
where $\epsilon$ is a smoothing parameter to avoid division by zero and $G_{t}$ is the running sum of squared gradients up to current iteration $t$ and is given by,
\begin{equation}
    \label{eqn:05}
    G_{t} = \sum_{\tau=1}^{t} g_{\tau} \odot g_{\tau} = \sum_{\tau=1}^{t} g_{\tau}^{2}
\end{equation}

AdaGrad suffers from a fundamental problem where the accumulated sum of current and past squared gradients grows in magnitude during training. This leads to a learning rate that diminishes over time ultimately resulting in no update of parameters. RMSProp~\cite{rmsprop} addresses the diminishing learning rate problem encountered in AdaGrad. RMSProp is an adaptive optimizer, that accumulates the past and current gradients as an exponential average of squared gradients. Hence RMSProp introduces an additional smoothing factor $\beta$ that influences the importance given to recent gradients and is set to a value of 0.9. Therefore $G_{t}$ in RMSProp can be written as,
\begin{equation}
    \label{eqn:06}
    G_{t} = \beta \cdot G_{t-1} + (1 - \beta) \cdot g_{t}^{2}
\end{equation}
Adam~\cite{adam} is one of the most popular and intuitive optimizers which has made its impact in achieving State-of-the-Art results in the field of deep learning. Adam combines the forthcoming of both Stochastic Gradient Descent with Momentum along with RMSProp and gets the best of both worlds. While SGD-M speeds up the optimization process for gradients whose directions are the same, RMSProp adapts the learning rate dynamically to be more cautious as training progresses. Hence the two optimizers can be used in unison which leads to a robust, intuitive update of parameters. The authors of Adam refer to the update made in terms of 1\textsuperscript{st} and 2\textsuperscript{nd} moment vectors. The 1\textsuperscript{st} moment and 2\textsuperscript{nd} moment refers to the mean and uncentered variance of the gradient respectively. 1\textsuperscript{st} and 2\textsuperscript{nd} moment vectors are initialized with zeros for the first iteration ($t = 0$) and are updated as follows,
\begin{align}
    \label{eqn:07}
     m_{t} & = \beta_{1} \cdot m_{t-1} + (1 - \beta_1) \cdot g_{t} \\
    v_{t} & = \beta_{2} \cdot v_{t-1} + (1 - \beta_2) \cdot g_{t}^{2}
\end{align}
where $\beta_{1}$ and $\beta_{2}$ are decay-hyperparameters for 1\textsuperscript{st} and 2\textsuperscript{nd} moment vectors respectively. The authors of Adam recommend a default value of 0.9 for $\beta_{1}$ and 0.999 for $\beta_{2}$. These default values require no tuning and can be used for most of the optimization problems. The initialization of the 1\textsuperscript{st} and 2\textsuperscript{nd} moment vectors to zero, leads to an biased estimate in the initial part of training. To eliminate the bias towards the initial value, Adam introduces bias-correction for the moment vectors. The bias-corrected moment vectors can be represented as,
\begin{align}
    \label{eqn:08}
     \hat m_{t} & = \frac{m_t}{1 - \beta^{t}_{1}} \\
     \hat v_{t} & = \frac{v_t}{1 - \beta^{t}_{2}}
\end{align}
The update rule of Adam can be written as,
\begin{align}
    \label{eqn:09}
    \theta_{t} & = \theta_{t-1} - \frac{\alpha}{\sqrt{\hat v_{t}} + \epsilon}\hat m_{t}
\end{align}

AMSGrad~\cite{amsgrad} addresses the convergence issue in Adam by providing an optimization setting where Adam converges to a sub-optimal solution. The authors show that the problem in Adam is due to the diminishing influence of the 2\textsuperscript{nd} moment vector during an update step. Hence they introduce an optimizer which inculcates a "long-term-memory" of the past gradients called AMSGrad. Unlike Adam, where the second moment vector is directly incorporated to the update rule, AMSGrad employs a maximum of past squared gradients to create a long term memory of encountered gradients along with eliminating the bias-correction for 2\textsuperscript{nd} moment vector. Hence the update rule can be defined as,
\begin{align}
    \label{eqn:10}
    v_{t} & = \beta_{2} \cdot v_{t-1} + (1 - \beta_2) \cdot g_{t}^{2} \\
     \hat v_{t} & = \max(v_{t-1}, v_{t}) \\
     \theta_{t} & = \theta_{t-1} - \frac{\alpha}{\sqrt{\hat v_{t}} + \epsilon}\hat m_{t}
\end{align}
Although AMSGrad has theoretical convergence guarantees, the results are comparable and consistent with Adam. Hence Adam is comparatively more popular in the field of deep learning due to its simplicity.

The present study proposes an optimizer called \emph{Tom} which is inspired from the principles of time series forecasting, specifically Holt's Linear Trend method (double exponential smoothing). The main contributions of this study can be summarized as follows,
\begin{itemize}
  \item We propose \emph{Tom} (Trend over Momentum), a novel variant of Adam that considers the rate of change of gradients between two successive time steps for boosting the speed of convergence.
  \item The proposed \emph{Tom} optimizer can be incorporated to any adaptive gradient based optimizer.
  \item We report the performance of the proposed \emph{Tom} optimizer against RMSProp, Adagrad and Adam for the task of classification and regression. We find a significant boost in convergence for the proposed \emph{Tom} optimizer during the initial phase of optimization and consistently outperform RMSProp, Adagrad and Adam.
\end{itemize}

\section{Proposed Trend over Momentum Optimizer}
The present section gives an overview of time series forecasting and its components. We also give a brief overview of three popular exponential smoothing forecasting methods which have found its application in predicting radio traffic for mobile networks~\cite{trafficpred}, forecasting food prices~\cite{javaisland}, predicting atmospheric pollutants~\cite{pollutant},  but also parameter averaging in training of Generative Adversarial Networks~\cite{unusualgan}. Inspired from the principles of time series forecasting, particularly exponential smoothing and its variants we introduce the proposed \emph{Tom} optimizer which employs these popular forecasting methods for estimating the weight updates of the neural network. \emph{Tom} can be incorporated into any adaptive gradient based optimizer to boost convergence with minimal additional overhead.
\subsection{Time Series Forecasting and Components}
Time series forecasting refers to the modeling of time-stamped data that are observed over regular intervals of time to predict future values. These future values are predicted based on two assumptions,
\begin{enumerate}
  \item Past values of the observed time series data is available
  \item Past values of the observed time series data is applicable for future demand
\end{enumerate}

Time series data can be decomposed into components which gives a better abstraction to understand the characteristics of the data and helps in separating the forecasts for irregular movements/noise. The components are as follows,
\begin{enumerate}
  \item \textbf{Level ($\mathbf{L_{t}}$):} Level refers to the mean value that captures the scale of the time series data.
  \item \textbf{Trend ($\mathbf{T_{t}}$):} Trend refers to the consistent increase or decrease in values of the time series data. Trend can be either linear, exponential, quadratic etc.
  \item \textbf{Seasonality ($\mathbf{S_{t}}$):} Seasonality refers to the repetitive variations that occur in the data across regular/fixed intervals of time. The interval can be hourly, daily, weekly, etc. 
  \item \textbf{Cyclical Variation ($\mathbf{C_{t}}$):} Cyclical variation refers to upward and downward movement of data which is not of fixed frequency. These movements occur due to economic changes or business cycles.
  \item \textbf{Noise ($\mathbf{e_{t}}$):} Noise refers to irregularity or randomness in the data that is unpredictable and unsystematic.
\end{enumerate}
Mathematically a time series can be modelled as,
\begin{align}
    \label{eqn:11}
    y_{t} = F(t)
\end{align}
where $y_{t}$ is the response variable which is a function of time $t$

The components of the time series data can be modelled either additively or multiplicatively.
\subsubsection{Additive Model}
Additive model can be represented as the sum of level, trend, seasonality, cyclical variation and noise. It is represented as,
\begin{align}
    \label{eqn:12}
    y_{t} = \ell_{t} + b_{t} + s_{t} + c_{t} + e_{t}
\end{align}
Additive model assumes that the seasonality and cyclical variation are independent of the trend. Hence the additive model is applicable when the seasonality is fixed over a period of time. This may not be the case since seasonality may not be constant for a time series data.

\subsubsection{Multiplicative Model}
Multiplicative model can be represented as the product of level, trend, seasonality, cyclical variation and noise. It is represented as, 
\begin{align}
    \label{eqn:13}
    y_{t} = \ell_{t} * b_{t} * s_{t} * c_{t} * e_{t}
\end{align}
Multiplicative model assumes that the seasonality is proportionate to the level observed over a period of time.

\subsection{Exponential Smoothing}
Exponential smoothing is a forecasting method in which forecasts are produced by assigning exponentially decreasing weights to past observations. Hence recent observations are associated with higher weights than the older observations. There are three types of exponential smoothing methods depending on the use cases. They are,
\subsubsection{Single Exponential Smoothing}
Single Exponential Smoothing is the simplest of exponential smoothing methods in which differential weights are assigned by exponentially decreasing weights to past observations. However single exponential smoothing is not suitable for data which has a trend or seasonality present in it, and only has the level component included in the forecast. Mathematically the forecast at time $t$ is defined as, 
\begin{align}
    \label{eqn:16}
    \ell_{t} = \alpha \cdot \ell_{t-1} + (1 - \alpha) \cdot y_{t} 
\end{align}
where 0 < $\alpha$ < 1 is the smoothing constant, $\ell_{t}$ is the estimate of level at time $t$ and $y_{t}$ is the current observation.

Single Exponential is an adaptive learning process, i.e. it adapts itself to the errors made in the previous time steps.  
\begin{align}
    \label{eqn:18}
     f_{t+1} &= \ell_{t} \notag \\
    \ell_{t} &= \alpha \cdot \ell_{t-1} + \ell_{t-1} - \ell_{t-1} + (1 - \alpha)\cdot y_{t}\notag\\
    \ell_{t} &= \ell_{t-1} - (1 - \alpha)\cdot \ell_{t-1} + (1-\alpha)\cdot y_{t}\notag\\
    \ell_{t} &= \ell_{t-1} + (1 - \alpha)\cdot (y_{t} - \ell_{t-1})\notag\\
    \ell_{t} &= \ell_{t-1} + (1 - \alpha)\cdot (y_{t} - f_{t})\notag\\
    f_{t+1} &= f_{t} + (1-\alpha)\cdot e_{t}
\end{align}

where $e_{t}$ is the error made in the forecasting at time step ${t}$.

\subsubsection{Double Exponential Smoothing}
One of the major drawbacks of single exponential smoothing is that it always lags behind the trend for the forecasts made.  Trend is a very useful property that identifies patterns in series that shows the series's movement to relatively higher or lower values over a long period of time. Hence an additional equation is introduced to address the trend associated with the time series data. This extension of single exponential smoothing is called double exponential smoothing or Holt's Linear Trend method~\cite{holtlineartrend}. Therefore the definition of level is modified in order to incorporate trend and is represented as, 

\begin{align}
    \label{eqn:20}
    \ell_{t} &= \alpha \cdot (\ell_{t-1} + b_{t-1}) + (1 - \alpha) \cdot y_{t}\notag \\
    b_{t} &= \beta \cdot b_{t-1} + (1-\beta)\cdot (\ell_{t} - \ell_{t-1})
\end{align}
where 0 < $\alpha$ < 1 is the smoothing constant for level and 0 < $\beta$ < 1 is the smoothing constant for trend. $\ell_{t}$ is the estimate of level and $b_{t}$ is the estimate of trend at time $t$.
Hence the forecast estimate $f_{t+1}$ at time $t+1$ is given by,
\begin{align}
    \label{eqn:21}
    f_{t+1} &= \ell_{t} + b_{t}
\end{align}

\subsubsection{Triple Exponential Smoothing}
Single and double exponential smoothing methods cannot handle data when there is a seasonal component associated with it. Hence in addition to the level and trend equations introduced in Holt's Linear Trend method, a third equation is introduced to capture the seasonality of the time series data. This extension is called Triple Exponential Smoothing or Holt-Winters' Seasonal method. Mathematically triple exponential smoothing with additive seasonality is represented as,
\begin{align}
    \label{eqn:22}
     \ell_{t} &= \alpha \cdot (\ell_{t-1} + b_{t-1}) + (1 - \alpha) \cdot (y_{t} - s_{t-c}) \notag \\
     b_{t} &= \beta \cdot b_{t-1} + (1-\beta) \cdot (\ell_{t} - \ell_{t-1}) \notag \\
     s_{t} &= \gamma \cdot s_{t-c} + (1 - \gamma) \cdot (y_{t} - \ell_{t-1} - b_{t-1})
\end{align}
where 0 < $\alpha$ < 1 is the smoothing constant for level, 0 < $\beta$ < 1 is the smoothing constant for trend and 0 < $\gamma$ < 1 is the smoothing constant for seasonality. $\ell_{t}$ is the estimate of level, $b_{t}$ is the estimate of trend, $s_{t}$ is the estimate at time $t$ and $c$ is the frequency of seasonality. Hence the forecast estimate $f_{t+1}$ at time $t+1$ is given by,
\begin{align}
    \label{eqn:23}
     f_{t+1} &= \ell_{t} + b_{t} + s_{t-c+1}
\end{align}

\subsection{Temporal Aspect Of Optimization In Neural Networks}
We introduce Holt's Linear Trend method to Adam and incorporate it into the existing optimization framework for faster convergence. Specifically we introduce the trend component to the existing framework of Adam to leverage the rate of change of gradients between two successive time steps for boosting convergence. The proposed \emph{Tom} optimizer is based upon the fact that the gradients computed during the optimization process have a consistent increasing, decreasing or constant trend which can be utilized to perform smarter weight updates. This property of consistency in trend of the gradients calculated between two time steps can be assigned exponentially decreasing weights. The trend component is incorporated to the 1st moment vector of Adam with an additional smoothing equation to handle the trend. The concrete algorithm for \emph{Tom} is preseneted in Algorithm~\ref{alg:tom}. The update rule for \emph{Tom} can be formulated as,
\begin{align}
    \label{eqn:24}
    \ell_{t} &= \beta_{1} \cdot (\ell_{t-1} + b_{t-1}) + (1 - \beta_{1}) \cdot g_{t}\notag \\
    b_{t} &= \beta_{2} \cdot b_{t-1} + (1 - \beta_{2}) \cdot (g_{t} - g_{t-1}) \notag \\
    v_{t} &= \beta_{3} \cdot v_{t-1} + (1 - \beta_{3}) \cdot g_{t}^{2}
\end{align}
where $\beta_1$, $\beta_2$ are decay-hyperparameters for 1\textsuperscript{st} moment vector and is set to default values of 0.9 and 0.99 respectively. $\beta_3$ which is the decay-hyperparameter for 2\textsuperscript{nd} moment vector is set to 0.999. These hyperparameters require no tuning and can be used with the mentioned default values. The final update rule \emph{Tom} without bias correction can be represented as,
\begin{align}
    \label{eqn:25}
    f_{t+1} &= \ell_{t} + b_{t} \notag \\
    \theta_{t} &= \theta_{t-1} - \frac{\alpha}{\sqrt{ v_{t}} + \epsilon} f_{t+1}
\end{align}

\begin{algorithm*}[!htbp]
\DontPrintSemicolon
\caption{Proposed \emph{Tom} optimizer with bias correction. Default values suggested for decay-hyperparameters are $\beta_1$ = 0.9, $\beta_2$ = 0.99, $\beta_3$ = 0.999, $\alpha$ = 0.001 and $\epsilon$ = 10\textsuperscript{-8}. All operations performed on vectors are element-wise.}
\label{alg:tom}
\begin{algorithmic}
\REQUIRE $\beta_1, \beta_2, \beta_3$: Smoothing constants
\REQUIRE $\alpha$: Step-size
\REQUIRE $J(\theta)$ is the objective function that has to be minimized with parameters $\theta$
\REQUIRE $\theta_0$: Initial parameters\\
\quad \quad \quad~$\ell_0 \leftarrow$ 0: Initial level vector (1\textsuperscript{st} moment vector)\\
\quad \quad \quad~$b_0 \leftarrow$ 0: Initial trend vector (1\textsuperscript{st} moment vector)\\
\quad \quad \quad~$v_0 \leftarrow$ 0: Initial 2\textsuperscript{nd} moment vector\\
\quad \quad \quad~$t \leftarrow$ 0: Initial time step\\
\hspace{-3.5mm}\KwResult{Parameters {$\theta_{t}^{*}$} after the model has converged}
\hspace{-3mm}\While{$\theta_{t}$ not converged}{
    $t \leftarrow t + 1$ \par
    $g_t \leftarrow \nabla_\theta J(\theta_{t-1})$ \Comment*[r]{Compute gradient w.r.t to objective function at timestep t} \par
    $\ell_{t} = \beta_{1} \cdot (\ell_{t-1} + b_{t-1}) + (1 - \beta_{1}) \cdot g_{t}$ \Comment*[r]{Update 1\textsuperscript{st} moment vector for level} \par
    $b_{t} = \beta_{2} \cdot b_{t-1} + (1 - \beta_{2}) \cdot (g_{t} - g_{t-1})$ \Comment*[r]{Update 1\textsuperscript{st} moment vector for trend} \par
    $v_{t} = \beta_{3} \cdot v_{t-1} + (1 - \beta_{3}) \cdot g_{t}^{2}$ \Comment*[r]{Update 2\textsuperscript{nd} moment vector} \par
    $f_{t+1} = \ell_{t} + b_{t}$ \Comment*[r]{Compute forecast estimate} \par
    $\hat f_{t+1} \leftarrow f_{t+1} \slash (1 -\ (\beta_1 \cdot \beta_2)^{t})$ \Comment*[r]{Compute bias corrected forecast estimate}\par
    $\hat v_{t} \leftarrow v_{t} \slash (1 - \beta_3^{t})$ \Comment*[r]{Compute bias corrected 2\textsuperscript{nd} moment vector} \par
    $\theta_{t} \leftarrow \theta_{t-1} - \alpha \cdot \hat f_{t+1} \slash (\sqrt{\hat v_{t}} + \epsilon) $  \Comment*[r]{Perform a step and update parameters}
}
\hspace{-3mm}\Return $\theta_{t}^{*}$ 
\end{algorithmic}
\end{algorithm*}

The reasoning behind addition of trend component to Adam is based on the fact that,
\begin{enumerate}
  \item A running history of rate of change of gradients between two successive time steps is beneficial to make informed weight updates during the optimization process.
\end{enumerate}
\section{Bias correction for Tom}
The following section presents the bias correction for \emph{Tom} optimizer. The trend component in \emph{Tom} is intertwined with the level component. Hence during the weight update process we perform the bias correction on the forecast estimate of level and trend.
The key assumption in deriving bias correction for \emph{Tom} is that the distribution of gradients at any time step is the same, i.e. 
\begin{align}
    \mathbb{E}(g_{t}) \approx \mathbb{E}(g) \ \forall~{t}
    \label{eqn:assumption}
\end{align}
where $t \in {1, 2, 3, .... }$
\subsection{Bias correction for trend}
The first moment vector for trend $b_{t}$ is a recurrence relation that unfolds into a geometric progression as follows,
\begin{align}
    b_{t} &= (1-\beta_{2})\cdot(\delta g_{t} + \beta_{2}\delta g_{t-1} + \beta_{2}^{2}\delta g_{t-2} + .... + \beta_{2}^{t-1}\delta g_{1})
\end{align}
where $\delta g_{t} = g_{t} - g_{t-1}$

Taking expectation on both sides results in,
\begin{align}
    \mathbb{E}(b_{t}) &= (1-\beta_{2})\cdot[(\mathbb{E}(g_t) - \mathbb{E}(g_{t-1})) + \beta_{2}\cdot(\mathbb{E}(g_{t-1}) - \mathbb{E}(g_{t-2})) + ..... + \beta_{2}^{t-1}\cdot(\mathbb{E}(g_1)-0)]
    \label{eqn:trend}
\end{align}
The assumption from Eqn~\ref{eqn:assumption} is applied to Eqn~\ref{eqn:trend}. Hence the expected value of trend can be written as,
\begin{align}
    \mathbb{E}(b_{t}) &= (1-\beta_{2})\cdot\beta_{2}^{t-1}\cdot\mathbb{E}(g)
\end{align}
The above relationship between expected value of trend and the expected value of gradients shows that the bias has occurred as multiplicative factor while performing exponential averaging on trend.

\subsection{Bias correction for level}
The level component of \emph{Tom} can be written as follows,
\begin{align}
    l_{t} &= \beta_{1}\cdot l_{t-1} + \beta_{1}\cdot b_{t-1} + (1 - \beta_{1})\cdot g_{t}
    \label{eqn:level}
\end{align}
Eqn~\ref{eqn:level} can be unfolded into two series A and B, and can be formulated as,
\begin{align}
    &\text{series A :} \ \beta_{1}\cdot b_{t-1} + \beta_{1}^{2}\cdot b_{t-1} + .... + \beta_{1}^{t-1}\cdot b_{1}\notag\\
    &\text{series B :} \ (1-\beta_{1})\cdot(g_{t} + \beta_{1}\cdot g_{t-1} + \beta_{1}^{2}\cdot g_{t-2} + .... +\beta_{1}^{t-1}\cdot g_{1})
\end{align}
Taking the expected value of series A results in,
\begin{align}
    \mathbb{E}(series A) &=  \beta_{1}\cdot \mathbb{E}(b_{t-1}) + \beta_{1}^{2}\cdot \mathbb{E}(b_{t-1}) + .... + \beta_{1}^{t-1}\cdot \mathbb{E}(b_{1})\notag\\
     \mathbb{E}(series A) &= \beta_{1}\cdot(1-\beta_{2})\cdot[\beta_{2}^{t-2} + \beta_{1}\cdot\beta_{2}^{t-3} + ....+ \beta_{1}^{t-3}\cdot\beta_{2} + \beta_{1}^{t-2}]\cdot \mathbb{E}(g)
    \label{eqn:pre-tele}
\end{align}
Eqn~\ref{eqn:pre-tele} can be rewritten as a telescoping series as follows,
\begin{align}
    \label{eqn:28}
     \mathbb{E}(series A) &= \beta_{1}\cdot(1-\beta_{2})\cdot[\frac{\beta_{2}^{t-1} - \beta_{1}^{t-1}}{\beta_{2} - \beta_{1}}]\cdot \mathbb{E}(g)
\end{align}
where telescoping series is defined as,
\begin{align}
    \frac{{x}^{n} - {y}^{n}}{{x} - {y}} = \sum_{k = 1}^{n} {x}^{n-k}\cdot{y}^{k-1} ~~ \forall{x, y}
\end{align}
where $x, y \in R$
 
The expected value of series B is a simple geometric progression which simplifies to the form.
\begin{align}
    \mathbb{E}(series B) = (1-\beta_{1}^{t})\cdot\mathbb{E}(g)\notag\\
\end{align}
Hence taking the sum of the expected values of both series A and B gives can be formulated as,
\begin{align}
    \label{eqn:final_bias_corr}
    \mathbb{E}(series A) + \mathbb{E}(series B) = [\beta_{1}\cdot(1-\beta_{2})\cdot(\frac{\beta_{2}^{t-1} - \beta_{1}^{t-1}}{\beta_{2} - \beta_{1}}) + (1-\beta_{1}^{t}) ]\cdot\mathbb{E}(g)
\end{align}
For smoothing constants $\beta_{1}$ = 0.9 and $\beta_{2}$ = 0.99, Eqn~\ref{eqn:final_bias_corr} can be approximated to, 
\begin{align}
    \label{eqn:31}
    \mathbb{E}(series A + series B) &\approx [1 - (\beta_{1}\cdot\beta_{2})^{t}]\cdot \mathbb{E}(g)
\end{align}

\newpage
\section{Experimental Setup}
\subsection{Classification Settings}
\label{ref:classification_settings}
The following section presents the dataset, architecture of the deep ConvNets and the training regime used for the task of image classification. 

\subsubsection{CIFAR-10 and CIFAR-100}
\label{ref:cifar-dataset}
CIFAR-10 and CIFAR-100 developed by~\cite{cifaralex} is a standard collection of images which is widely used to train machine learning models and benchmark results against various State-of-the-Art machine learning and computer vision algorithms. CIFAR-10 and CIFAR-100 are subsets of a larger dataset called 80 million tiny images dataset~\cite{80milimages}. CIFAR-10 consists of 60000 color images of size 32 x 32 belonging to 10 different classes which are airplanes, cars, birds, cats, deer, dogs, frogs, horses, ships, and trucks. The dataset is further split into 50000 training images and 10000 test images. Similarly, CIFAR-100 consists of 60000 color images of size 32 x 32, belonging to 100 different classes. Hence CIFAR-100 contains 600 images per class, in which 500 are training samples and 100 are testing samples.   

\subsubsection{CINIC-10}
\label{ref:cinic-dataset}
CINIC-10 developed by~\cite{cinicluke} was created to address the shortcomings of CIFAR-10 being too small, easy and ImageNet being too large, computationally expensive for deep learning models to train on. Hence CINIC-10 bridges the bench-marking gap between CIFAR-10 and ImageNet. CINIC-10 consists of all 60000 samples from CIFAR-10 and a subset of 210000 samples from ImageNet which are downsampled to size 32 x 32. Therefore the dataset consists of 270000 samples in total belonging to 10 different classes that are equally split into train, validation and test sets. 

\subsubsection{Architecture}
Residual Neural Network (ResNet) introduced by~\cite{resnet} is considered for the task of image classification. ResNets solve the degradation problem of deep ConvNets by introducing residual units. These residual units consists of skip connections between two non-adjacent layers. These skip connections are governed by a weight matrix in case of~\cite{highwaynetworks} and an identity matrix in case of ResNets. Additionally these skip connections also contain non-linearity layers such as ReLU~\cite{relu} and normalization layers such as batch normalization~\cite{batchnorm}. We consider 18 and 34 layered ResNet called ResNet-18 and ResNet-34 which is publicly available through Github\footnote[1]{https://github.com/kuangliu/pytorch-cifar}.

\subsubsection{Training}
In the present study, the training of deep ConvNets (ResNet-18 and ResNet-34) is performed with three different batch sizes (512, 1024 and 4096). We initialize the deep ConvNets with five different seeds and run the experiments once for every seed value. We use RMSProp, AdaGrad, Adam and the proposed \emph{Tom} optimizer to optimize the paramaters of the ConvNets with categorical cross-entropy as the objective function. All the optimizers are initialized with a fixed learning rate of 0.001 across the entire training process of 100 epochs. AdaGrad is initialized with an initial accumulator value of 0.1. The images from the training set are randomly cropped and horizontally flipped. Additionally both the training and test set images are normalized with the training set's mean and standard deviation. We eliminate bias correction term for the forecast estimate in the proposed \emph{Tom} optimizer as it decreases the performance of the test accuracy observed. 

\subsection{Regression Settings}
The following section presents the dataset, architecture of the dense neural network and the training regime used for the task of regression.

\subsubsection{Boston Housing}
Boston Housing dataset introduced by~\cite{boston} is a popular dataset that consists of information related to housing in the area of Boston. The dataset consists of  506 samples in total with 13 independent and 1 dependent variable. The independent variables include CRIM (per capita crime rate by town), ZN (proportion of residential land), INDUS (proportion of non-retail business acres per town) etc, while the dependent/target variable is MEDV (median value of the house). We split the dataset with an 80-20 split ratio and normalize the train and test set with the training set's mean and standard deviation. 

\subsubsection{California Housing}
California Housing dataset introduced by~\cite{cali} consists of 20640 samples comprising 8 independent variables and 1 dependent variable. The independent variables include MedInc (median income in the block), HouseAge (median house age in the block) etc. while the dependent/target variable is the median house value. Similar to Boston Housing dataset, we split the dataset with an 80-20 split ratio and normalize the train and test set with the training set's mean and standard deviation.

\subsubsection{Diabetes Dataset}
Diabetes dataset introduced by~\cite{diabetes} consists of 10 baseline independent variables and 1 dependent variable. The independent variables include age, sex, body mass index etc. while the dependent/target variable is a quantitative measure of the progression of disease 1 year after baseline. The dataset consists of 442 samples which are split with an 80-20 train to test split ratio and normalized with the training set's mean and standard deviation.  

\section{Architecture}
A dense neural network consisting of 2 hidden layers with 8 and 10 units each is used for the task of regression. ReLU non-linearity is used as the activation function. The output layer consists of 1 unit with linear activation.

\section{Training}
In the present study, the dense neural network is trained with batch gradient descent. We initialize the dense neural network with five different seeds and run the experiments once for every seed value. The configuration of the optimizers are identical to that of classification settings. We use mean squared error (MSE) as the objective function and train the dense neural network for 200 epochs. 

\section{Experimental Results}
\subsection{Classification Settings}
\begin{figure}[!htbp]
     \centering
     \begin{subfigure}[b]{0.3\textwidth}
         \centering
         \includegraphics[width=\textwidth]{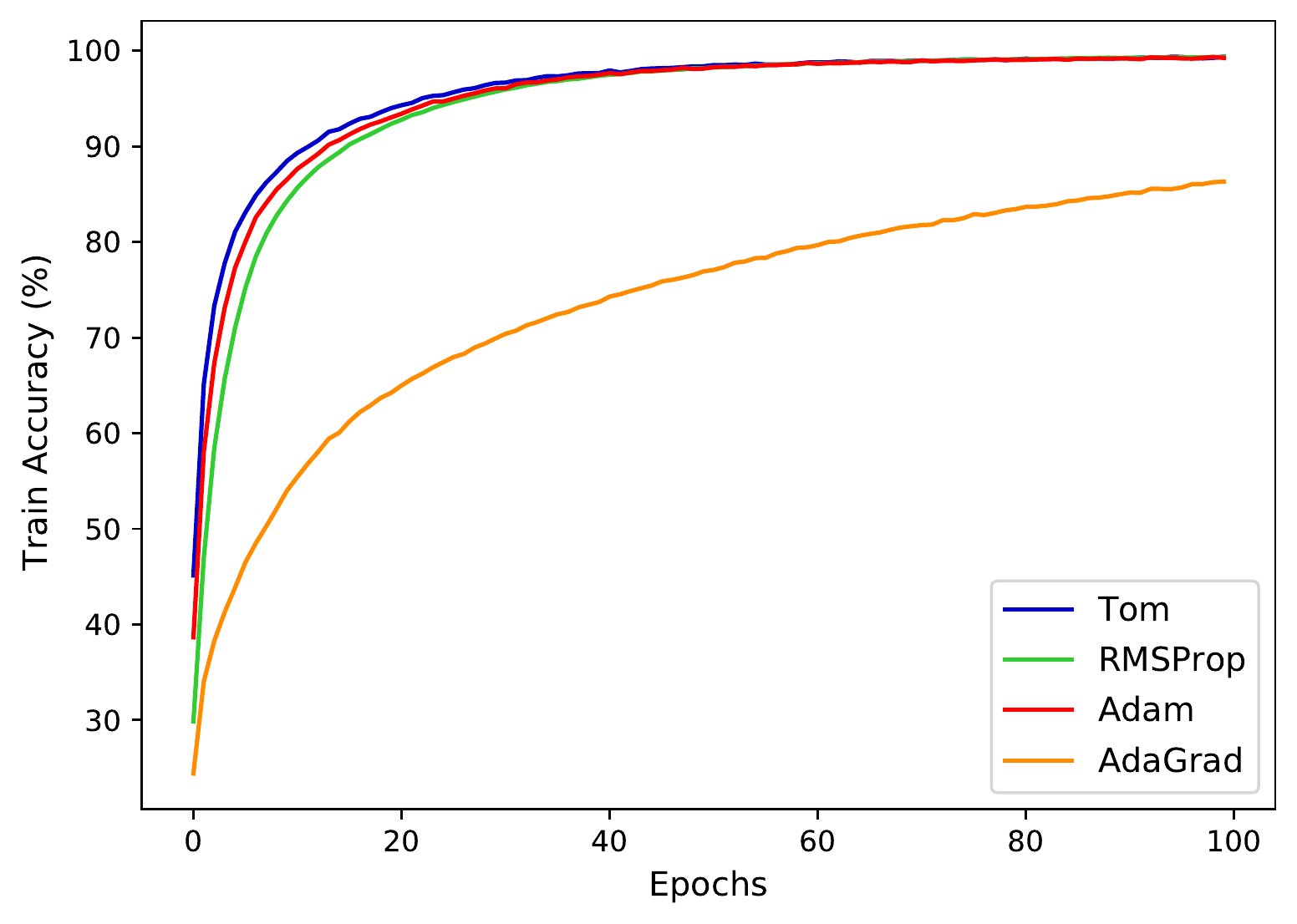}
         \caption{Training accuracy for ResNet-18 with batch size 512 on CIFAR-10}
     \end{subfigure}
     \hfill
     \begin{subfigure}[b]{0.3\textwidth}
         \centering
         \includegraphics[width=\textwidth]{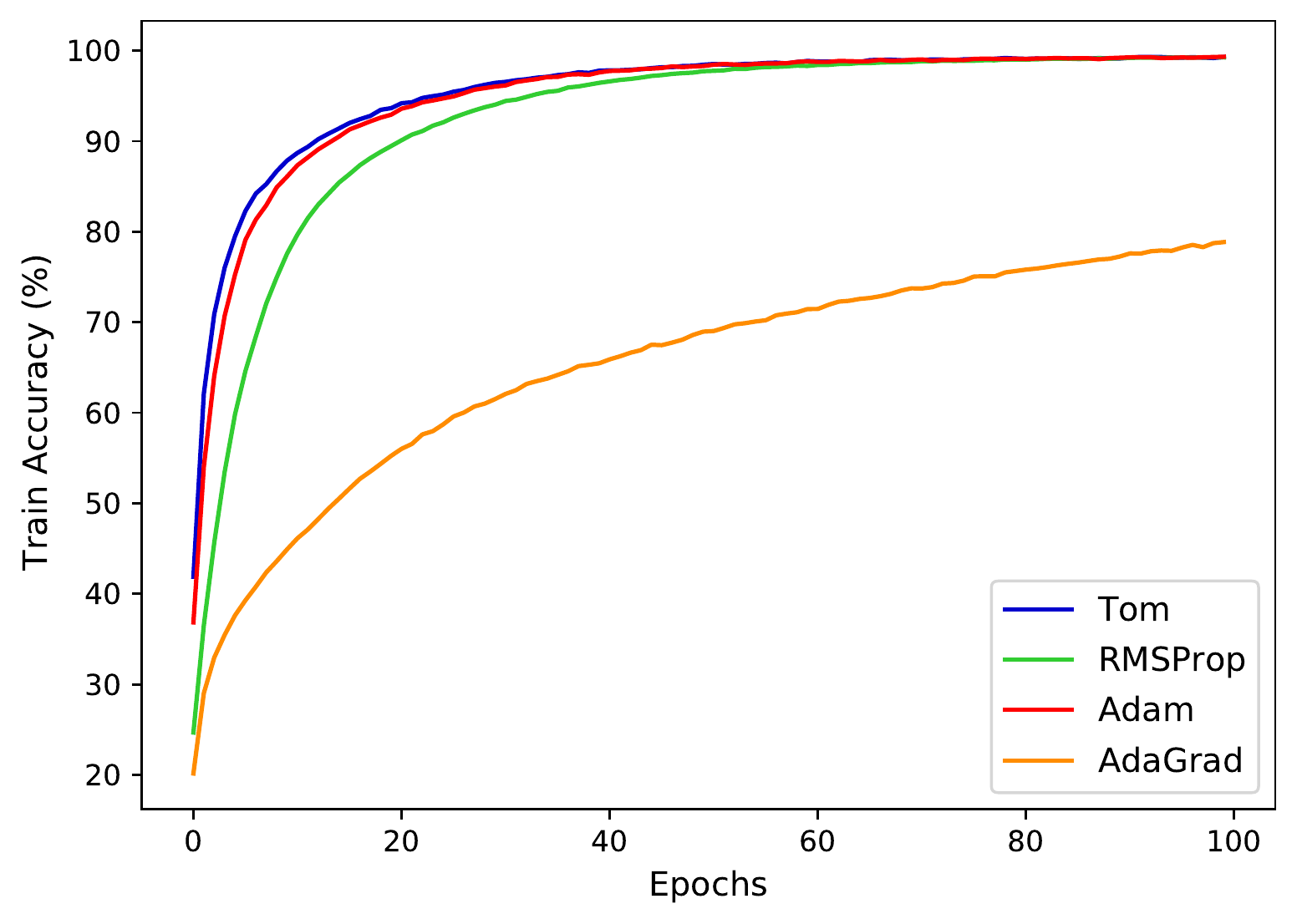}
         \caption{Training accuracy for ResNet-18 with batch size 1024 on CIFAR-10}
     \end{subfigure}
     \hfill
     \begin{subfigure}[b]{0.3\textwidth}
         \centering
         \includegraphics[width=\textwidth]{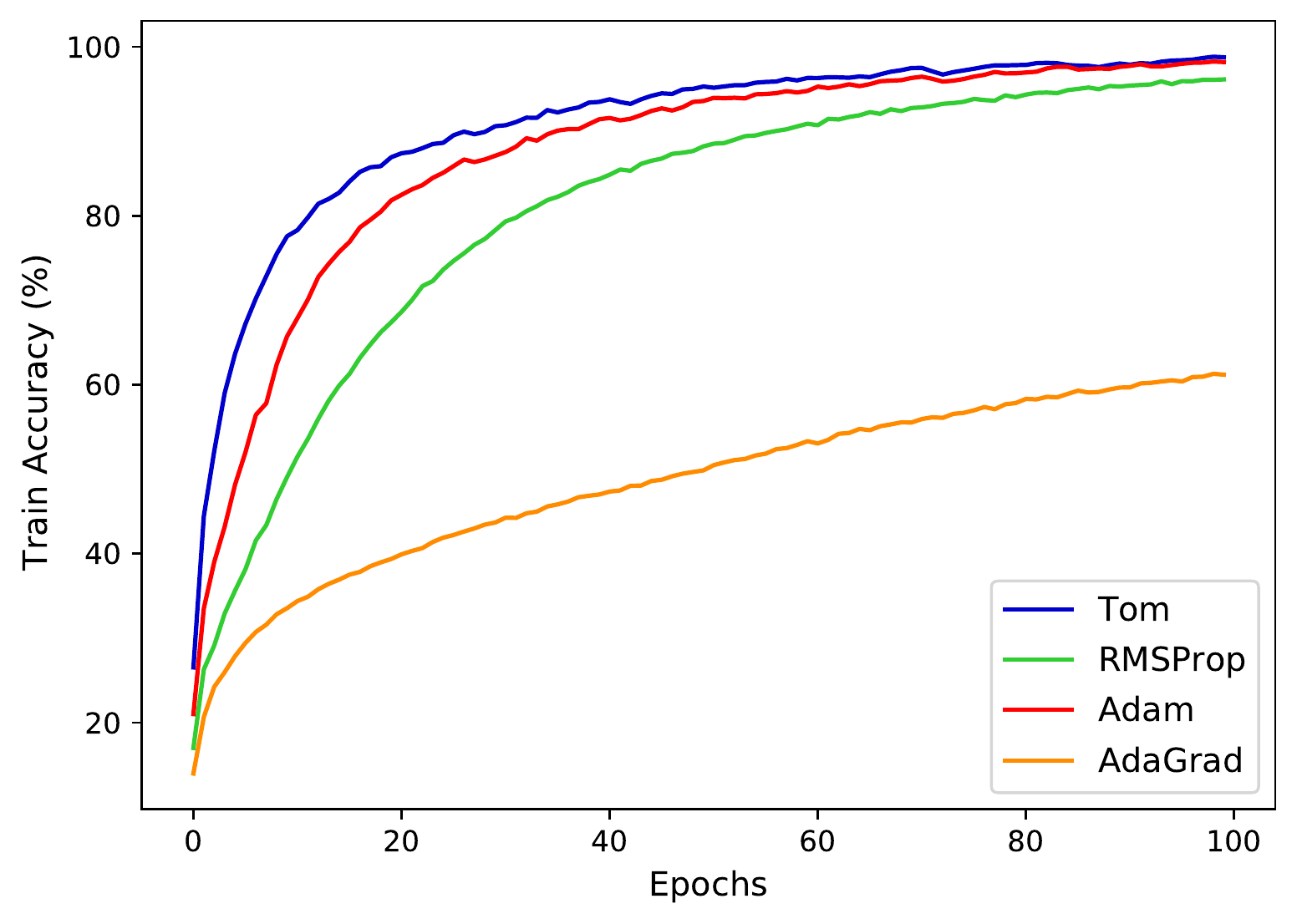}
         \caption{Training accuracy for ResNet-18 with batch size 4096 on CIFAR-10}
     \end{subfigure}
     \vfill
     \vspace{5mm}
     \begin{subfigure}[b]{0.3\textwidth}
         \centering
         \includegraphics[width=\textwidth]{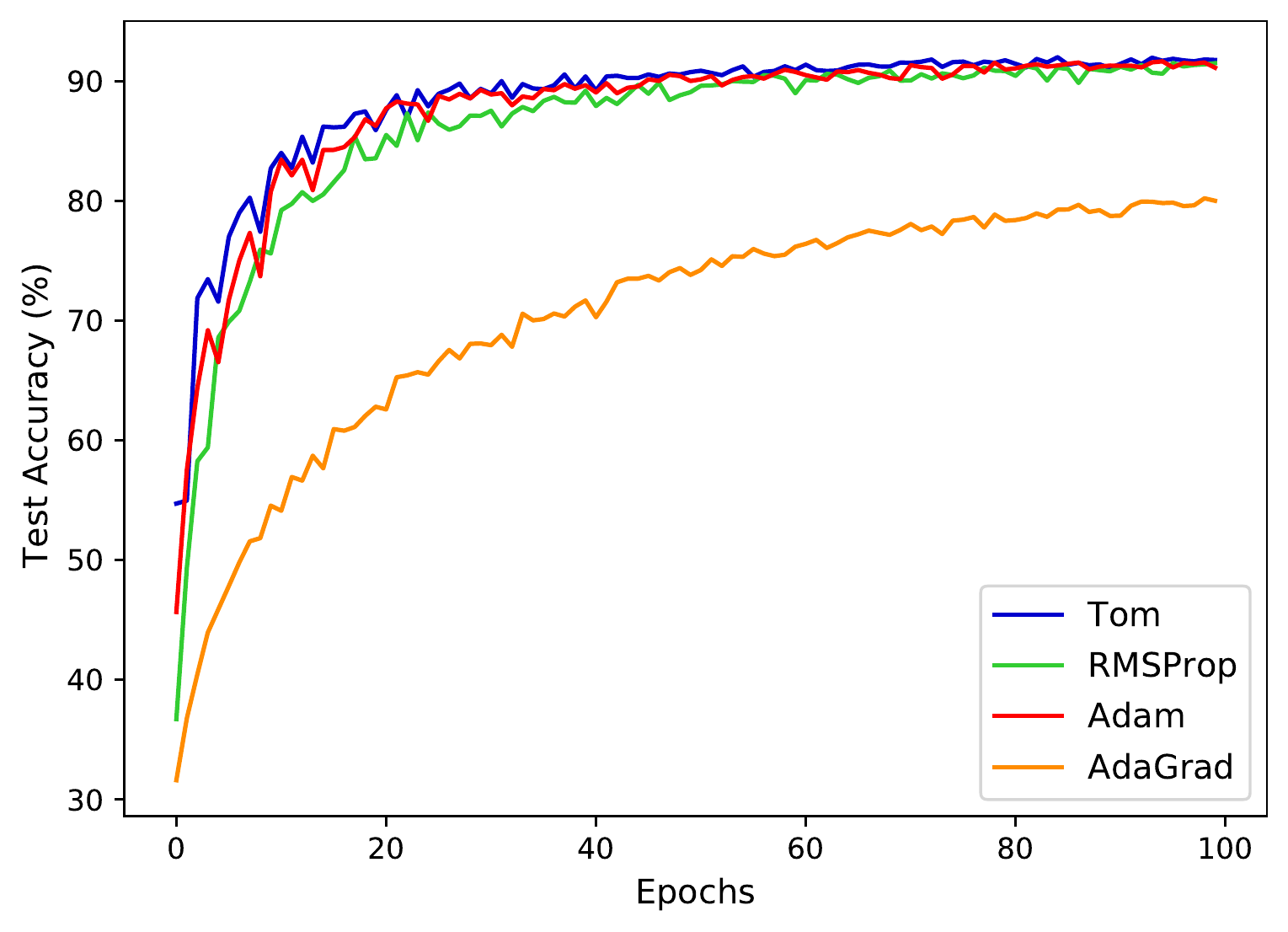}
         \caption{Test accuracy for ResNet-18 with batch size 512 on CIFAR-10}
     \end{subfigure}
     \hfill
     \begin{subfigure}[b]{0.3\textwidth}
         \centering
         \includegraphics[width=\textwidth]{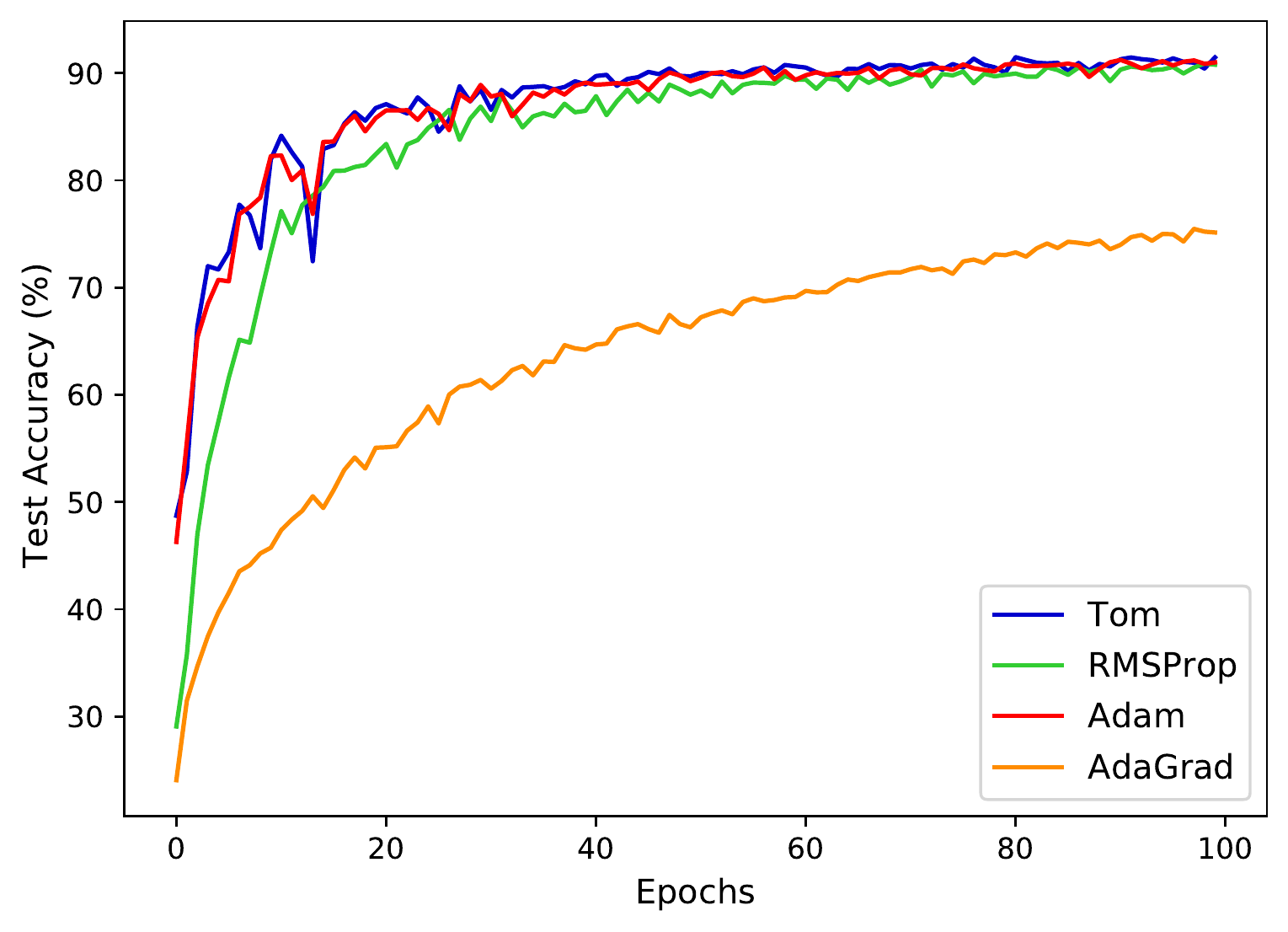}
         \caption{Test accuracy for ResNet-18 with batch size 1024 on CIFAR-10}
     \end{subfigure}
     \hfill
     \begin{subfigure}[b]{0.3\textwidth}
         \centering
         \includegraphics[width=\textwidth]{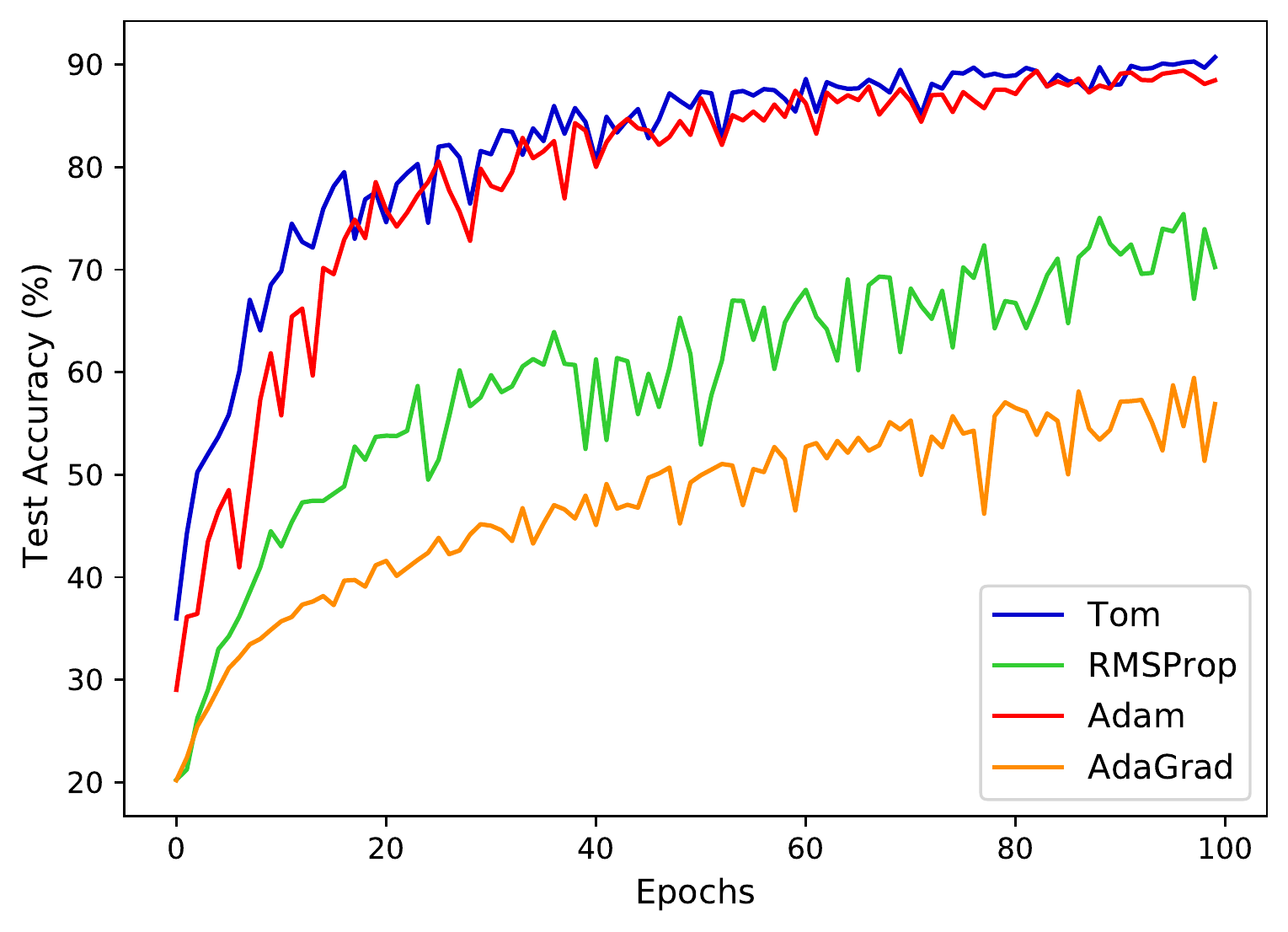}
         \caption{Test accuracy for ResNet-18 with batch size 4096 on CIFAR-10}
     \end{subfigure}
        \caption{Training and test accuracy evolution for ResNet-18 with three different batch sizes (512, 1024, 4096) on CIFAR-10}
        \label{fig:resnet18_cifar10_fig}
\end{figure}

\begin{figure}[!htbp]
     \centering
     \begin{subfigure}[b]{0.3\textwidth}
         \centering
         \includegraphics[width=\textwidth]{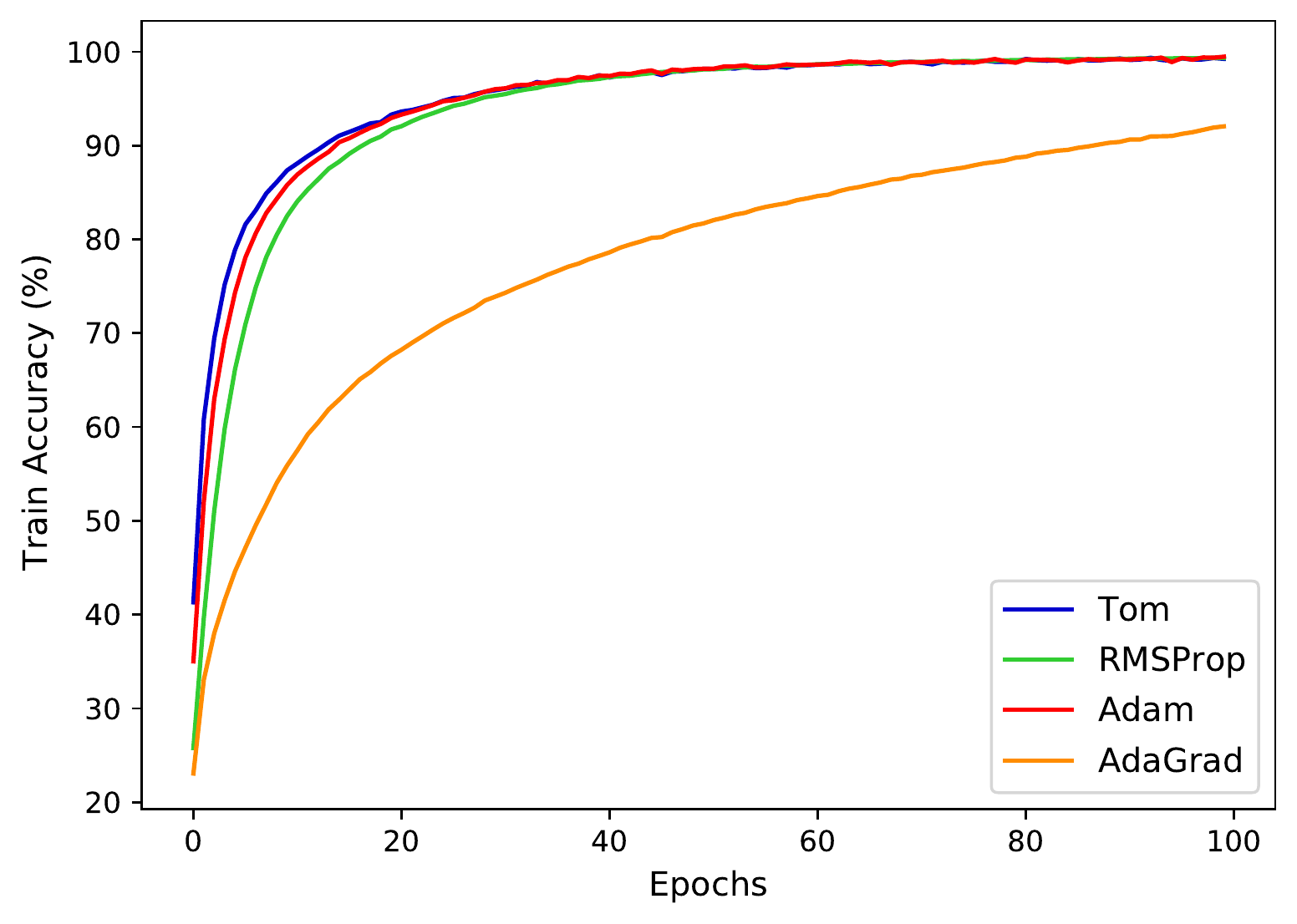}
         \caption{Training accuracy for ResNet-34 with batch size 512 on CIFAR-10}
     \end{subfigure}
     \hfill
     \begin{subfigure}[b]{0.3\textwidth}
         \centering
         \includegraphics[width=\textwidth]{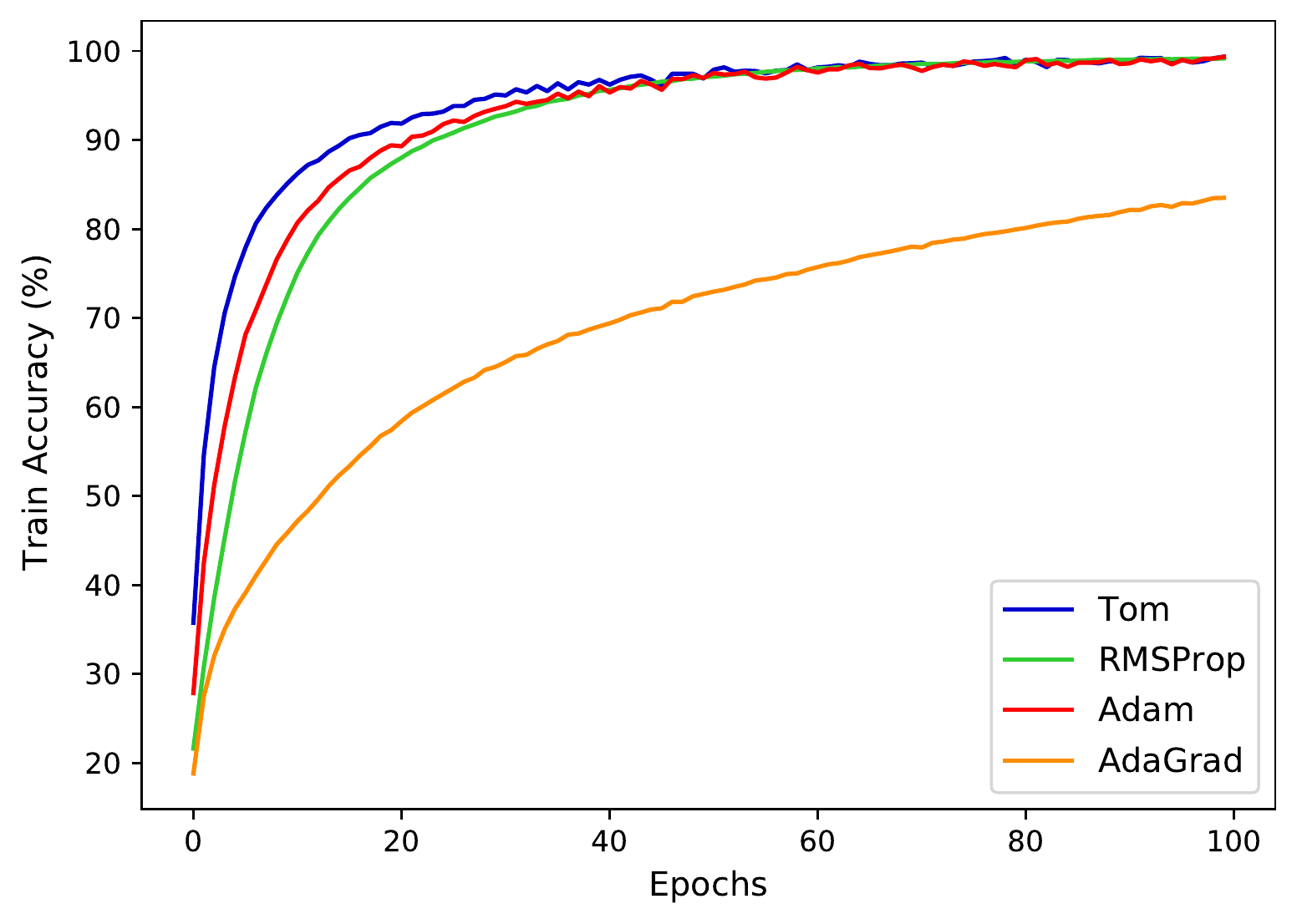}
         \caption{Training accuracy for ResNet-34 with batch size 1024 on CIFAR-10}
     \end{subfigure}
     \hfill
     \begin{subfigure}[b]{0.3\textwidth}
         \centering
         \includegraphics[width=\textwidth]{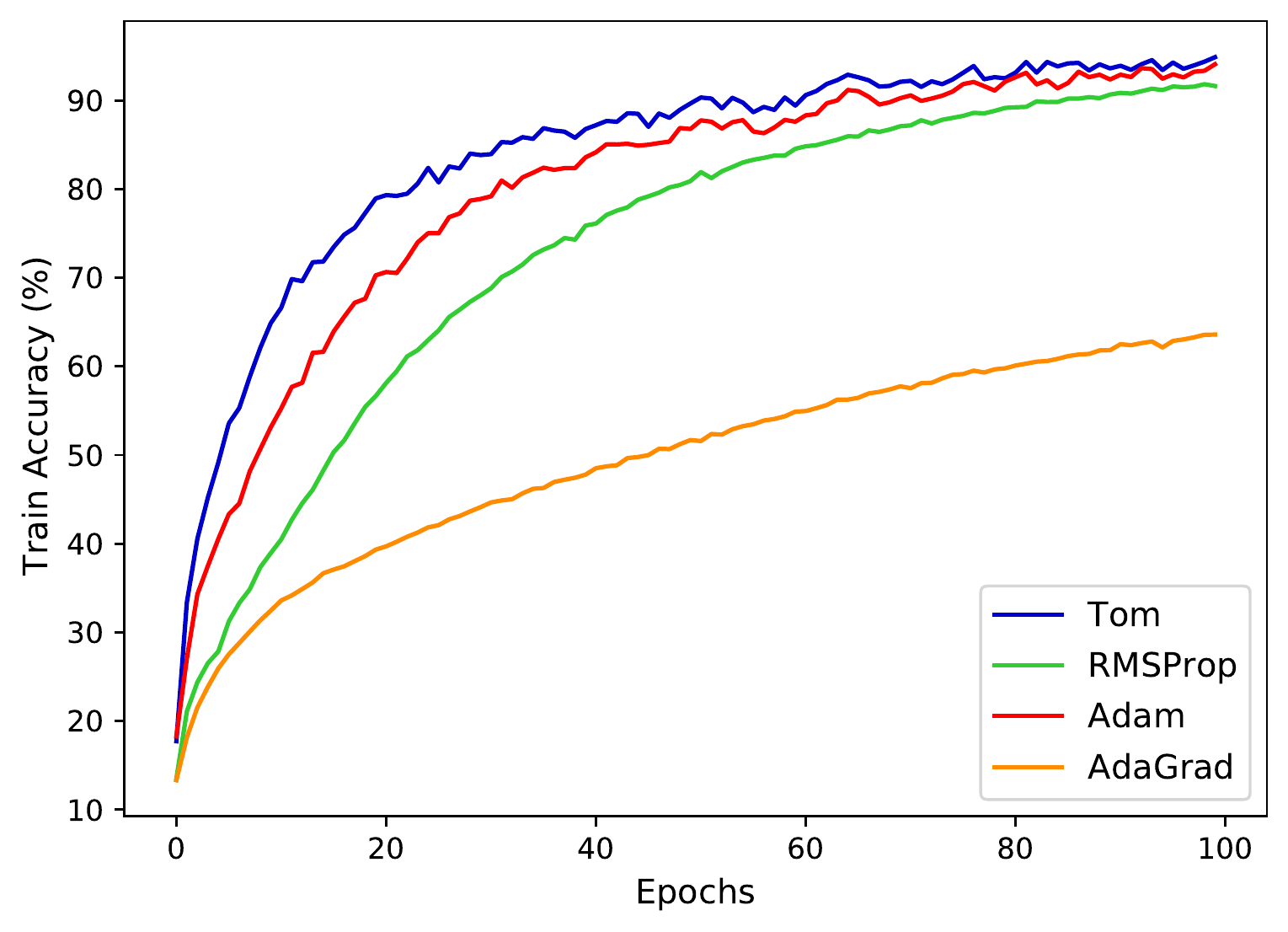}
         \caption{Training accuracy for ResNet-34 with batch size 4096 on CIFAR-10}
     \end{subfigure}
     \vfill
     \vspace{5mm}
     \begin{subfigure}[b]{0.3\textwidth}
         \centering
         \includegraphics[width=\textwidth]{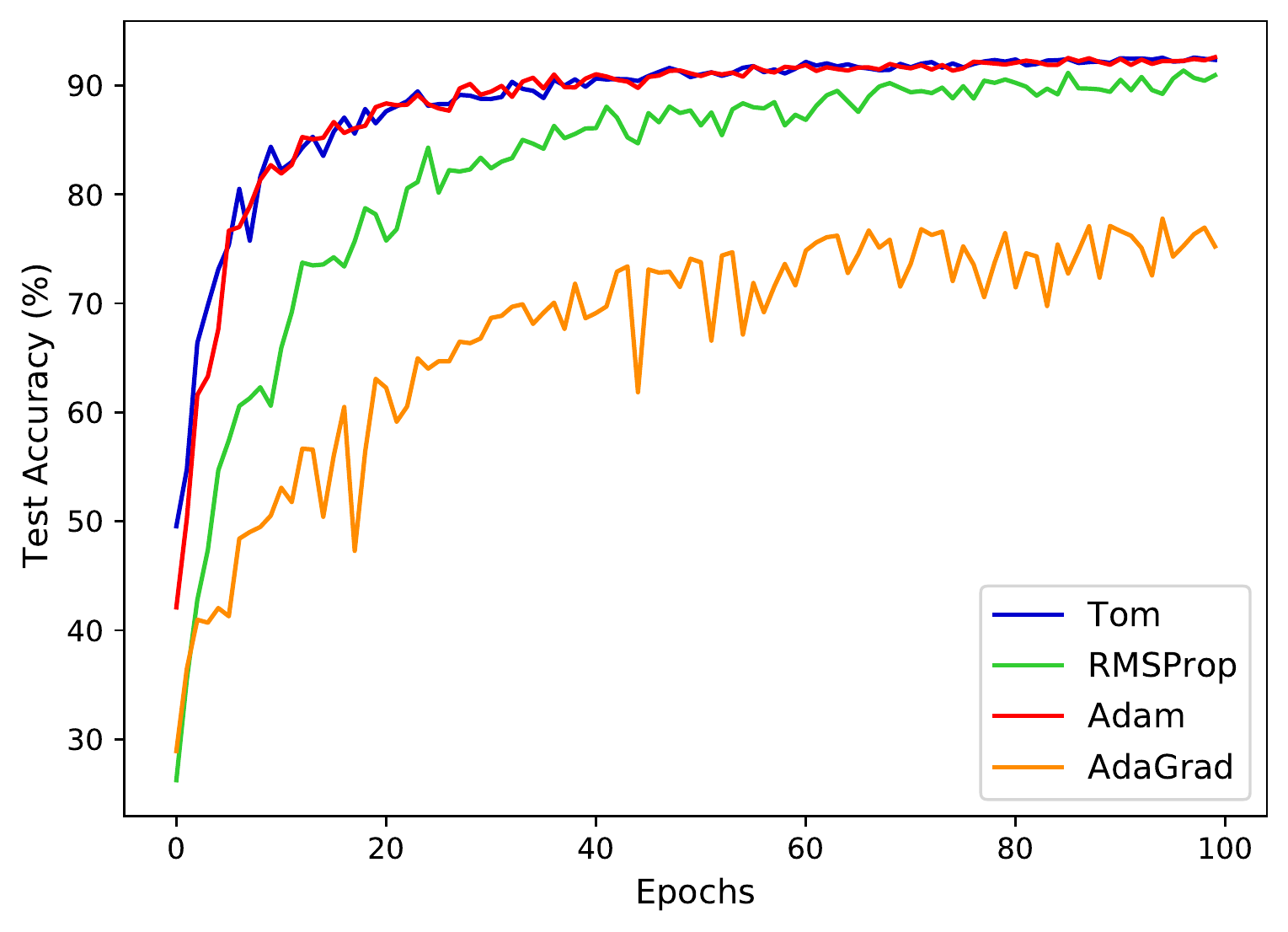}
         \caption{Test accuracy for ResNet-34 with batch size 512 on CIFAR-10}
     \end{subfigure}
     \hfill
     \begin{subfigure}[b]{0.3\textwidth}
         \centering
         \includegraphics[width=\textwidth]{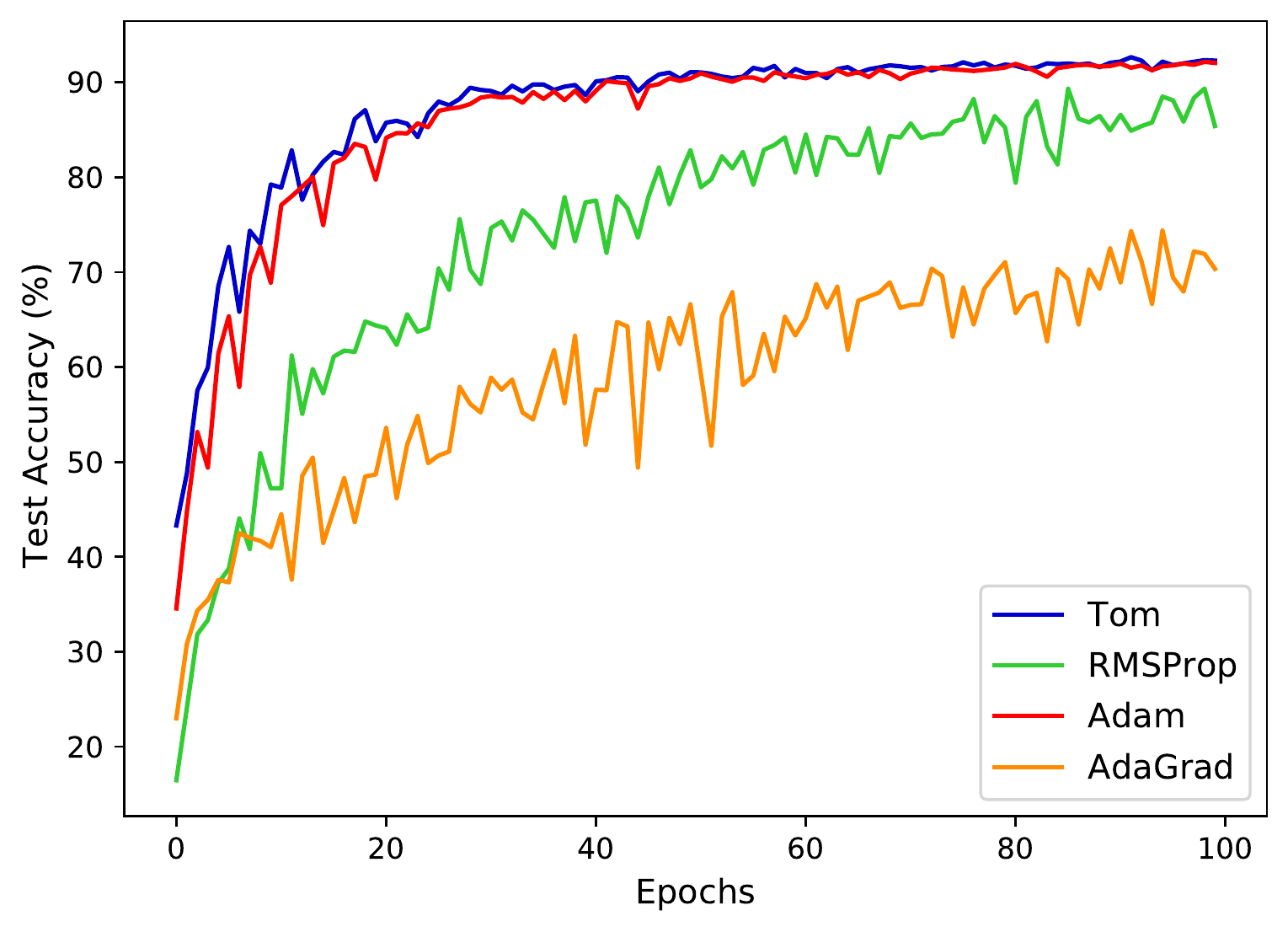}
         \caption{Test accuracy for ResNet-34 with batch size 1024 on CIFAR-10}
     \end{subfigure}
     \hfill
     \begin{subfigure}[b]{0.3\textwidth}
         \centering
         \includegraphics[width=\textwidth]{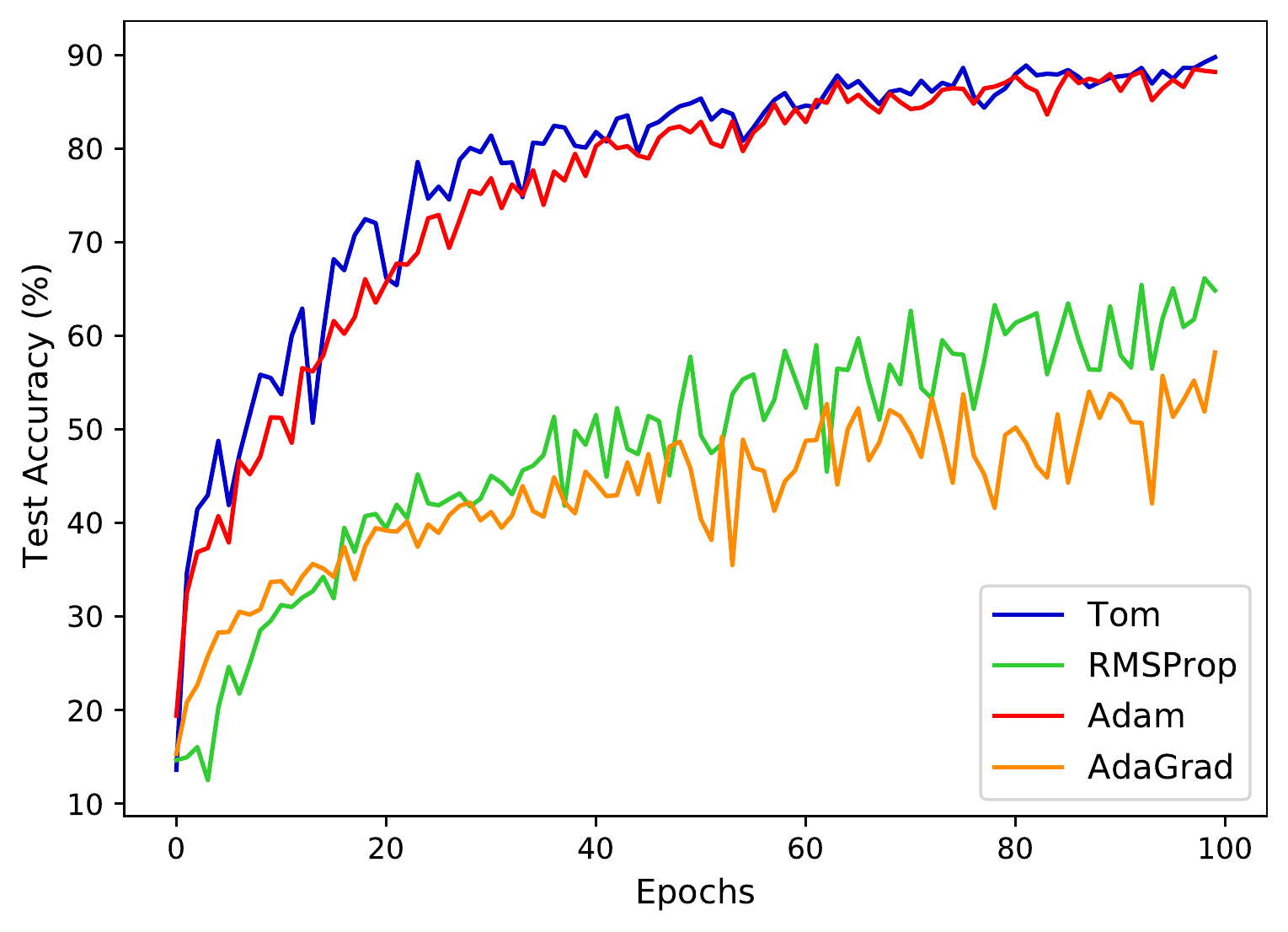}
         \caption{Test accuracy for ResNet-34 with batch size 4096 on CIFAR-10}
     \end{subfigure}
        \caption{Training and test accuracy evolution for ResNet-34 with three different batch sizes (512, 1024, 4096) on CIFAR-10}
        \label{fig:resnet34_cifar10_fig}
\end{figure}

\begin{figure}[!htbp]
     \centering
     \begin{subfigure}[b]{0.3\textwidth}
         \centering
         \includegraphics[width=\textwidth]{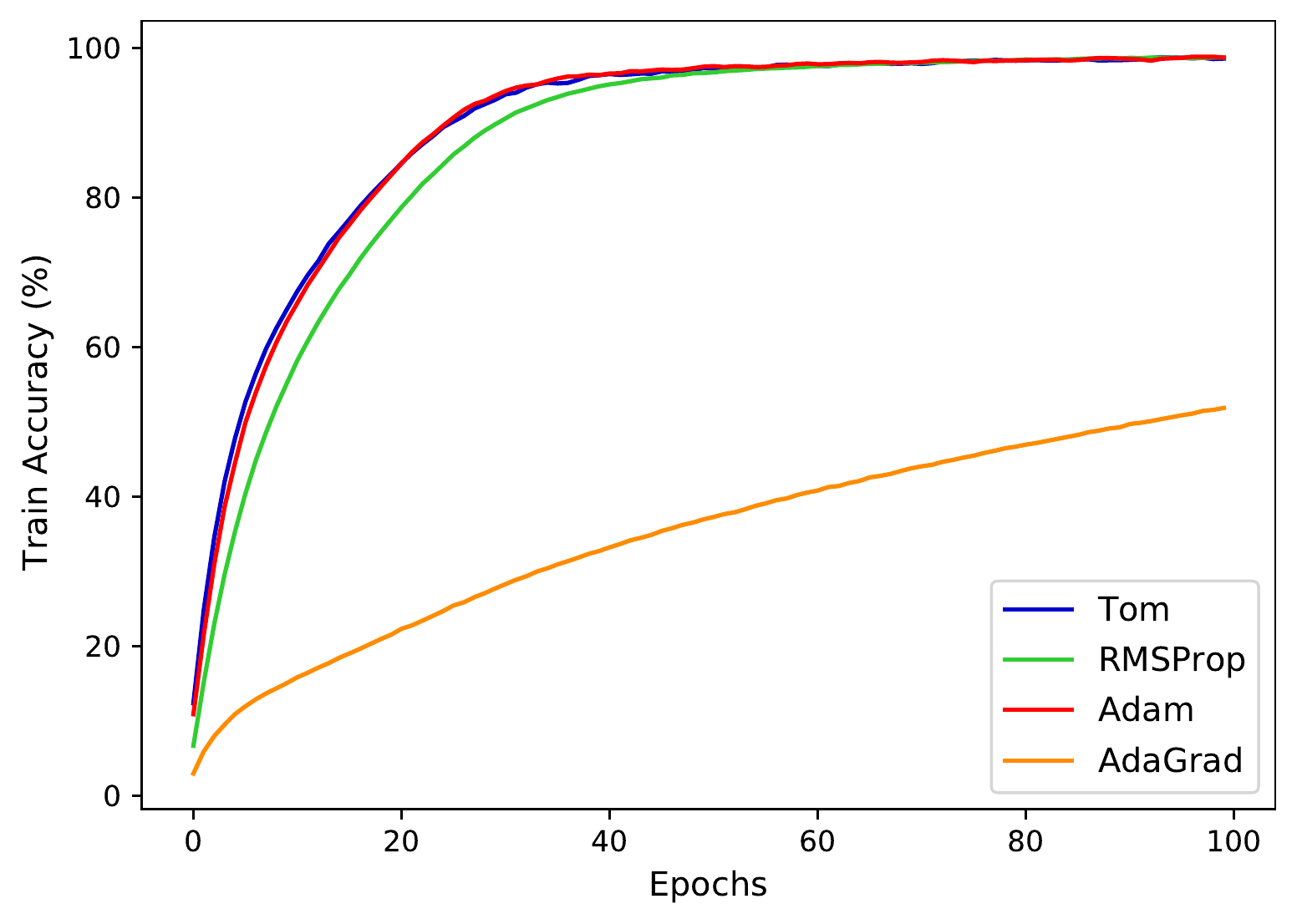}
         \caption{Training accuracy for ResNet-18 with batch size 512 on CIFAR-100}
     \end{subfigure}
     \hfill
     \begin{subfigure}[b]{0.3\textwidth}
         \centering
         \includegraphics[width=\textwidth]{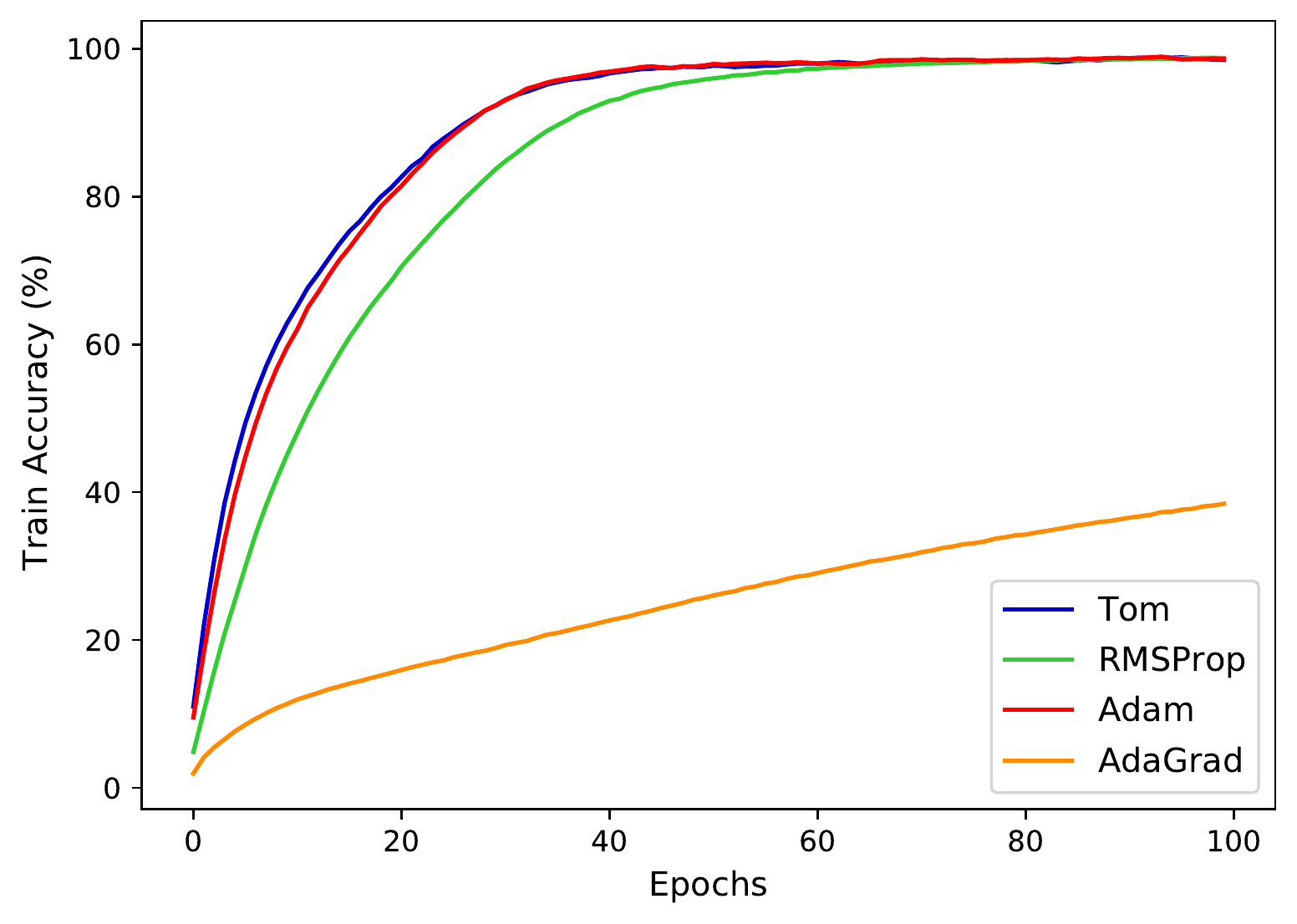}
         \caption{Training accuracy for ResNet-18 with batch size 1024 on CIFAR-100}
     \end{subfigure}
     \hfill
     \begin{subfigure}[b]{0.3\textwidth}
         \centering
         \includegraphics[width=\textwidth]{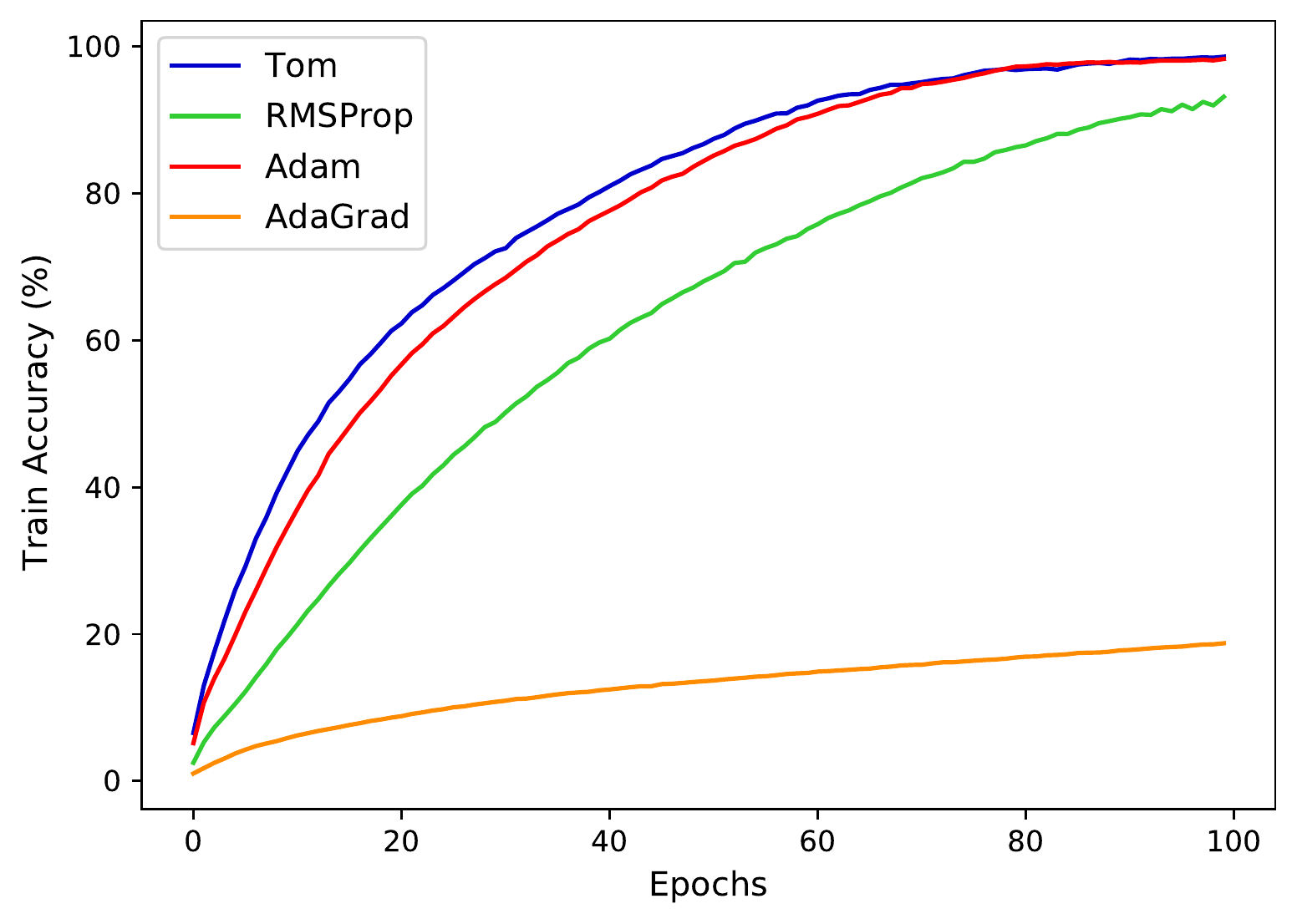}
         \caption{Training accuracy for ResNet-18 with batch size 4096 on CIFAR-100}
     \end{subfigure}
     \vfill
     \vspace{5mm}
     \begin{subfigure}[b]{0.3\textwidth}
         \centering
         \includegraphics[width=\textwidth]{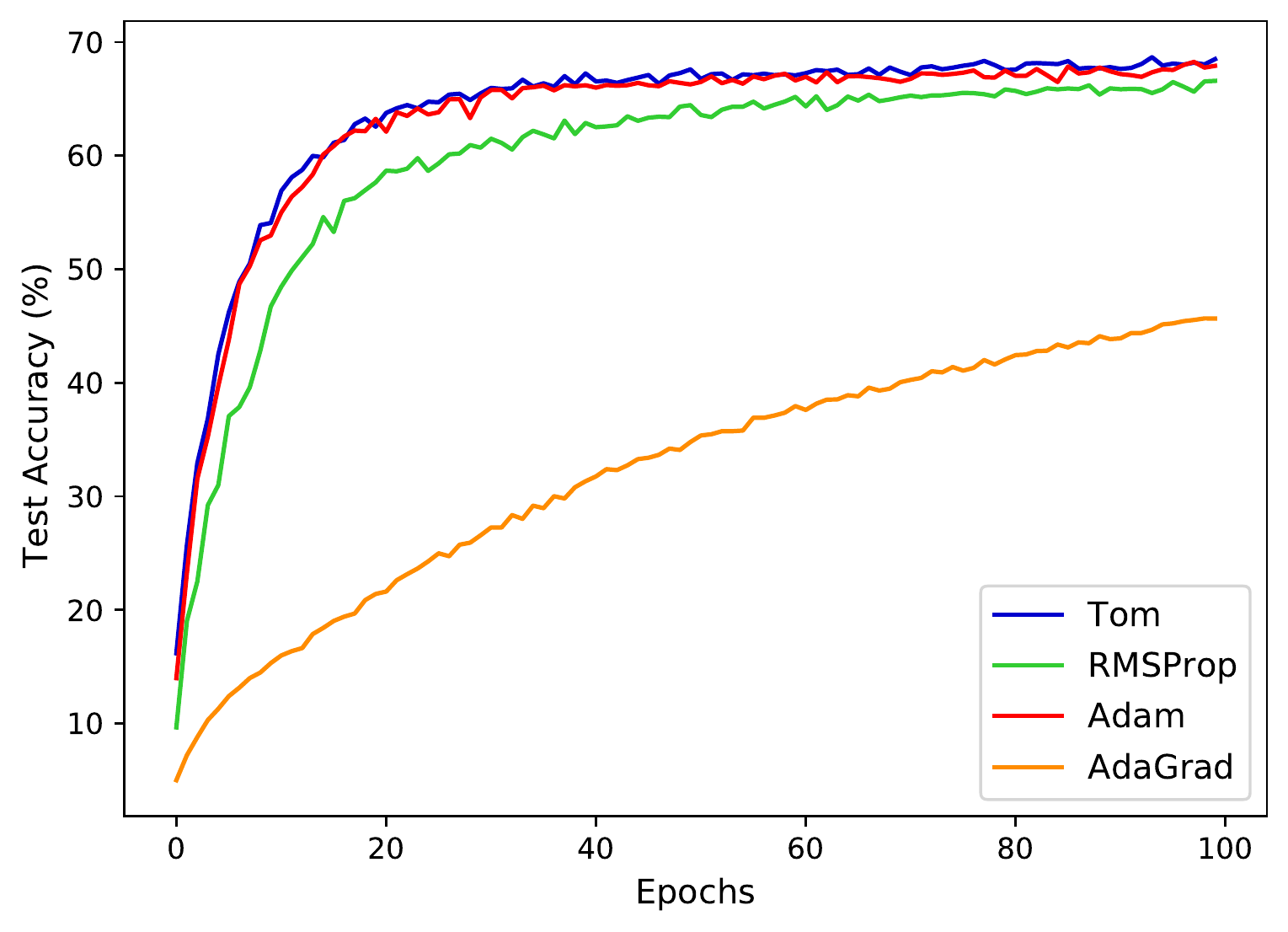}
         \caption{Test accuracy for ResNet-18 with batch size 512 on CIFAR-100}
     \end{subfigure}
     \hfill
     \begin{subfigure}[b]{0.3\textwidth}
         \centering
         \includegraphics[width=\textwidth]{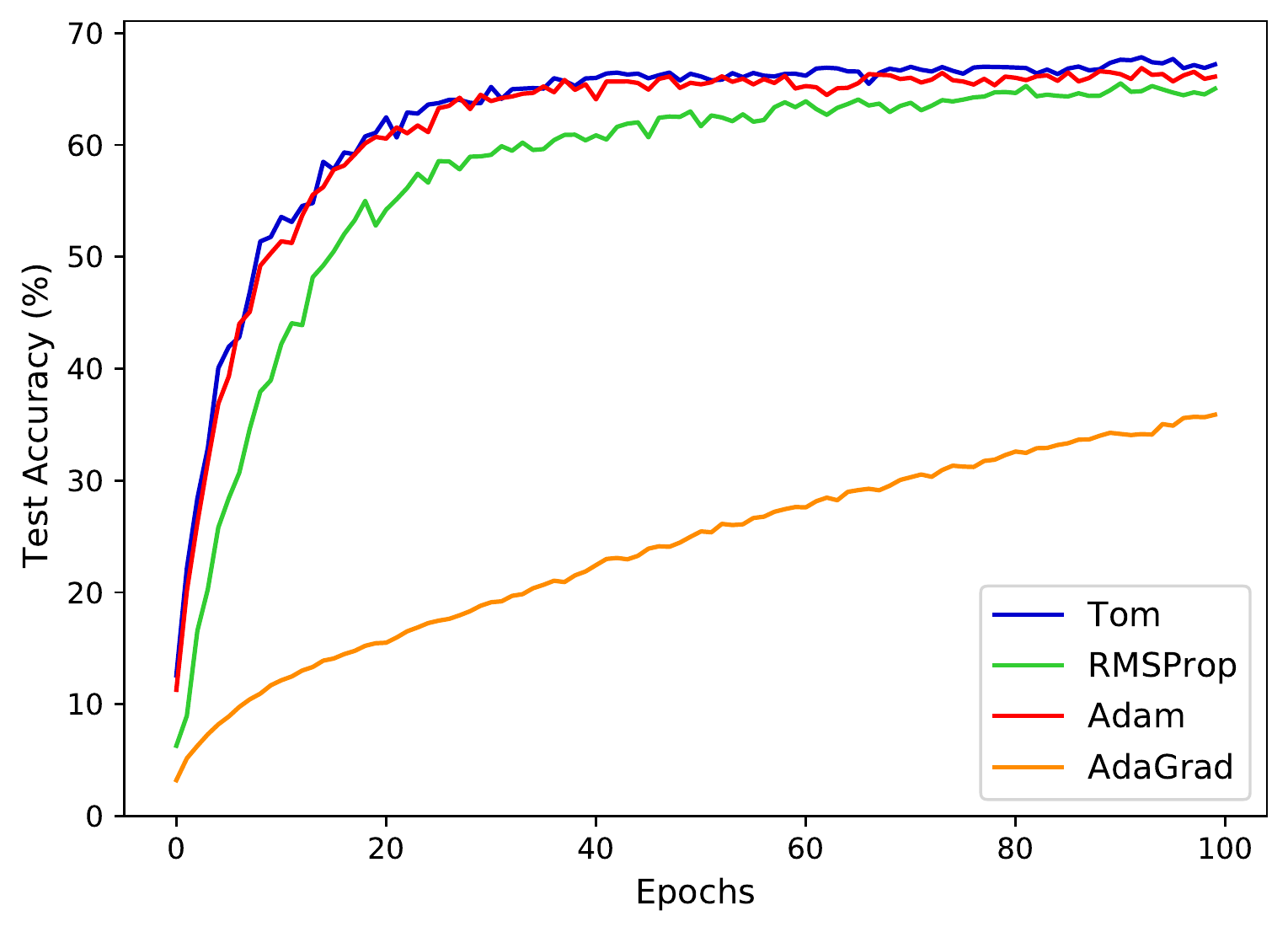}
         \caption{Test accuracy for ResNet-18 with batch size 1024 on CIFAR-100}
     \end{subfigure}
     \hfill
     \begin{subfigure}[b]{0.3\textwidth}
         \centering
         \includegraphics[width=\textwidth]{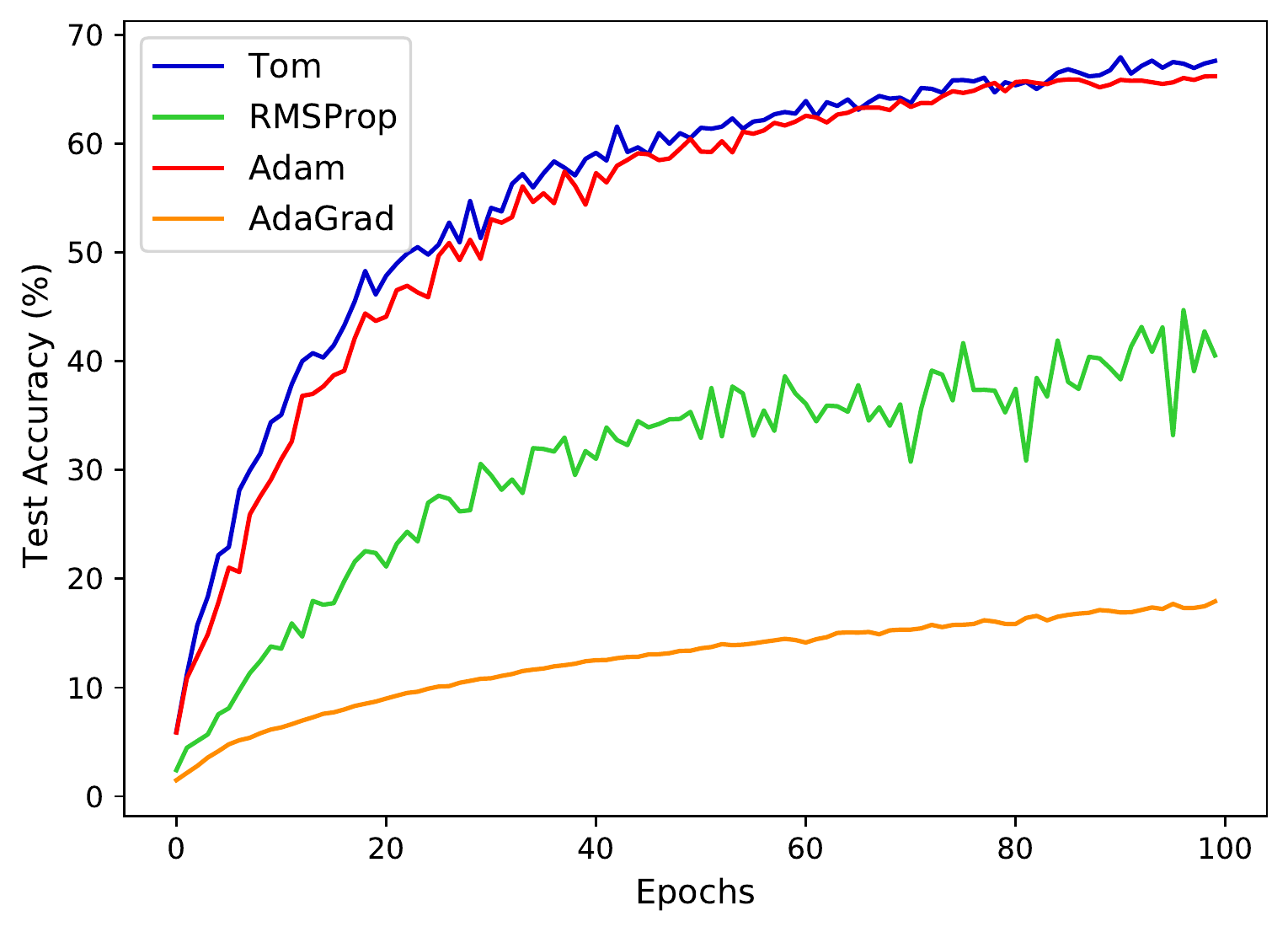}
         \caption{Test accuracy for ResNet-18 with batch size 4096 on CIFAR-100}
     \end{subfigure}
        \caption{Training and test accuracy evolution for ResNet-18 with three different batch sizes (512, 1024, 4096) on CIFAR-100}
        \label{fig:resnet18_cifar100_fig}
\end{figure}

\begin{figure}[!htbp]
     \centering
     \begin{subfigure}[b]{0.3\textwidth}
         \centering
         \includegraphics[width=\textwidth]{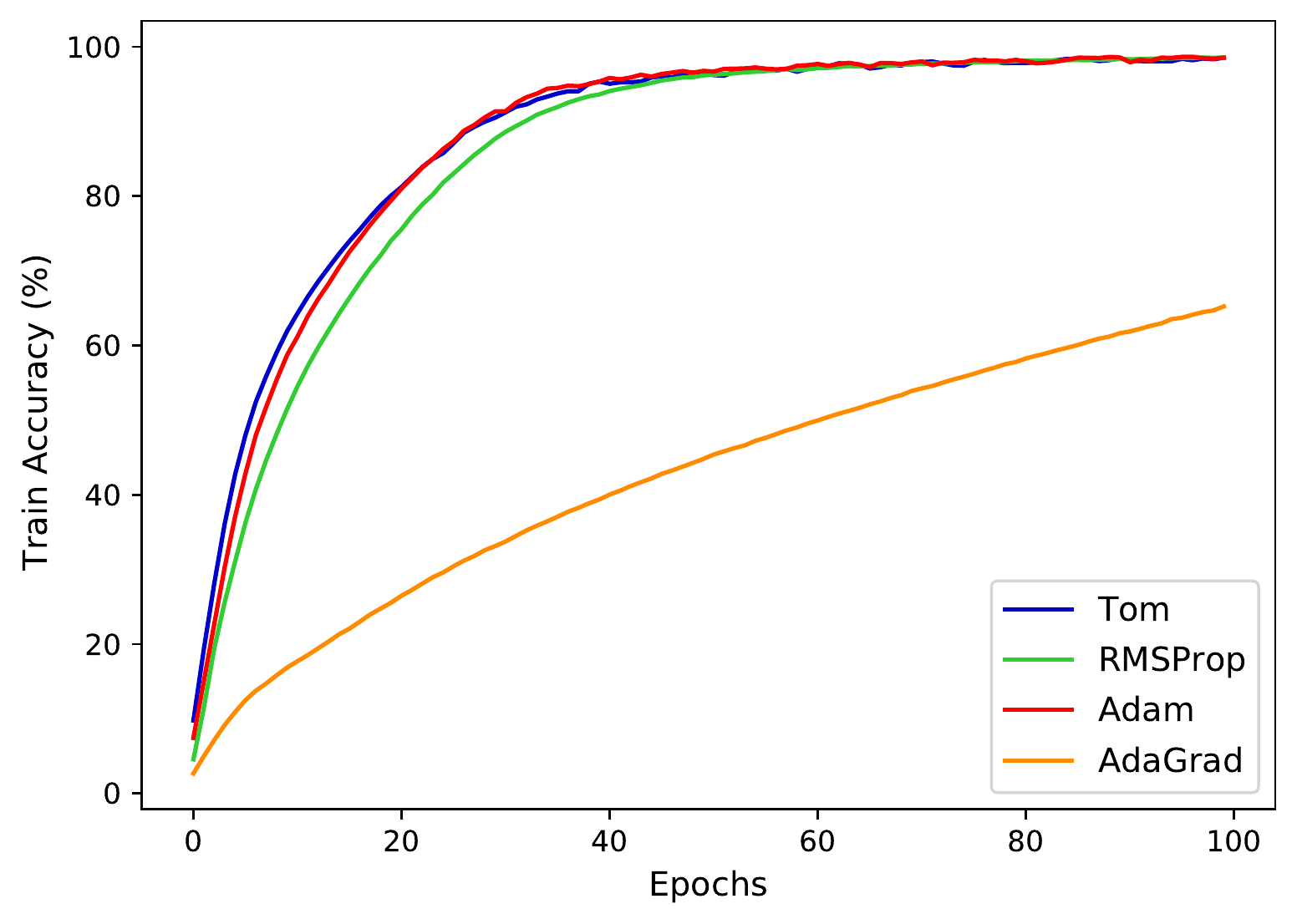}
         \caption{Training accuracy for ResNet-34 with batch size 512 on CIFAR-100}
     \end{subfigure}
     \hfill
     \begin{subfigure}[b]{0.3\textwidth}
         \centering
         \includegraphics[width=\textwidth]{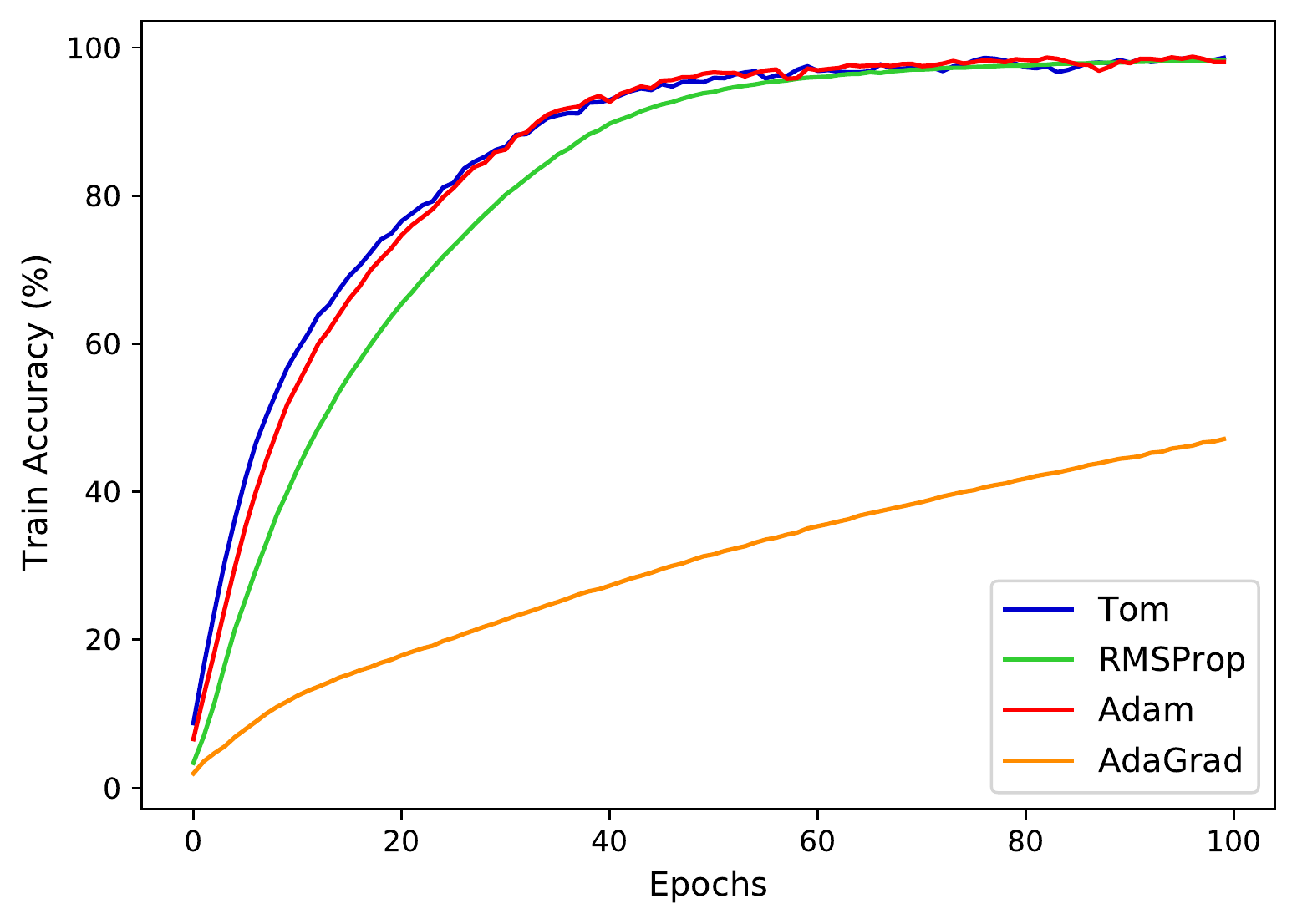}
         \caption{Training accuracy for ResNet-34 with batch size 1024 on CIFAR-100}
     \end{subfigure}
     \hfill
     \begin{subfigure}[b]{0.3\textwidth}
         \centering
         \includegraphics[width=\textwidth]{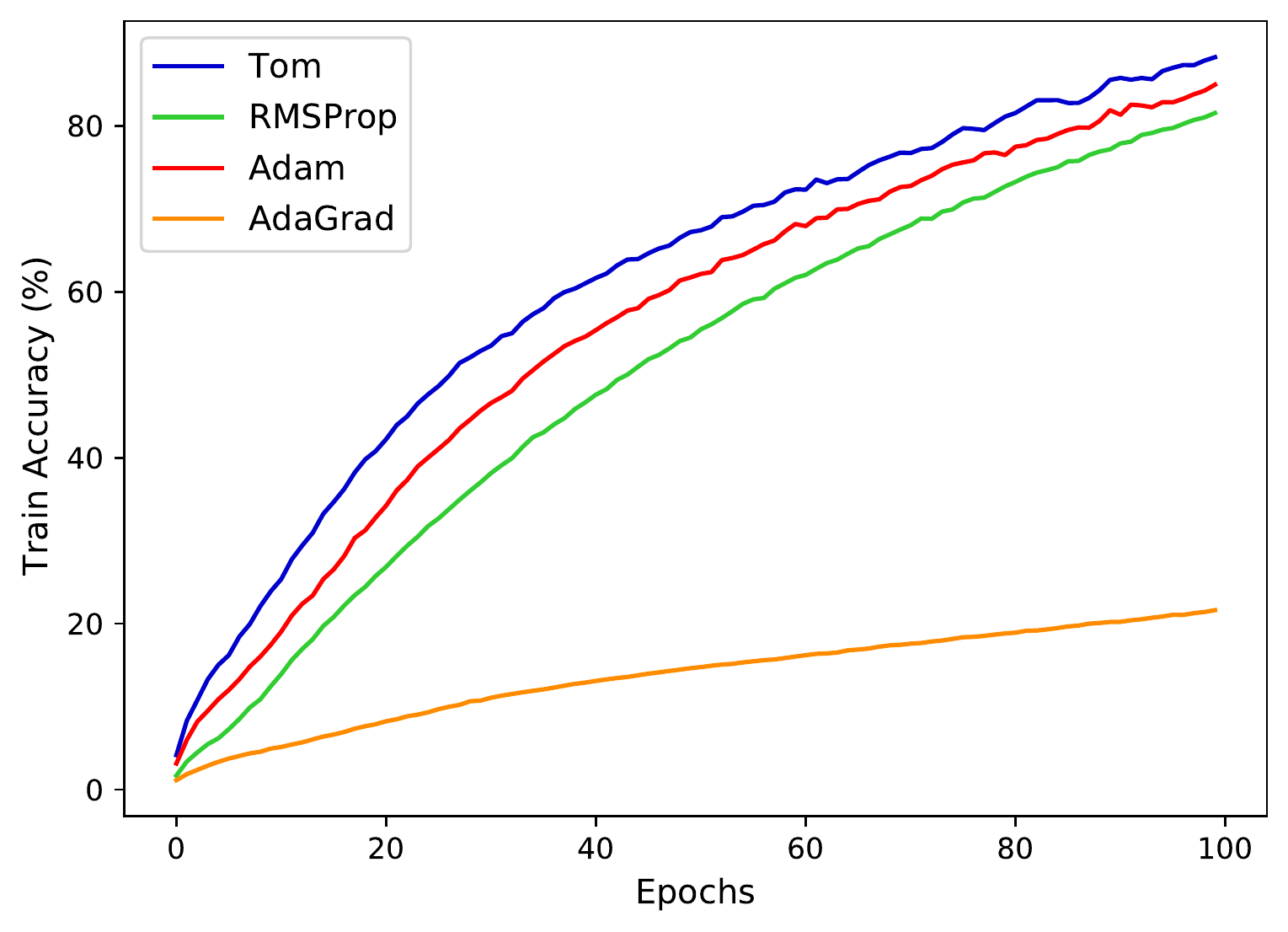}
         \caption{Training accuracy for ResNet-34 with batch size 4096 on CIFAR-100}
     \end{subfigure}
     \vfill
     \vspace{5mm}
     \begin{subfigure}[b]{0.3\textwidth}
         \centering
         \includegraphics[width=\textwidth]{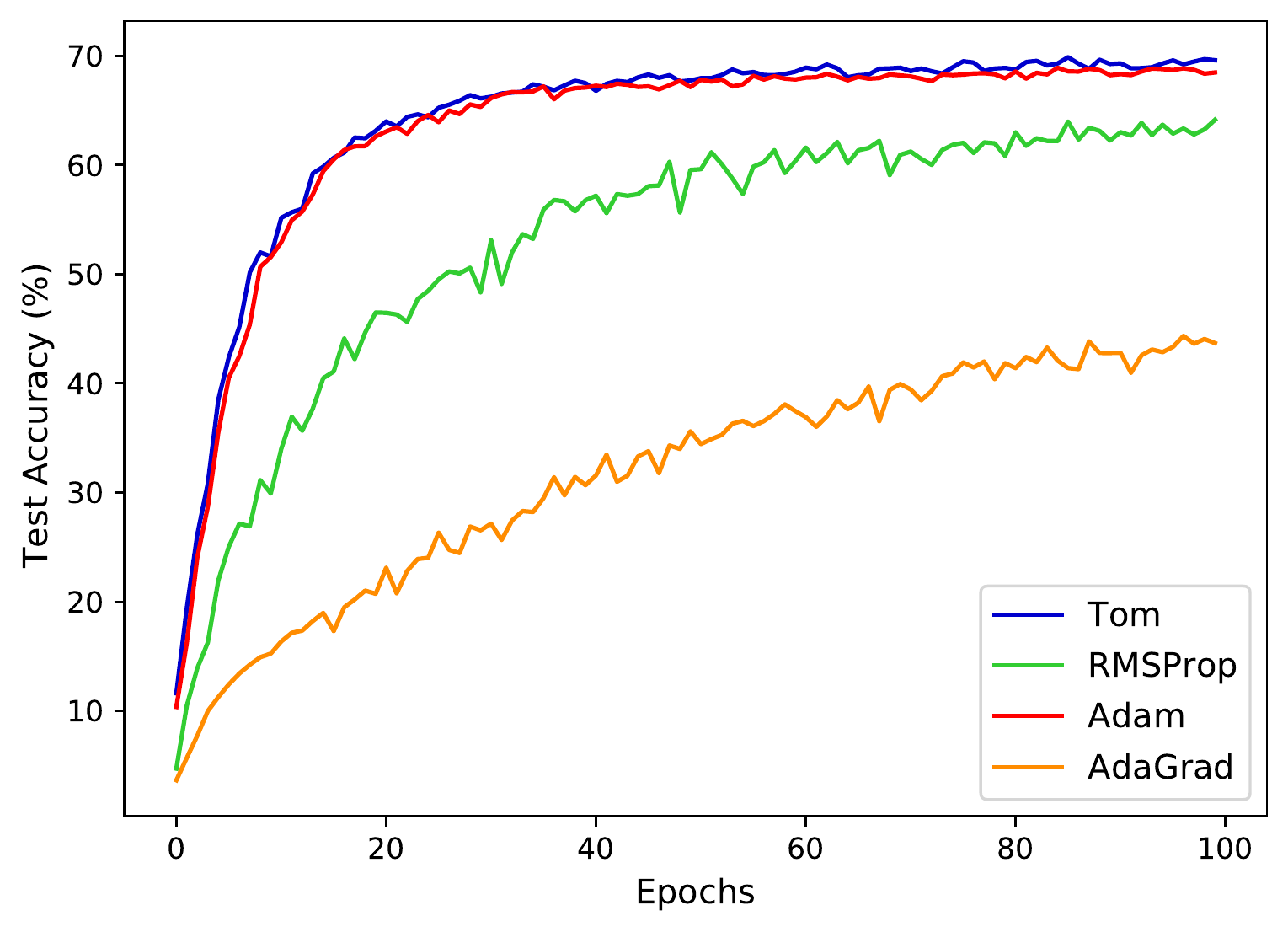}
         \caption{Test accuracy for ResNet-34 with batch size 512 on CIFAR-100}
     \end{subfigure}
     \hfill
     \begin{subfigure}[b]{0.3\textwidth}
         \centering
         \includegraphics[width=\textwidth]{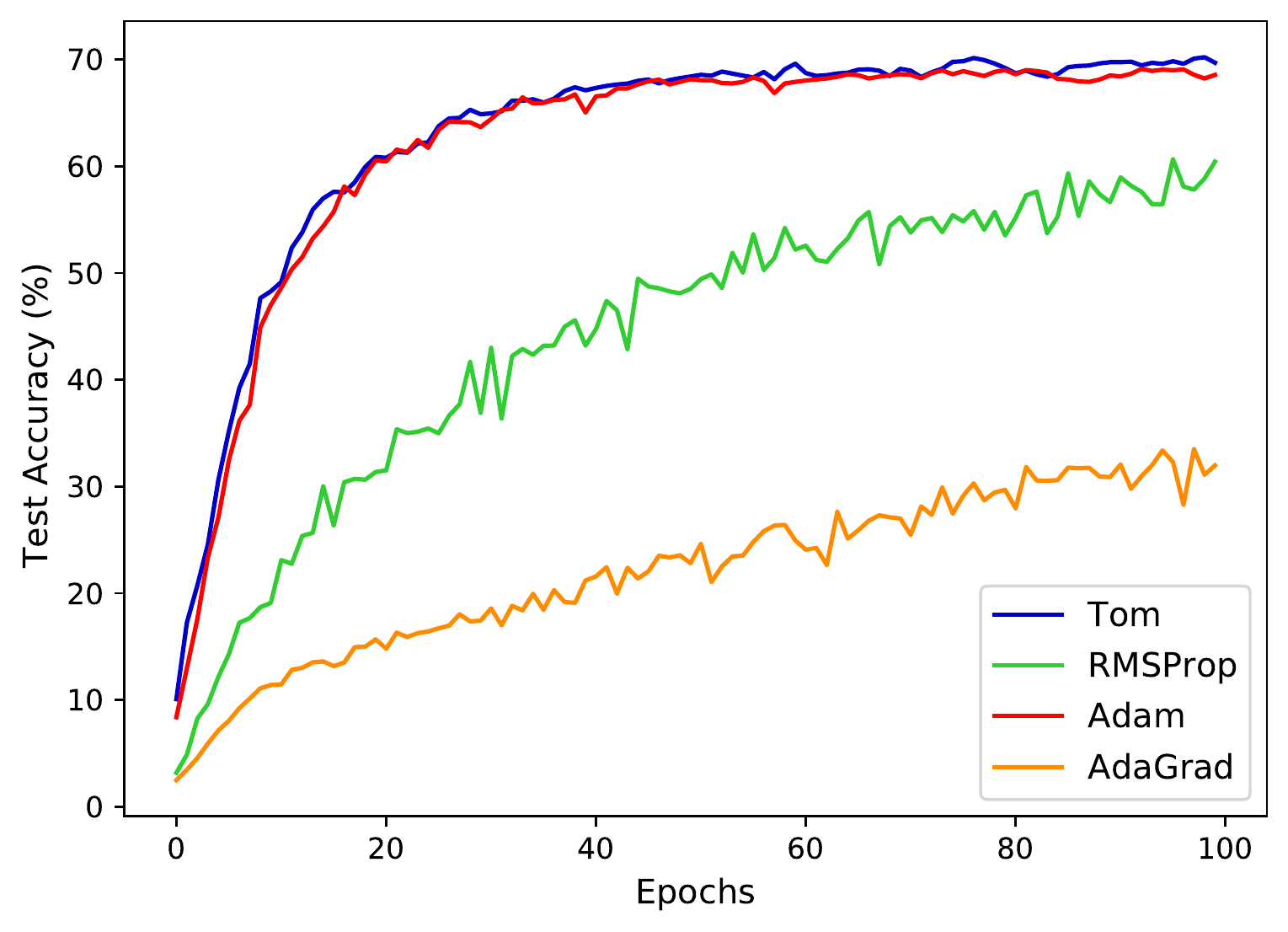}
         \caption{Test accuracy for ResNet-34 with batch size 1024 on CIFAR-100}
     \end{subfigure}
     \hfill
     \begin{subfigure}[b]{0.3\textwidth}
         \centering
         \includegraphics[width=\textwidth]{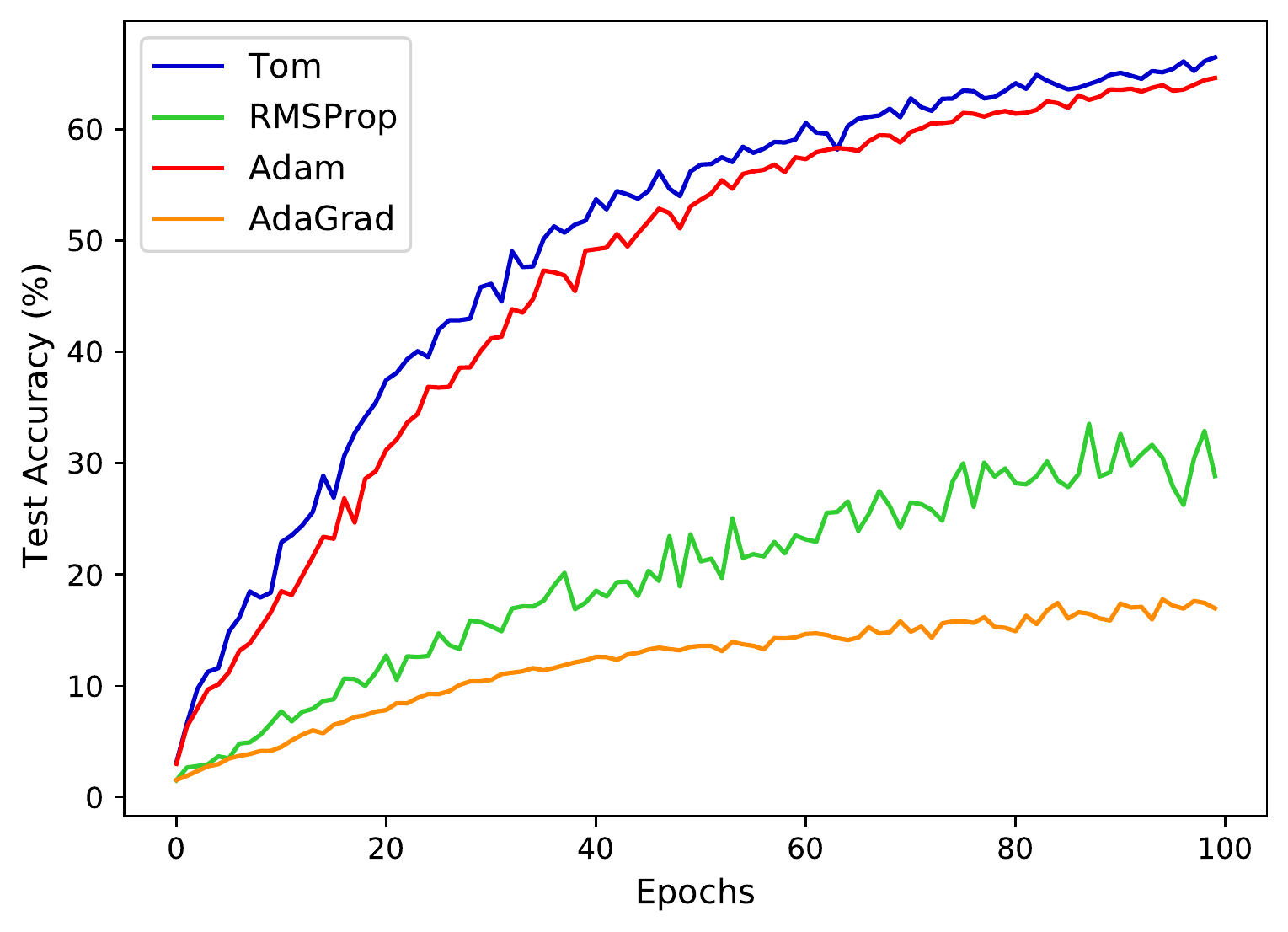}
         \caption{Test accuracy for ResNet-34 with batch size 4096 on CIFAR-100}
     \end{subfigure}
        \caption{Training and test accuracy evolution for ResNet-34 with three different batch sizes (512, 1024, 4096) on CIFAR-100}
        \label{fig:resnet34_cifar100_fig}
\end{figure}

\begin{figure}[!htbp]
     \centering
     \begin{subfigure}[b]{0.3\textwidth}
         \centering
         \includegraphics[width=\textwidth]{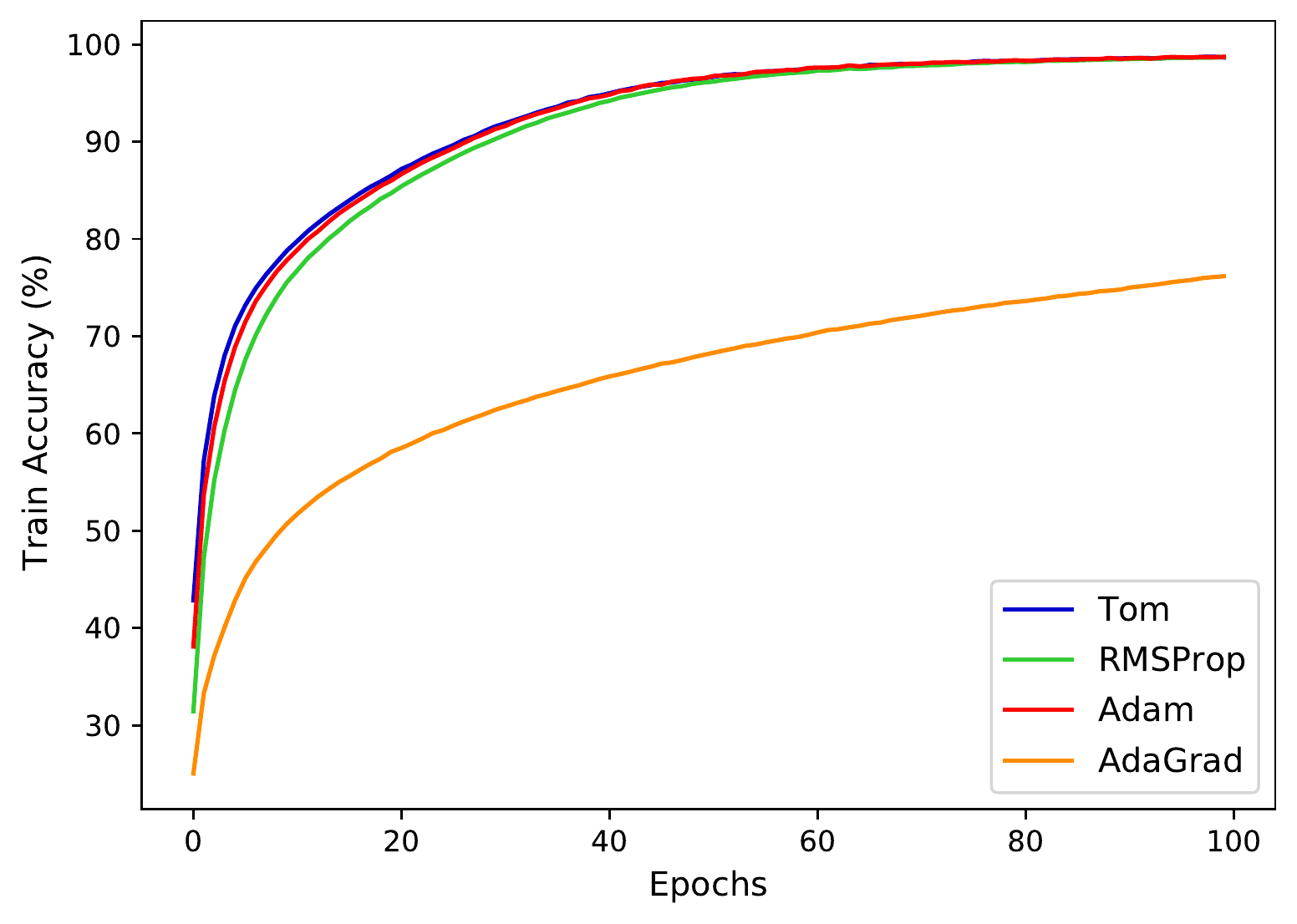}
         \caption{Training accuracy for ResNet-18 with batch size 512 on CINIC-10}
     \end{subfigure}
     \hfill
     \begin{subfigure}[b]{0.3\textwidth}
         \centering
         \includegraphics[width=\textwidth]{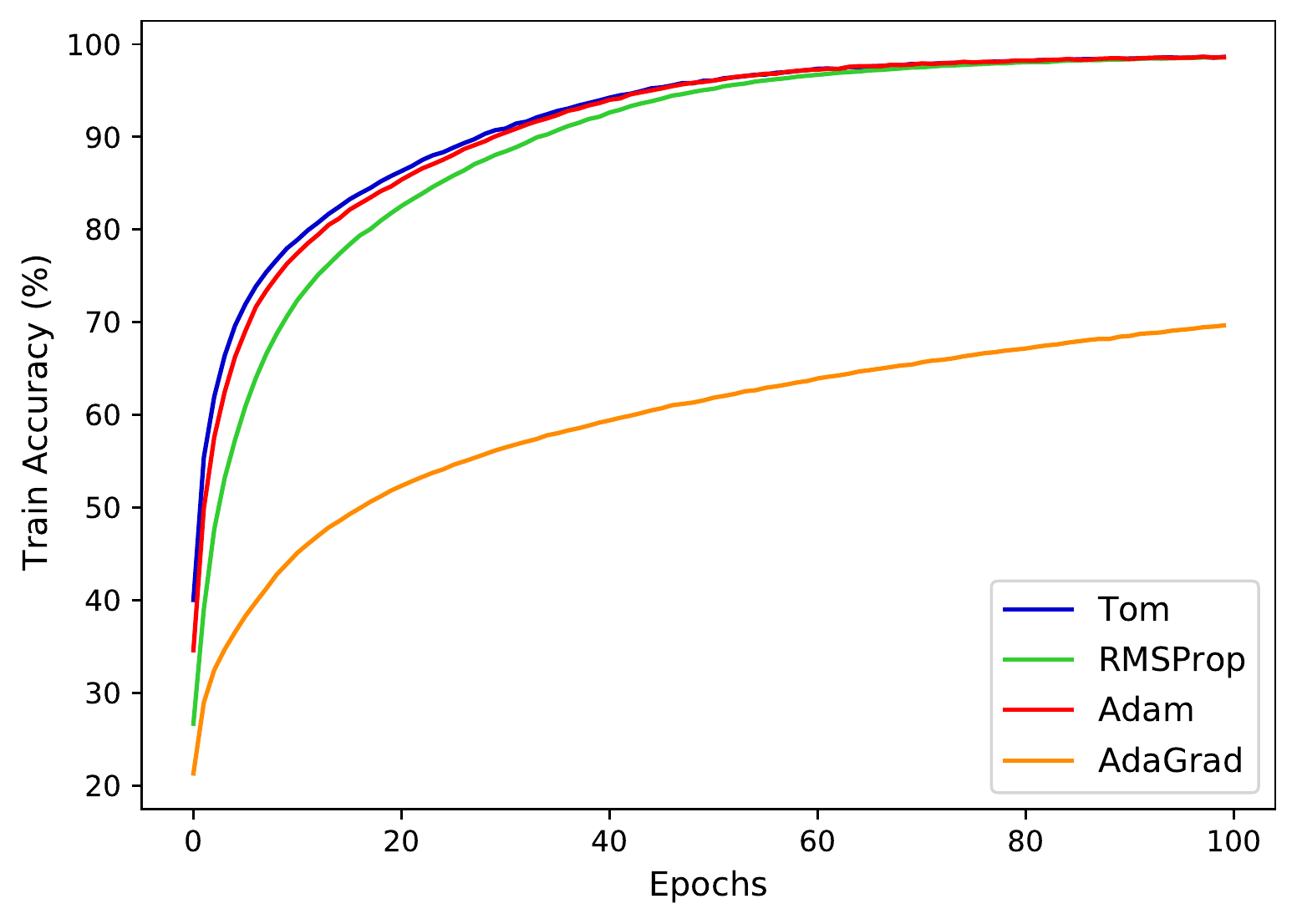}
         \caption{Training accuracy for ResNet-18 with batch size 1024 on CINIC-10}
     \end{subfigure}
     \hfill
     \begin{subfigure}[b]{0.3\textwidth}
         \centering
         \includegraphics[width=\textwidth]{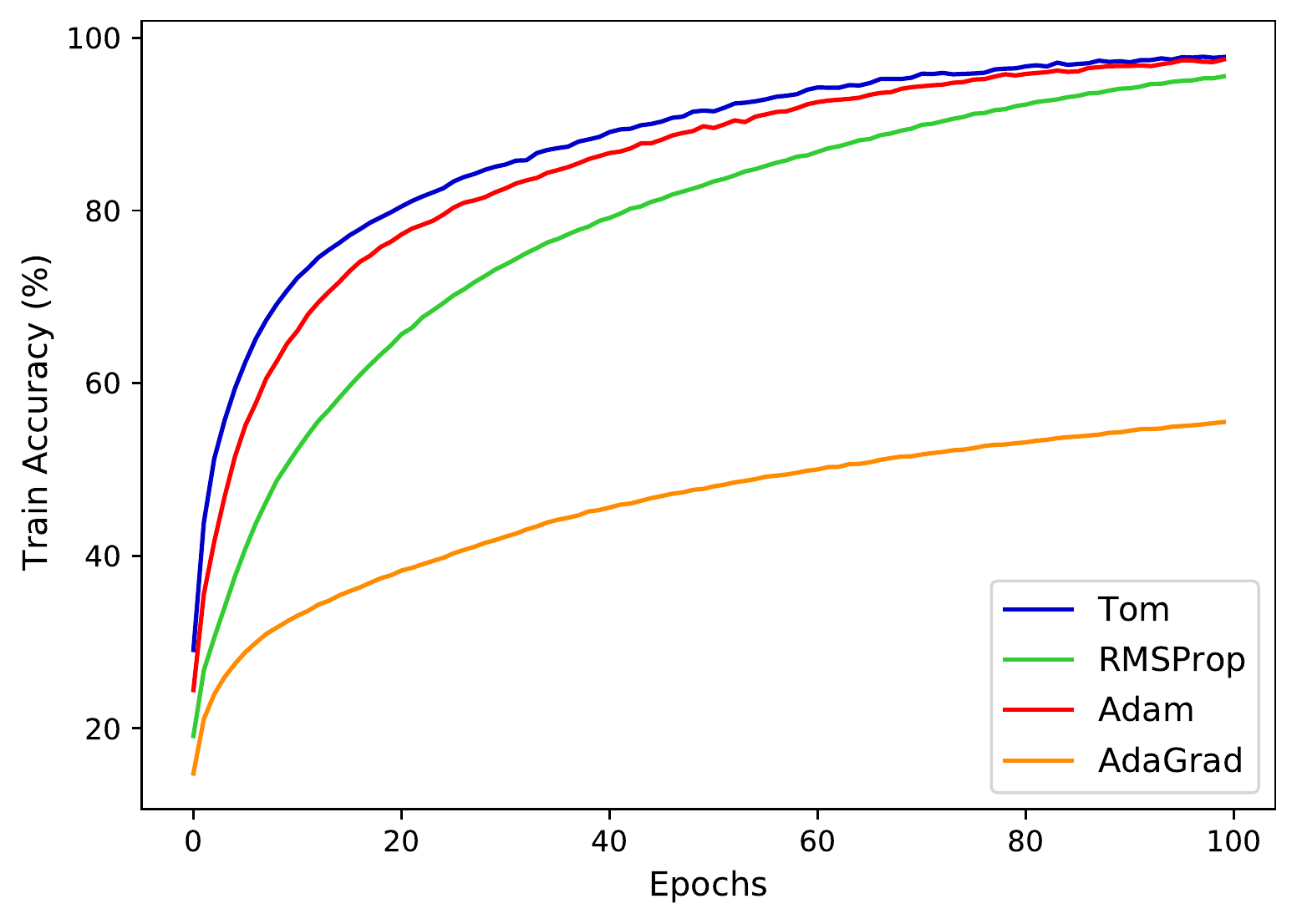}
         \caption{Training accuracy for ResNet-18 with batch size 4096 on CINIC-10}
     \end{subfigure}
     \vfill
     \vspace{5mm}
     \begin{subfigure}[b]{0.3\textwidth}
         \centering
         \includegraphics[width=\textwidth]{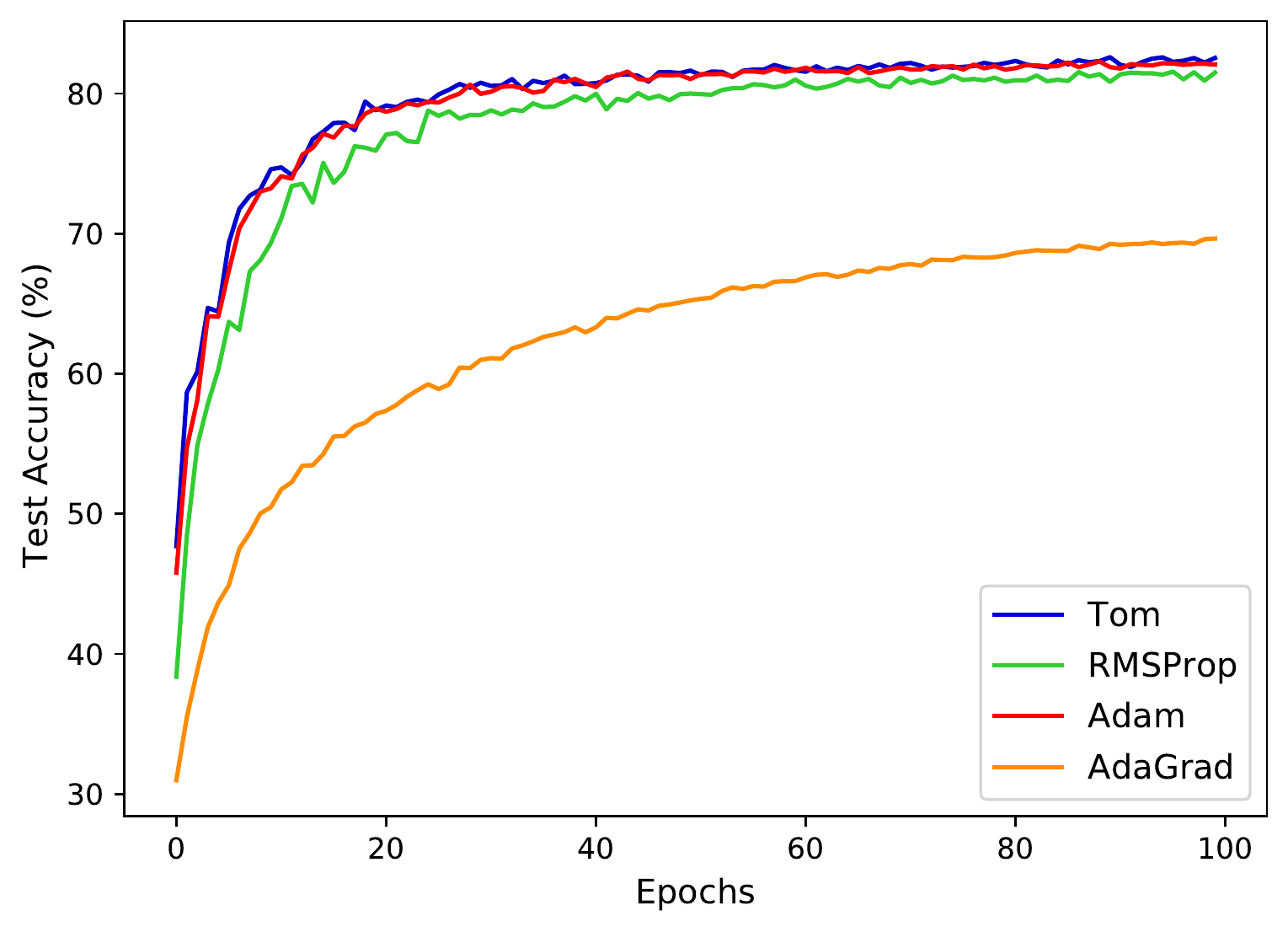}
         \caption{Test accuracy for ResNet-18 with batch size 512 on CINIC-10}
     \end{subfigure}
     \hfill
     \begin{subfigure}[b]{0.3\textwidth}
         \centering
         \includegraphics[width=\textwidth]{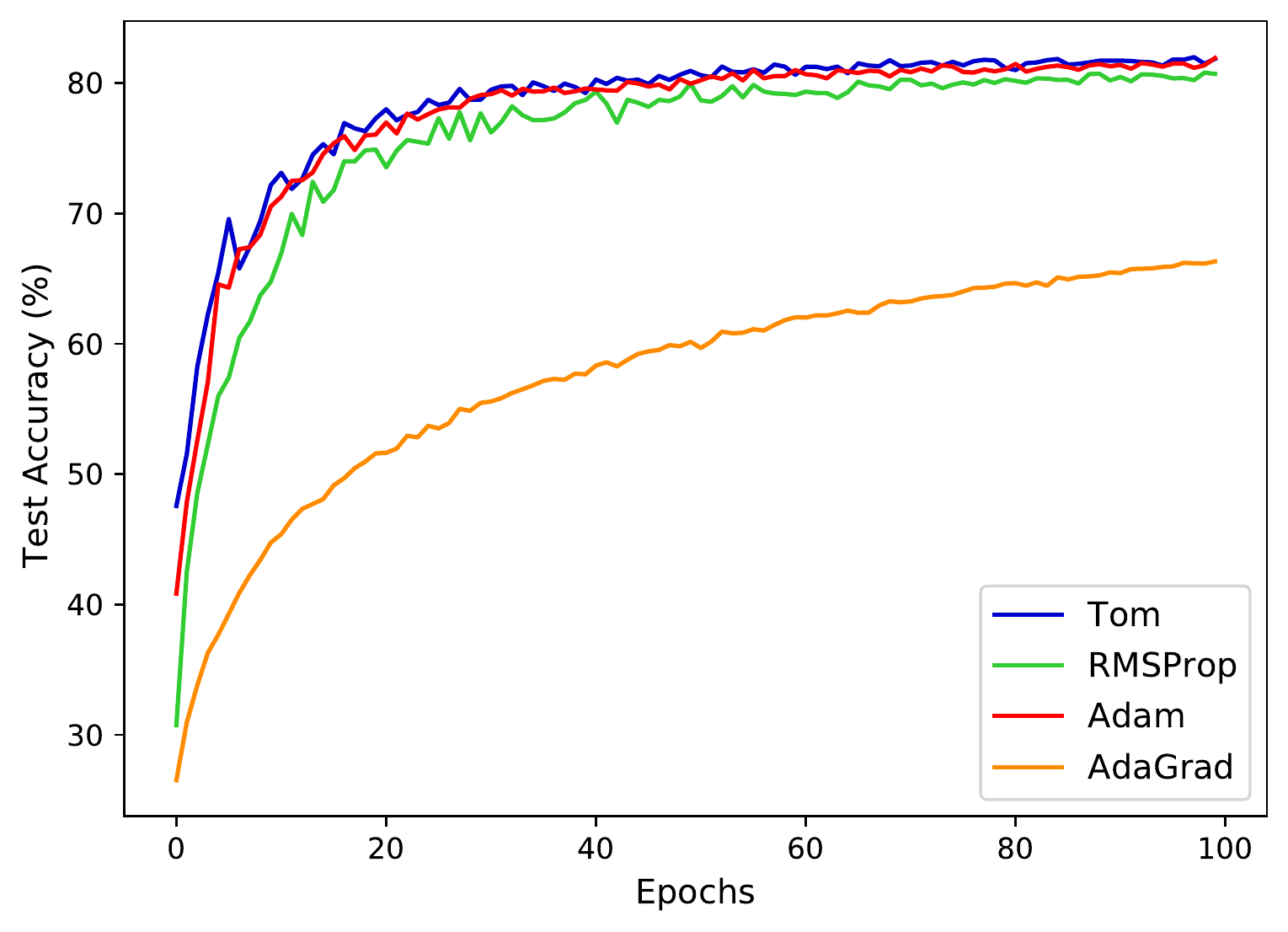}
         \caption{Test accuracy for ResNet-18 with batch size 1024 on CINIC-10}
     \end{subfigure}
     \hfill
     \begin{subfigure}[b]{0.3\textwidth}
         \centering
         \includegraphics[width=\textwidth]{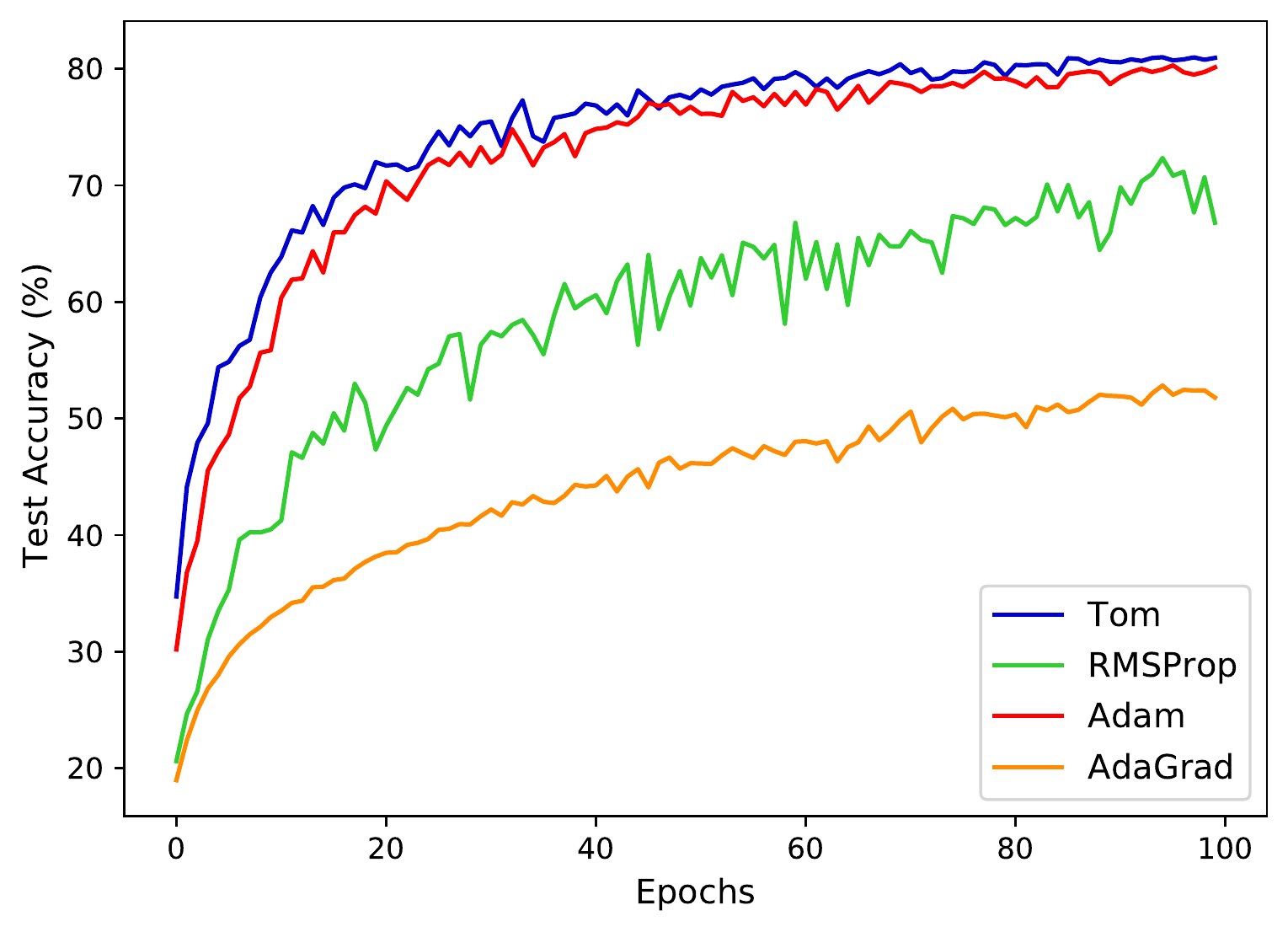}
         \caption{Test accuracy for ResNet-18 with batch size 4096 on CINIC-10}
     \end{subfigure}
        \caption{Training and test accuracy evolution for ResNet-18 with three different batch sizes (512, 1024, 4096) on CINIC-10}
        \label{fig:resnet18_cinic10_fig}
\end{figure}

\begin{figure}[!htbp]
     \centering
     \begin{subfigure}[b]{0.3\textwidth}
         \centering
         \includegraphics[width=\textwidth]{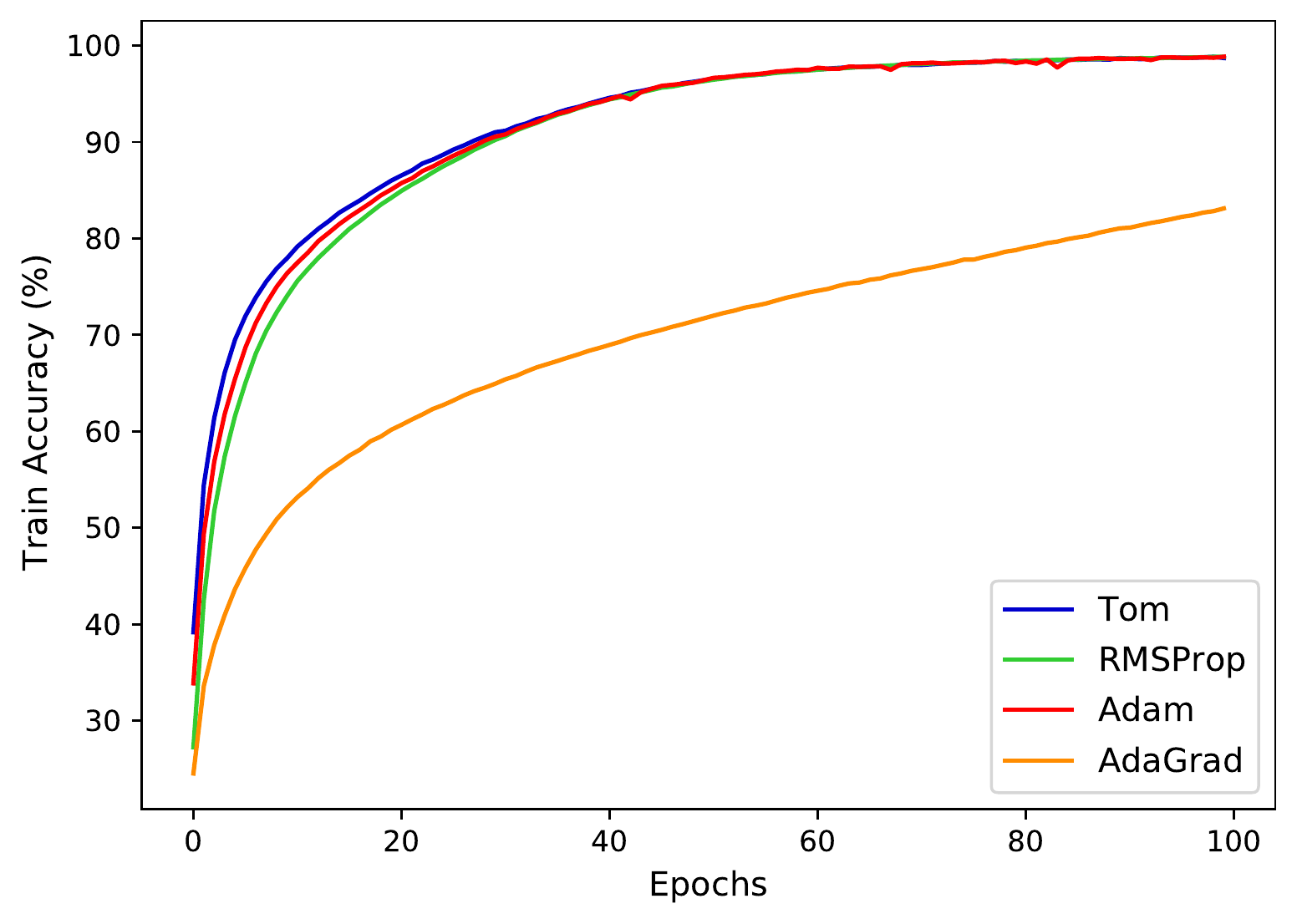}
         \caption{Training accuracy for ResNet-34 with batch size 512 on CINIC-10}
     \end{subfigure}
     \hfill
     \begin{subfigure}[b]{0.3\textwidth}
         \centering
         \includegraphics[width=\textwidth]{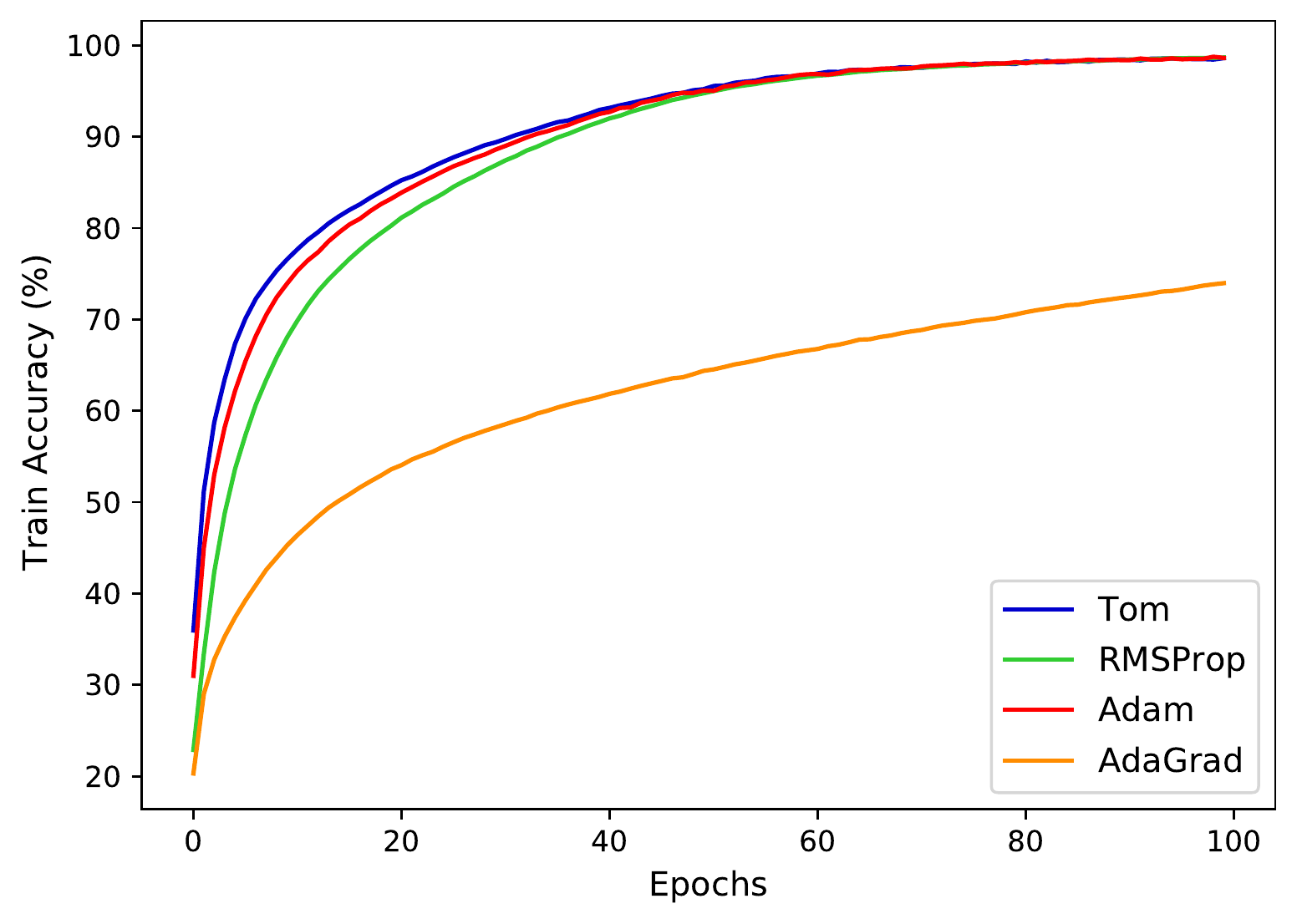}
         \caption{Training accuracy for ResNet-34 with batch size 1024 on CINIC-10}
     \end{subfigure}
     \hfill
     \begin{subfigure}[b]{0.3\textwidth}
         \centering
         \includegraphics[width=\textwidth]{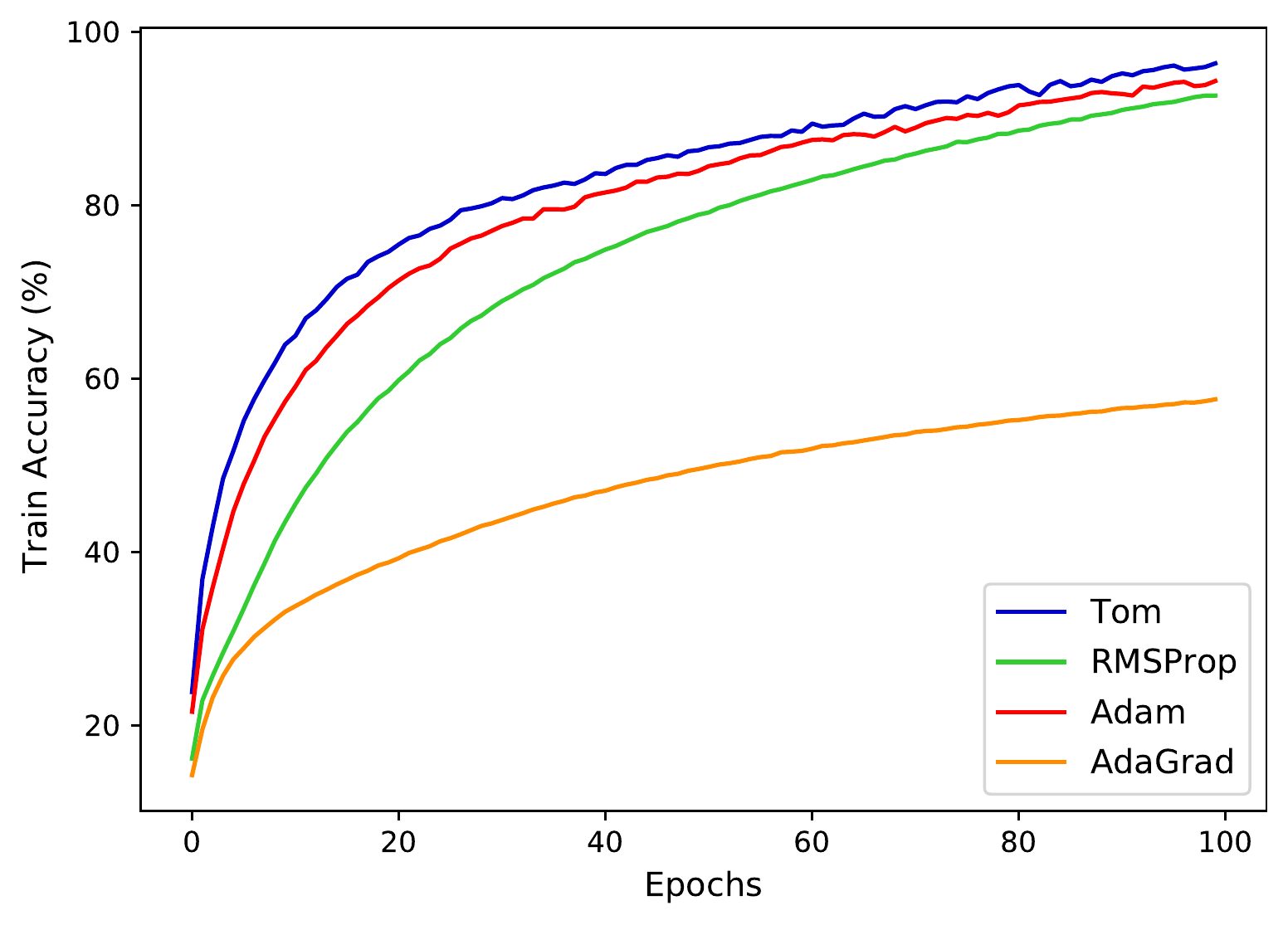}
         \caption{Training accuracy for ResNet-34 with batch size 4096 on CINIC-10}
     \end{subfigure}
     \vfill
     \vspace{5mm}
     \begin{subfigure}[b]{0.3\textwidth}
         \centering
         \includegraphics[width=\textwidth]{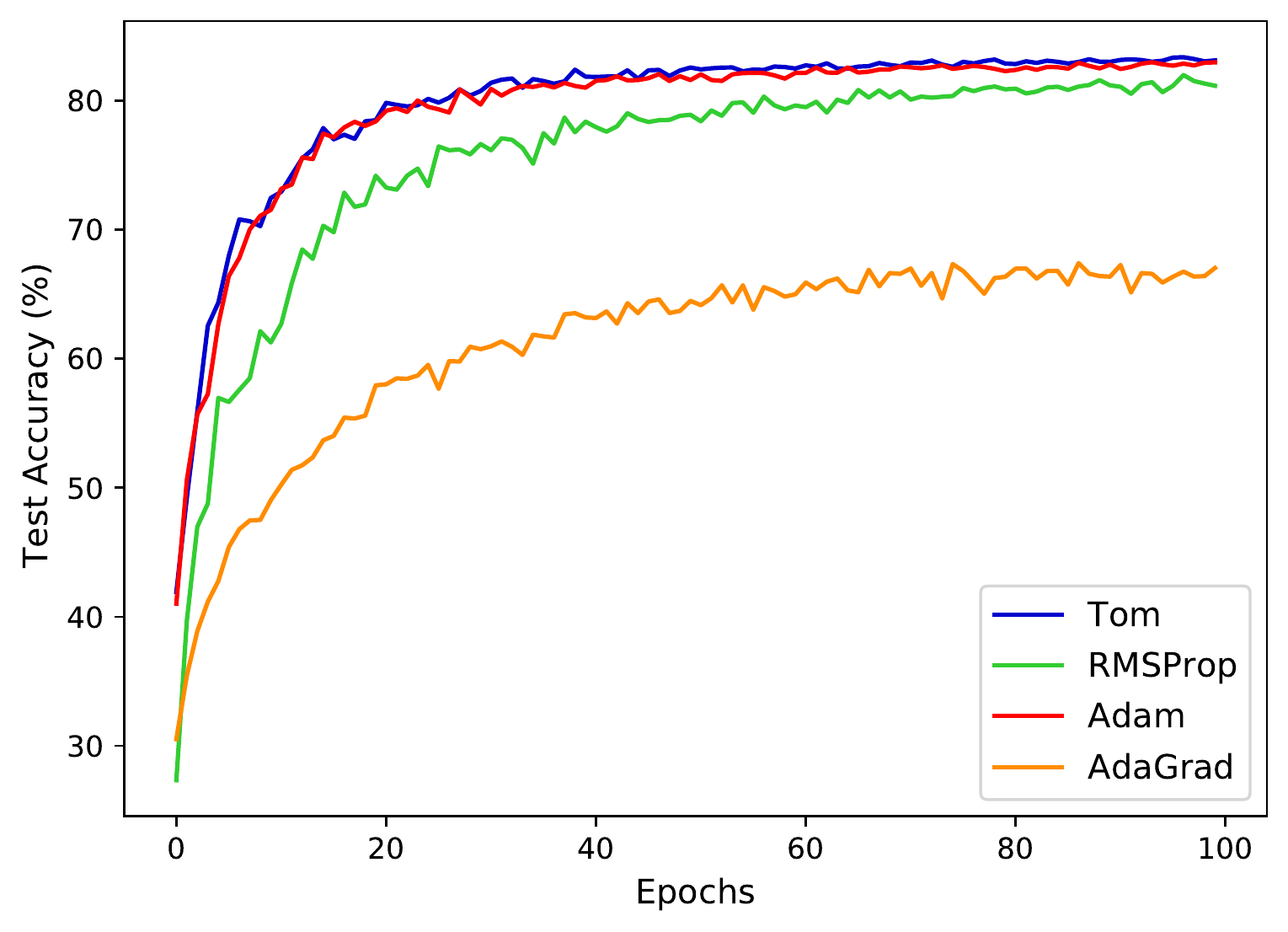}
         \caption{Test accuracy for ResNet-34 with batch size 512 on CINIC-10}
     \end{subfigure}
     \hfill
     \begin{subfigure}[b]{0.3\textwidth}
         \centering
         \includegraphics[width=\textwidth]{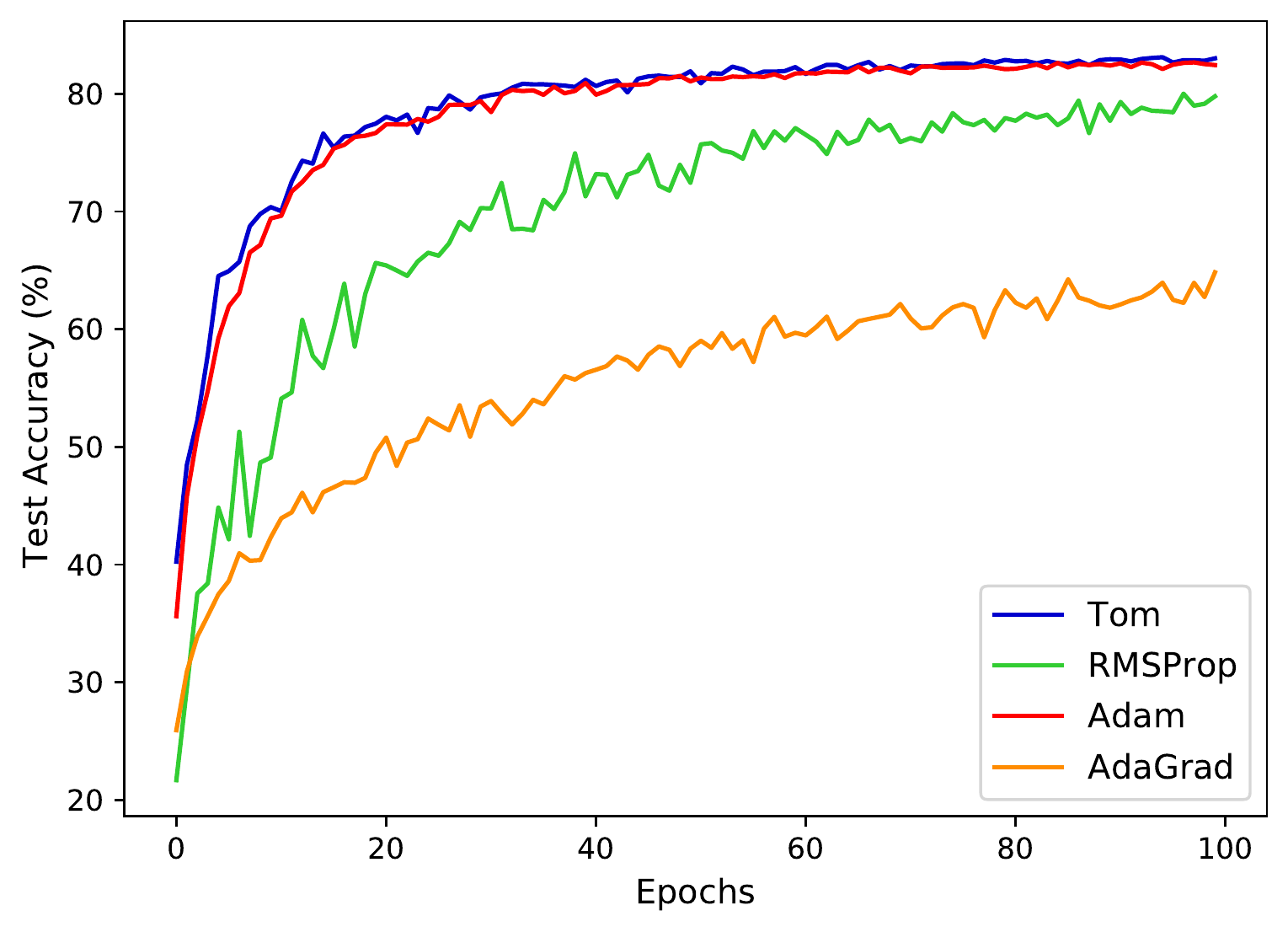}
         \caption{Test accuracy for ResNet-34 with batch size 1024 on CINIC-10}
     \end{subfigure}
     \hfill
     \begin{subfigure}[b]{0.3\textwidth}
         \centering
         \includegraphics[width=\textwidth]{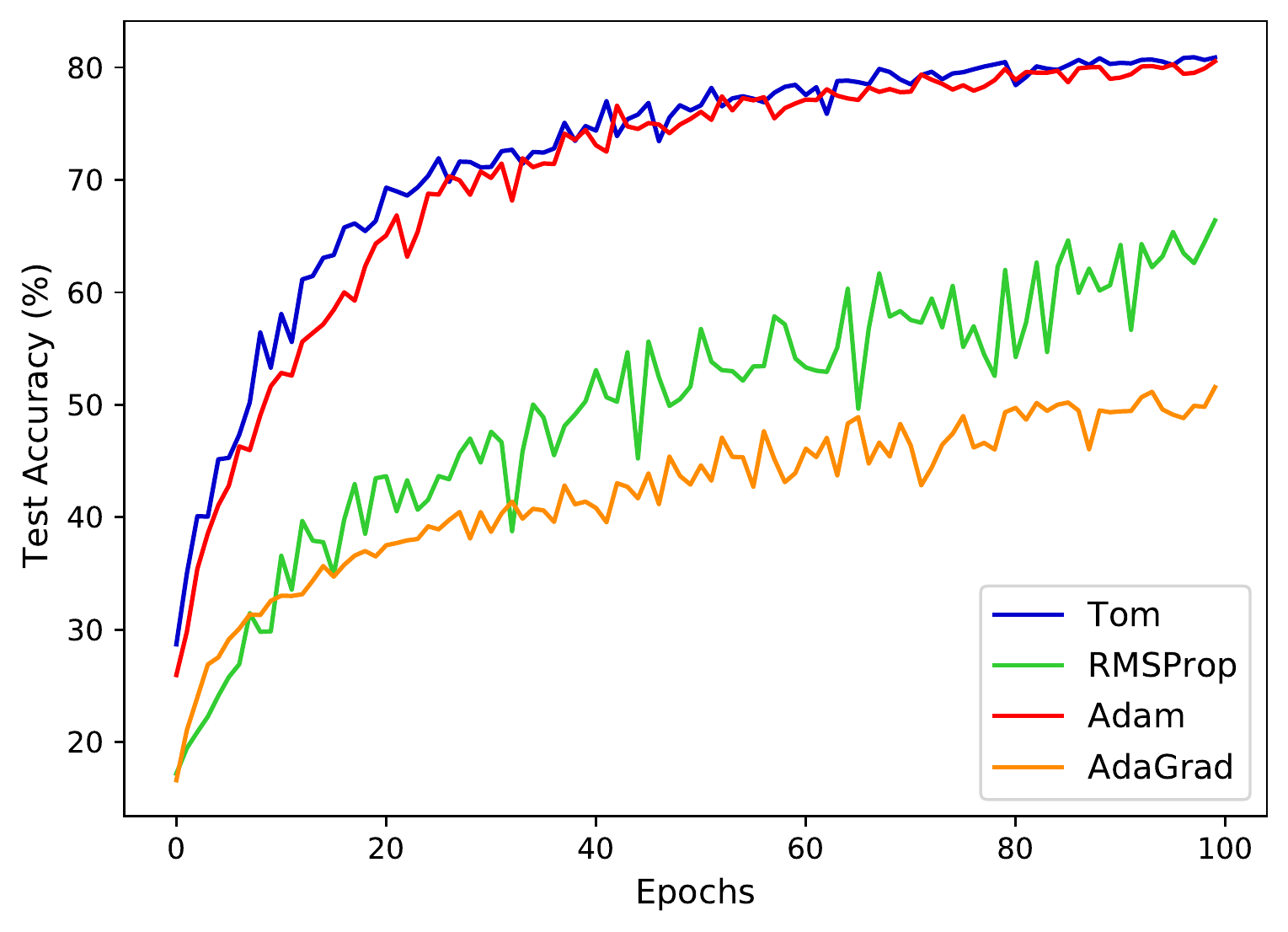}
         \caption{Test accuracy for ResNet-34 with batch size 4096 on CINIC-10}
     \end{subfigure}
        \caption{Training and test accuracy evolution for ResNet-34 with three different batch sizes (512, 1024, 4096) on CINIC-10}
        \label{fig:resnet34_cinic10_fig}
\end{figure}

The test accuracy of ResNet-18 and ResNet-34 for CIFAR-10, CIFAR-100 and CINIC-10 datasets across 3 different batch sizes is tabulated in Table~[\ref{tab:resnet18_cifar10_tab},\ref{tab:resnet34_cifar10_tab},\ref{tab:resnet18_cifar100_tab},\ref{tab:resnet34_cifar100_tab},\ref{tab:resnet18_cinic10_tab},\ref{tab:resnet34_cinic10_tab}]. We run each experiment five times with five different seeds and report the mean and standard deviation of the test accuracy. \emph{Tom} performs consistently better when compared to RMSProp, Adam, Adagrad. \emph{Tom} performs significantly better in the initial phase of training due to the addition of trend component and has a test accuracy gain as much as 6.68\% (CIFAR-10,ResNet-18, Epoch 10, Batch-size 4096). We can also observe that the gain in accuracy is higher as the batch size increases. Our hypothesis is that the trend observed is more representative and accurate as the number of data points in a gradient step increases. The train and test accuracy evolution shown in Figure~[\ref{fig:resnet18_cifar10_fig},\ref{fig:resnet34_cifar10_fig},\ref{fig:resnet18_cifar100_fig},\ref{fig:resnet34_cifar100_fig},\ref{fig:resnet18_cinic10_fig},\ref{fig:resnet34_cinic10_fig}] depicts that \emph{Tom} is on par or makes rapid progress when compared to other optimizers due to the addition of trend component. 
\subsection{Regression Settings}
\begin{figure}[!htbp]
     \centering
     \begin{subfigure}[b]{0.4\textwidth}
         \centering
         \includegraphics[width=\textwidth]{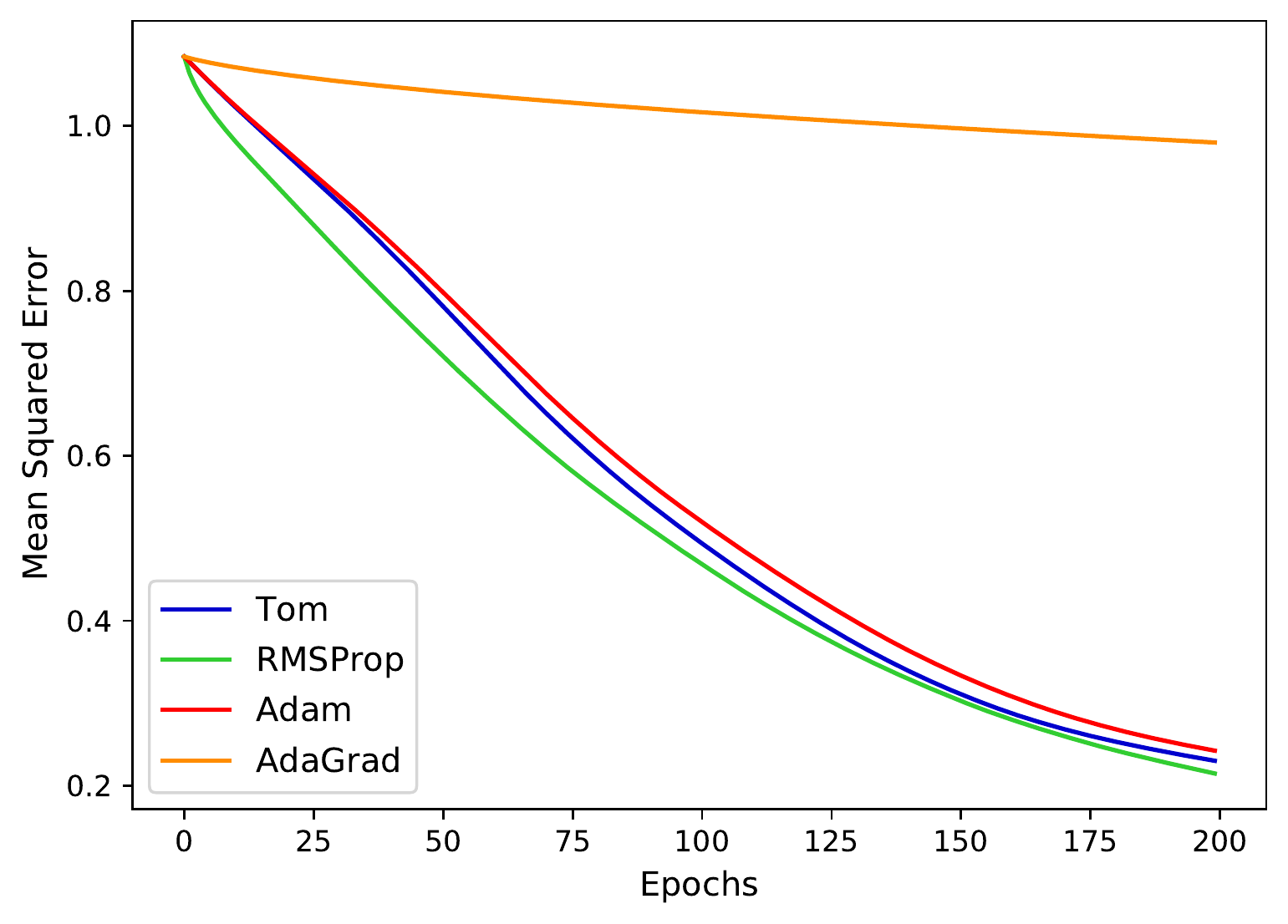}
         \caption{Mean squared error on the training set for Boston Housing dataset}
     \end{subfigure}
     \hfill
     \begin{subfigure}[b]{0.4\textwidth}
         \centering
         \includegraphics[width=\textwidth]{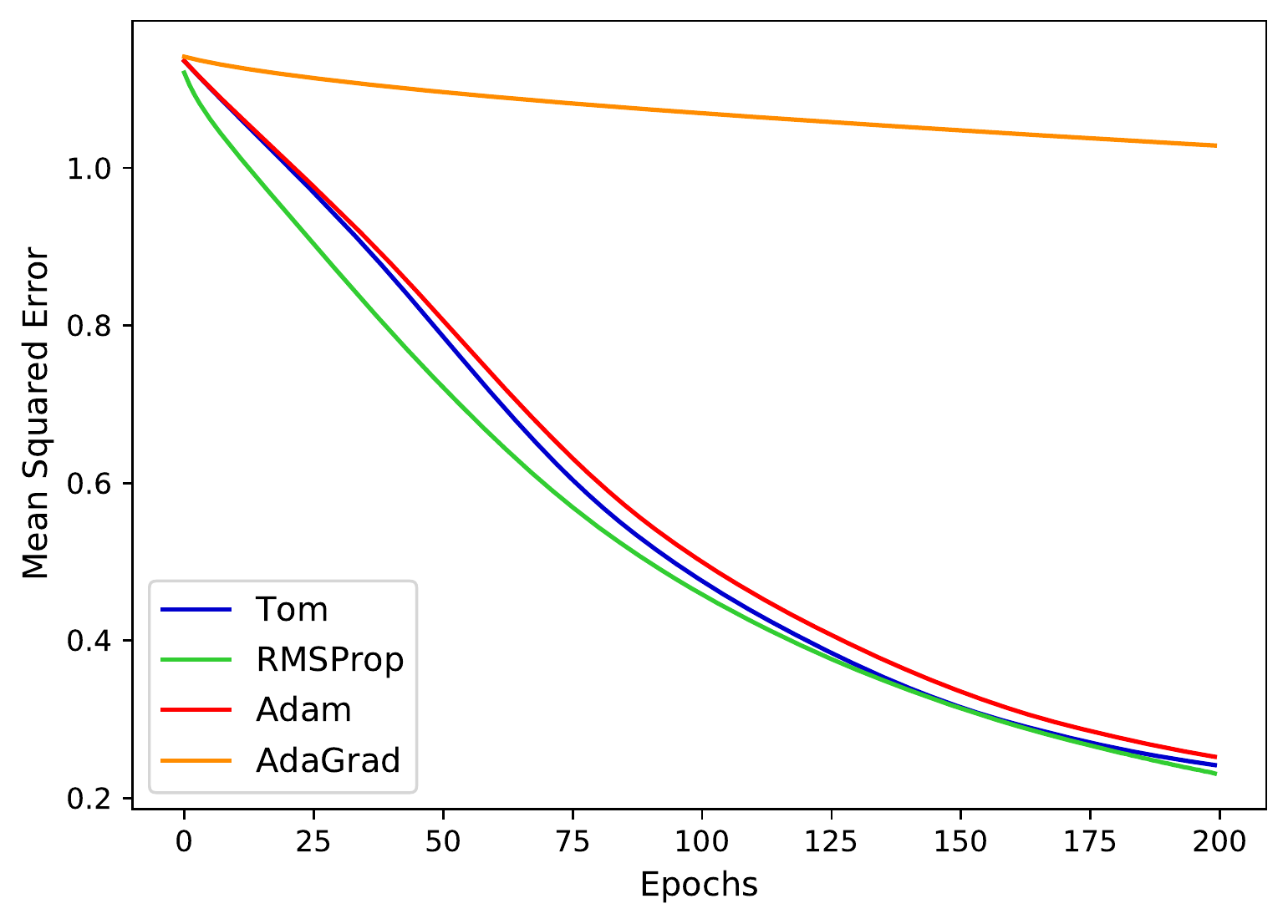}
         \caption{Mean squared error on the test set for Boston Housing dataset}
     \end{subfigure}
        \caption{Mean squared error evolution for the training and test set on Boston Housing dataset}
        \label{fig:boston_reg_fig}
\end{figure}

\begin{figure}[!htbp]
     \centering
     \begin{subfigure}[b]{0.4\textwidth}
         \centering
         \includegraphics[width=\textwidth]{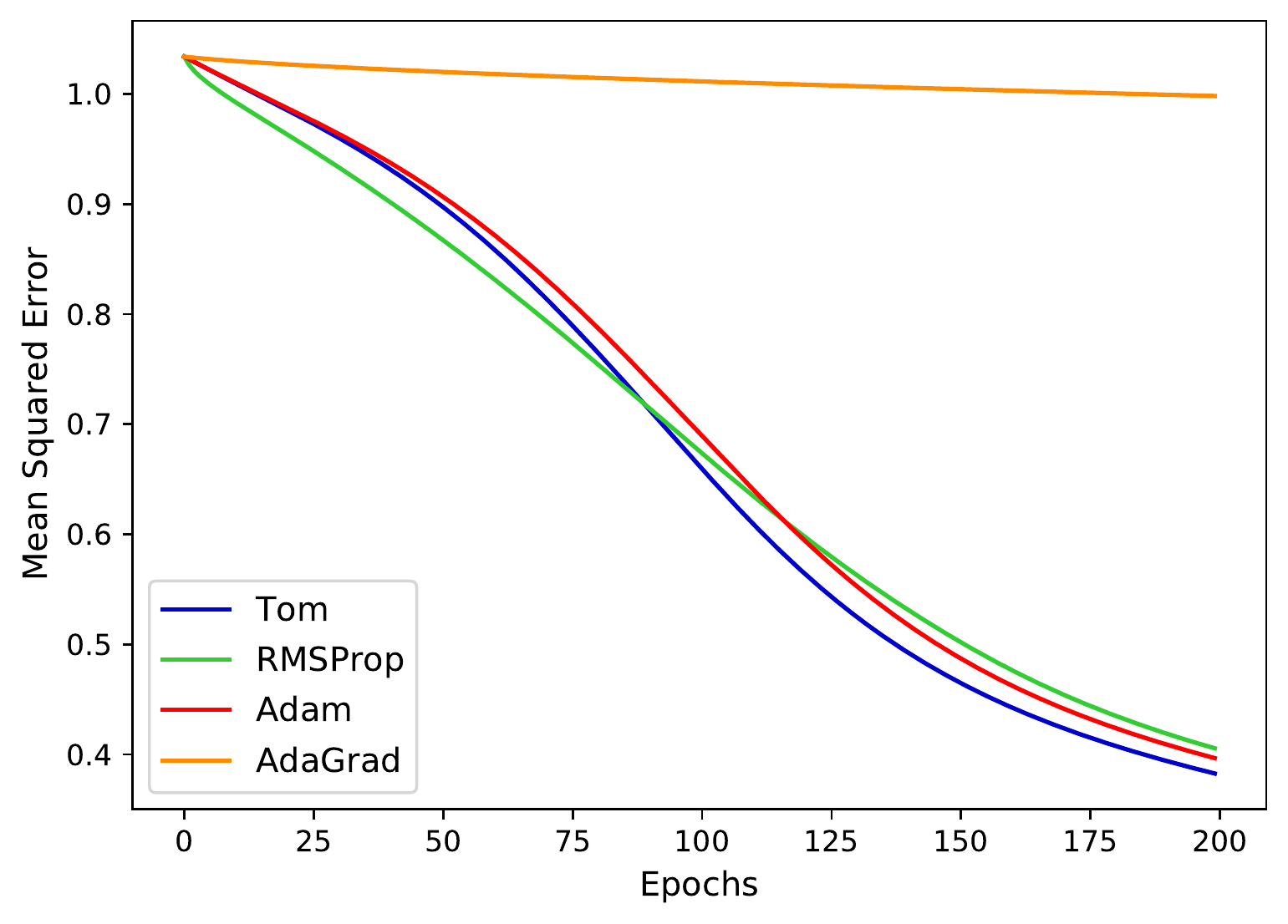}
         \caption{Mean squared error on the training set for California Housing dataset}
     \end{subfigure}
     \hfill
     \begin{subfigure}[b]{0.4\textwidth}
         \centering
         \includegraphics[width=\textwidth]{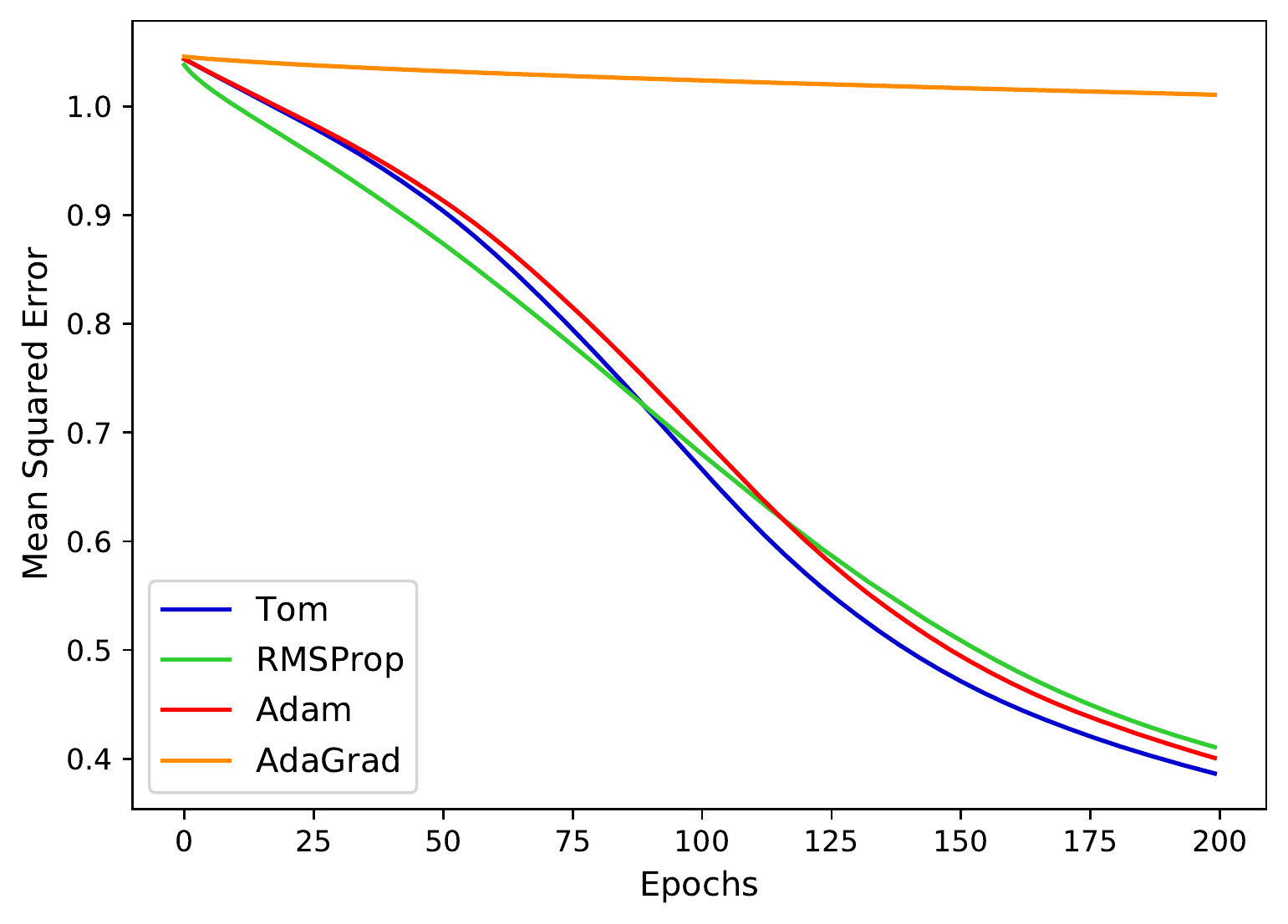}
         \caption{Mean squared error on the test set for California Housing dataset}
     \end{subfigure}
        \caption{Mean squared error evolution for the training and test set on California Housing dataset}
        \label{fig:cali_reg_fig}
\end{figure}

\begin{figure}[!htbp]
     \centering
     \begin{subfigure}[b]{0.4\textwidth}
         \centering
         \includegraphics[width=\textwidth]{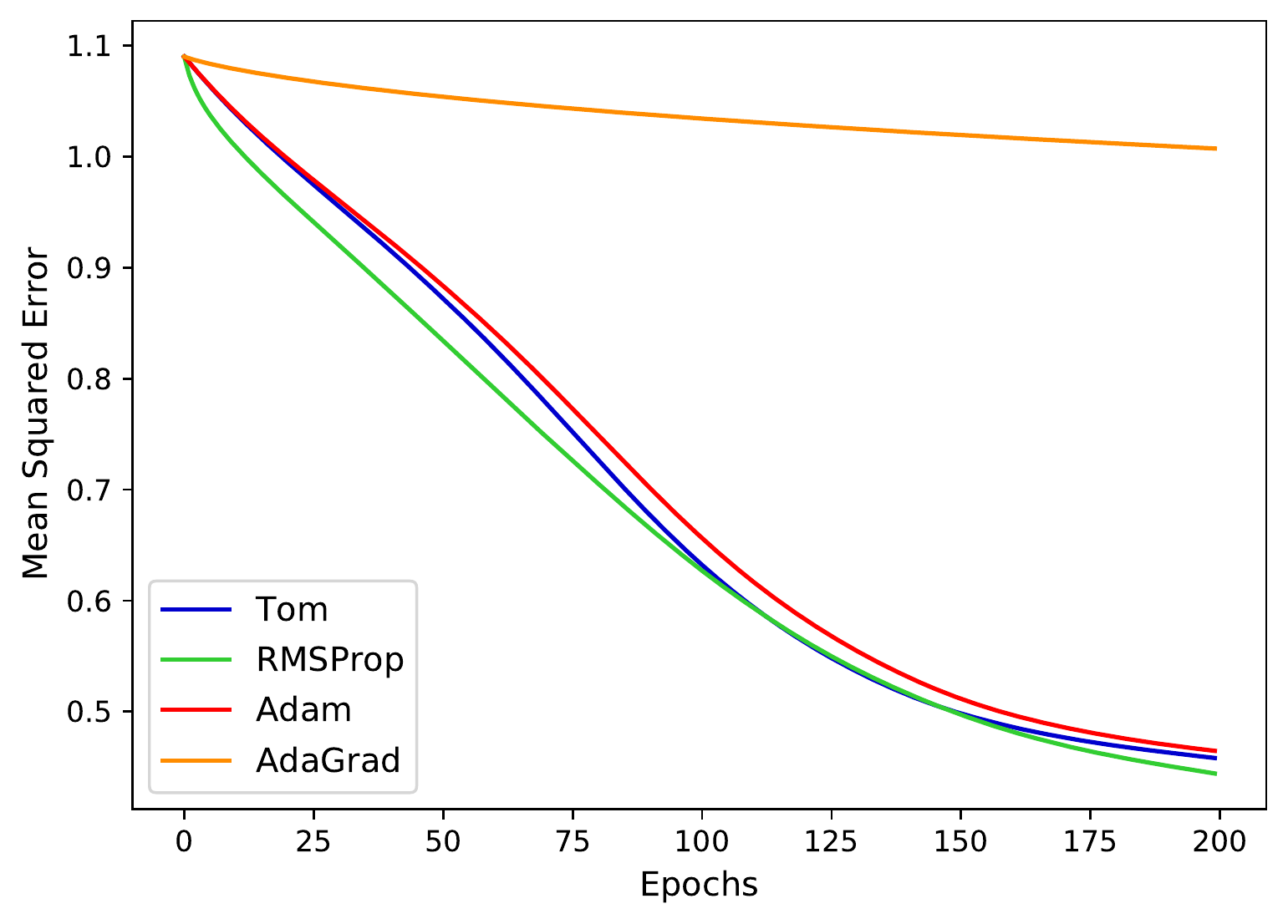}
         \caption{Mean squared error on the training set for Diabetes dataset}
     \end{subfigure}
     \hfill
     \begin{subfigure}[b]{0.4\textwidth}
         \centering
         \includegraphics[width=\textwidth]{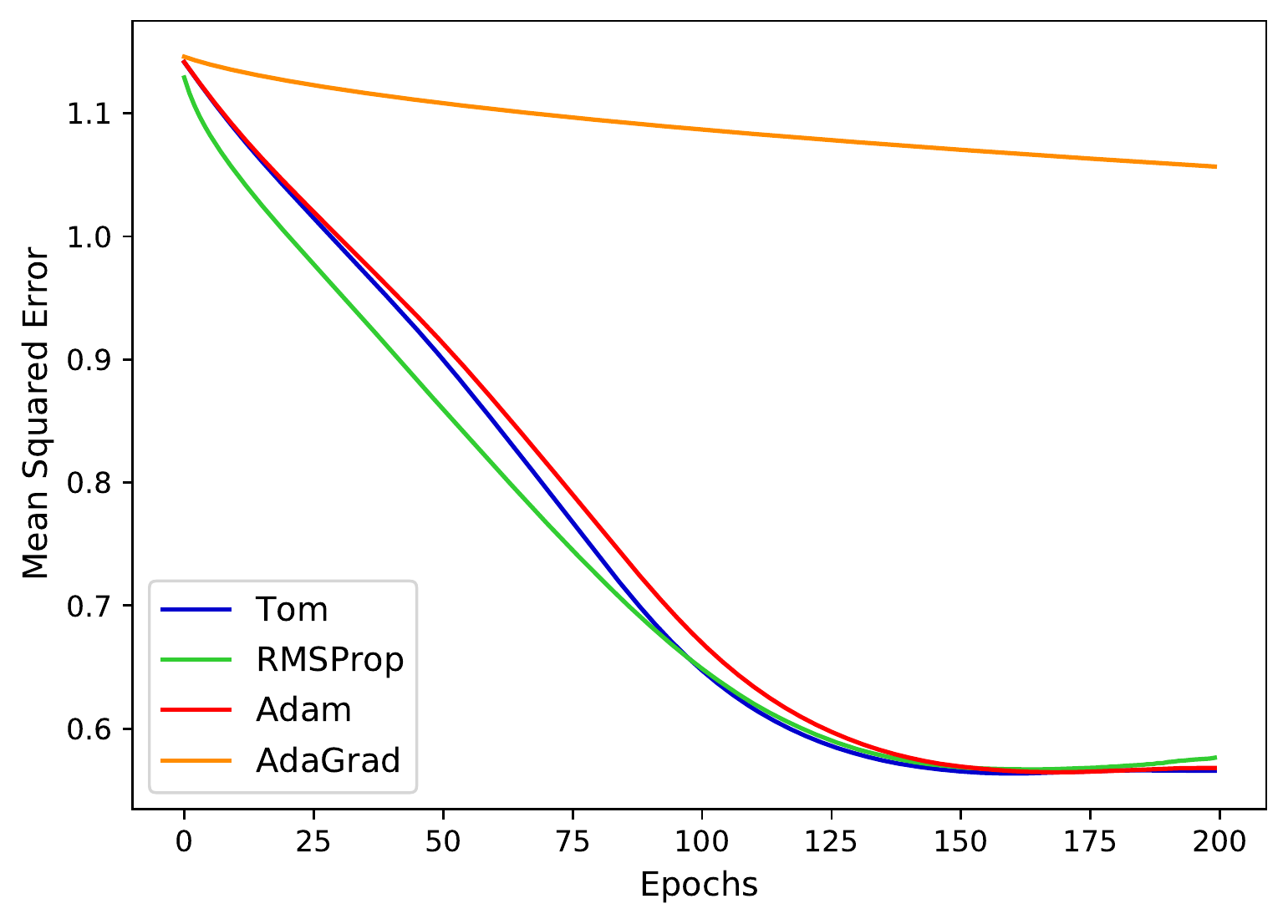}
         \caption{Mean squared error on the test set for Diabetes dataset}
     \end{subfigure}
        \caption{Mean squared error evolution for the training and test set on Diabetes dataset}
        \label{fig:dia_reg_fig}
\end{figure}

The test mean squared error for the dense neural network trained with batch gradient descent is reported in Table~\ref{tab:regression_tab}. RMSProp performs significantly better for dense neural networks which indicates that the trend in observed gradients is not significant for dense neurons to take advantage of. The train and test mean squared error evolution for the different datasets is shown in Figure [\ref{fig:boston_reg_fig}, \ref{fig:cali_reg_fig}, \ref{fig:dia_reg_fig}]

\begin{table}[!htbp]
\centering
\begin{tabular}{|c|c|cccc|}
\hline
\multicolumn{1}{|l|}{\multirow{3}{*}{Batch size}} & \multicolumn{1}{l|}{\multirow{3}{*}{Optimizer}} & \multicolumn{4}{c|}{CIFAR-10}                                     \\ \cline{3-6} 
\multicolumn{1}{|l|}{}                            & \multicolumn{1}{l|}{}                           & \multicolumn{4}{c|}{Epochs}                                       \\ \cline{3-6} 
\multicolumn{1}{|l|}{}                            & \multicolumn{1}{l|}{}                           & 10             & 30             & 75             & 100            \\ \hline
\multirow{4}{*}{512}                              & RMSProp                                      & 75.63$\pm$3.26          & 87.13$\pm$0.87          & 90.53$\pm$0.49          & 91.52$\pm$0.60          \\
                                                  & Adagrad                                         & 54.56$\pm$0.22          & 68.10$\pm$0.20          & 78.37$\pm$0.25          &  80.03$\pm$0.24          \\
                                                  & Adam                                            & 80.79$\pm$1.79 & 89.29$\pm$0.66 & 90.58$\pm$0.40 & 91.15$\pm$0.45 \\
                                                  & Tom                                             & \textbf{82.74$\pm$2.09} & \textbf{89.38$\pm$0.66} & \textbf{91.61$\pm$0.37} & \textbf{91.80$\pm$0.34} \\ \hline
\multirow{4}{*}{1024}                             & RMSProp                                        & 73.30$\pm$2.86          & 86.88$\pm$1.33          & 89.79$\pm$0.56          & 90.79$\pm$0.55          \\
                                                  & Adagrad                                         & 45.74$\pm$0.16          & 61.39$\pm$0.12          & 71.29$\pm$0.21          & 75.14$\pm$0.10          \\
                                                  & Adam                                            & 82.33$\pm$1.47          & 88.07$\pm$1.16          & \textbf{90.81$\pm$0.71}          & 91.0$\pm$0.21          \\
                                                  & Tom                                             & \textbf{84.17$\pm$1.13}          & \textbf{88.43$\pm$1.17}         & 90.52$\pm$0.75          & \textbf{91.46$\pm$0.40}          \\ \hline
\multirow{4}{*}{4096}                             & RMSProp                                         & 44.48$\pm$3.29          & 57.53$\pm$1.78          & 62.41$\pm$13.55          & 70.25$\pm$15.1          \\
                                                  & Adagrad                                         & 34.86$\pm$0.10         & 45.16$\pm$0.10          & 55.71$\pm$0.26          & 56.89$\pm$0.35          \\
                                                  & Adam                                            & 61.84$\pm$1.14          & 72.81$\pm$1.45          & 85.38$\pm$2.26          & 88.47$\pm$0.98          \\
                                                  & Tom                                             & \textbf{68.52$\pm$3.25}          & \textbf{76.44$\pm$4.73}          & \textbf{89.23$\pm$0.19}          & \textbf{90.72$\pm$0.38}      \\ \hline
\end{tabular}
\caption{Test Accuracy of Resnet-18 on CIFAR-10 dataset. The best accuracy among different optimizers are highlighted in bold.}
\label{tab:resnet18_cifar10_tab}
\end{table}
\begin{table}[!htbp]
\centering
\begin{tabular}{|c|c|cccc|}
\hline
\multicolumn{1}{|l|}{\multirow{3}{*}{Batch size}} & \multicolumn{1}{l|}{\multirow{3}{*}{Optimizer}} & \multicolumn{4}{c|}{CIFAR-10}                                     \\ \cline{3-6} 
\multicolumn{1}{|l|}{}                            & \multicolumn{1}{l|}{}                           & \multicolumn{4}{c|}{Epochs}                                       \\ \cline{3-6} 
\multicolumn{1}{|l|}{}                            & \multicolumn{1}{l|}{}                           & 10             & 30             & 75             & 100            \\ \hline
\multirow{4}{*}{512}                              & RMSProp                                         & 60.61$\pm$2.52         & 83.38$\pm$3.03          & 88.83$\pm$1.98          & 90.95$\pm$0.55          \\
                                                  & Adagrad                                         & 50.52$\pm$1.55          & 66.78$\pm$2.01          & 72.05$\pm$2.28          & 75.23$\pm$2.46          \\
                                                  & Adam                                            & 82.70$\pm$0.59 & \textbf{89.16$\pm$0.62} & 91.37$\pm$0.51 & \textbf{92.60$\pm$0.14} \\
                                                  & Tom                                             & \textbf{84.38$\pm$0.62} & 88.78$\pm$0.60 & \textbf{92.03$\pm$0.37} & 92.36$\pm$0.25 \\ \hline
\multirow{4}{*}{1024}                             & RMSProp                                          & 47.22$\pm$6.49          & 68.75$\pm$4.75          & 85.86$\pm$1.88          & 85.40$\pm$2.62          \\
                                                  & Adagrad                                         & 41.02$\pm$1.41          & 55.23$\pm$2.08          & 63.20$\pm$3.66          & 70.35$\pm$1.78          \\
                                                  & Adam                                             & 68.89$\pm$2.84          & 88.40$\pm$0.35          & 91.30$\pm$0.40          & 92.05$\pm$0.57          \\
                                                  & Tom                                              & \textbf{79.24$\pm$2.04}          & \textbf{89.21$\pm$0.55}          & \textbf{92.12$\pm$0.18}          & \textbf{92.30$\pm$0.56}          \\ \hline
\multirow{4}{*}{4096}                             & RMSProp                                         & 29.55$\pm$0.85          & 42.61$\pm$10.14          & 58.11$\pm$7.20          & 64.86$\pm$8.36          \\
                                                  & Adagrad                                         & 33.70$\pm$0.68          & 40.30$\pm$1.70          & 44.30$\pm$1.51          & 58.23$\pm$1.34          \\
                                                  & Adam                                            & 51.31$\pm$1.55         & 75.53$\pm$1.21          & 86.40$\pm$0.98          & 88.21$\pm$1.01          \\
                                                  & Tom                                             & \textbf{55.50$\pm$5.9}          & \textbf{80.09$\pm$1.03}          & \textbf{88.65$\pm$0.45}          & \textbf{89.784$\pm$1.16}      \\ \hline
\end{tabular}
\caption{Test Accuracy of Resnet-34 for CIFAR-10 dataset. The best accuracy among different optimizers are highlighted in bold.}
\label{tab:resnet34_cifar10_tab}
\end{table}
\begin{table}[!htbp]
\centering
\begin{tabular}{|c|c|cccc|}
\hline
\multicolumn{1}{|l|}{\multirow{3}{*}{Batch size}} & \multicolumn{1}{l|}{\multirow{3}{*}{Optimizer}} & \multicolumn{4}{c|}{CIFAR-100}                                     \\ \cline{3-6} 
\multicolumn{1}{|l|}{}                            & \multicolumn{1}{l|}{}                           & \multicolumn{4}{c|}{Epochs}                                       \\ \cline{3-6} 
\multicolumn{1}{|l|}{}                            & \multicolumn{1}{l|}{}                           & 10             & 30             & 75             & 100            \\ \hline
\multirow{4}{*}{512}                              & RMSProp                               & 46.72$\pm$1.19         & 60.71$\pm$1.91          & 65.41$\pm$0.74          & 66.59$\pm$0.63          \\
                                                  & Adagrad                                         & 15.32$\pm$0.47          & 26.58$\pm$0.35          & 41.39$\pm$0.48          & 45.66$\pm$0.40          \\
                                                  & Adam                                            & 52.97$\pm$1.89 & 65.12$\pm$1.22 & 67.20$\pm$0.27 & 67.91$\pm$0.32 \\
                                                  & Tom                                              & \textbf{54.08$\pm$1.56} & \textbf{65.48$\pm$0.27} & \textbf{67.75$\pm$0.47} & \textbf{68.50$\pm$0.86} \\ \hline
\multirow{4}{*}{1024}                             & RMSProp                                         & 38.95$\pm$2.35          & 58.98$\pm$1.72          & 63.89$\pm$.49          & 65.03$\pm$0.83          \\
                                                  & Adagrad                                         & 11.68$\pm$0.44          & 18.80$\pm$0.25          & 31.32$\pm$0.30          & 35.88$\pm$0.58          \\
                                                  & Adam                                             & 50.32$\pm$2.14          & \textbf{64.49$\pm$0.52}          & 65.79$\pm$0.96          & 66.11$\pm$0.47          \\
                                                  & Tom                                              & \textbf{51.77$\pm$1.28}          & 63.74$\pm$1.19          & \textbf{66.63$\pm$0.96}          & \textbf{67.20$\pm$0.73}          \\ \hline
\multirow{4}{*}{4096}                             & RMSProp                                         & 13.78$\pm$1.02          & 30.57$\pm$1.84          & 36.39$\pm$3.37          & 40.52$\pm$4.41          \\
                                                  & Adagrad                                         & 6.14$\pm$0.43          & 10.81$\pm$0.29          & 15.75$\pm$0.42          & 17.92$\pm$0.38          \\
                                                  & Adam                                             & 29.08$\pm$1.03         & 49.40$\pm$5.16          & 64.80$\pm$0.78  &        66.18$\pm$1.28          \\
                                                  & Tom                                             & \textbf{34.38$\pm$2.04}          & \textbf{51.30$\pm$3.69}          & \textbf{65.80$\pm$0.75}          & \textbf{67.59$\pm$1.02}      \\ \hline
\end{tabular}
\caption{Test Accuracy of Resnet-18 for CIFAR-100 dataset. The best accuracy among different optimizers are highlighted in bold.}
\label{tab:resnet18_cifar100_tab}
\end{table}
\begin{table}[!htbp]
\centering
\begin{tabular}{|c|c|cccc|}
\hline
\multicolumn{1}{|l|}{\multirow{3}{*}{Batch size}} & \multicolumn{1}{l|}{\multirow{3}{*}{Optimizer}} & \multicolumn{4}{c|}{CIFAR-100}                                     \\ \cline{3-6} 
\multicolumn{1}{|l|}{}                            & \multicolumn{1}{l|}{}                           & \multicolumn{4}{c|}{Epochs}                                       \\ \cline{3-6} 
\multicolumn{1}{|l|}{}                            & \multicolumn{1}{l|}{}                           & 10             & 30             & 75             & 100            \\ \hline
\multirow{4}{*}{512}                              & RMSProp                                         & 29.91$\pm$4.12          & 48.33$\pm$3.31          & 61.85$\pm$1.24          & 64.14$\pm$1.39          \\
                                                  & Adagrad                                         & 15.24$\pm$1.14          & 26.53$\pm$0.83          & 40.90$\pm$0.99          &  43.68$\pm$1.42          \\
                                                  & Adam                                            & 51.56$\pm$1.09 & 65.33$\pm$0.56 & 68.23$\pm$0.51 & 68.48$\pm$0.41 \\
                                                  & Tom                                              & \textbf{51.63$\pm$1.55} & \textbf{66.11$\pm$0.36} & \textbf{68.95$\pm$0.97} & \textbf{69.60$\pm$0.69} \\ \hline
\multirow{4}{*}{1024}                             & RMSProp                                         & 19.10$\pm$3.29          & 36.90$\pm$2.98          & 55.42$\pm$3.48          & 60.43$\pm$1.58          \\
                                                  & Adagrad                                        & 11.41$\pm$0.42          & 17.44$\pm$1.69          & 27.47$\pm$1.36          & 31.98$\pm$2.39          \\
                                                  & Adam                                            & 47.00$\pm$2.29          & 63.68$\pm$0.60          & 68.62$\pm$1.17          & 68.56$\pm$0.95          \\
                                                  & Tom                         & \textbf{48.29$\pm$1.26}          & \textbf{64.88$\pm$1.24}         & \textbf{69.79$\pm$0.42}          & \textbf{69.71$\pm$0.57}          \\ \hline
\multirow{4}{*}{4096}                             & RMSProp                                         & 6.62$\pm$1.47          & 15.74$\pm$2.05          & 28.39$\pm$4.09          & 28.86$\pm$5.36          \\
                                                  & Adagrad                                        & 4.16$\pm$0.61         & 10.42$\pm$0.49          & 15.80$\pm$0.68          & 16.96$\pm$1.63          \\
                                                  & Adam                     & 16.60$\pm$1.33          & 40.07$\pm$3.60          & 60.69$\pm$0.74          & 64.62$\pm$0.83          \\
                                                  & Tom                               & \textbf{18.38$\pm$1.77}          & \textbf{45.83$\pm$1.00}          & \textbf{62.77$\pm$0.61}          & \textbf{66.48$\pm$1.52}      \\ \hline
\end{tabular}
\caption{Test Accuracy of Resnet-34 for CIFAR-100 dataset. The best accuracy among different optimizers are highlighted in bold.}
\label{tab:resnet34_cifar100_tab}
\end{table}
\begin{table}[!htbp]
\centering
\begin{tabular}{|c|c|cccc|}
\hline
\multicolumn{1}{|l|}{\multirow{3}{*}{Batch size}} & \multicolumn{1}{l|}{\multirow{3}{*}{Optimizer}} & \multicolumn{4}{c|}{CINIC-10}                                     \\ \cline{3-6} 
\multicolumn{1}{|l|}{}                            & \multicolumn{1}{l|}{}                           & \multicolumn{4}{c|}{Epochs}                                       \\ \cline{3-6} 
\multicolumn{1}{|l|}{}                            & \multicolumn{1}{l|}{}                           & 10             & 30             & 75             & 100            \\ \hline
\multirow{4}{*}{512}                              & RMSProp                                        & 69.32$\pm$2.27          & 78.46$\pm$1.36          & 81.27$\pm$0.33          & 81.49$\pm$0.43          \\
                                                  & Adagrad                                         & 50.47$\pm$0.50          & 60.99$\pm$0.18          & 68.10$\pm$0.42          &  69.65$\pm$0.45          \\
                                                  & Adam                          & 73.22$\pm$1.73 & 79.97$\pm$0.63 & \textbf{81.95$\pm$0.22} & 82.06$\pm$0.38 \\
                                                  & Tom                                             & \textbf{74.59$\pm$0.87} & \textbf{80.77$\pm$0.34} & 81.85$\pm$0.48 & \textbf{82.53$\pm$0.24} \\ \hline
\multirow{4}{*}{1024}                             & RMSProp                                      & 64.76$\pm$2.63          & 77.69$\pm$0.58          & 79.88$\pm$0.40          & 80.71$\pm$0.34          \\
                                                  & Adagrad                                          & 44.75$\pm$0.32          & 55.47$\pm$0.21         & 63.76$\pm$0.38          & 66.31$\pm$0.50          \\
                                                  & Adam                     & 70.53$\pm$1.19          & \textbf{79.08$\pm$0.88}          & 81.27$\pm$0.23         & \textbf{81.92$\pm$0.23}          \\
                                                  & Tom                                     & \textbf{72.17$\pm$0.97}          & 78.73$\pm$0.83         & \textbf{81.63$\pm$0.40}          & 81.86$\pm$0.27          \\ \hline
\multirow{4}{*}{4096}                             & RMSProp                                      & 40.48$\pm$5.89          & 56.31$\pm$6.01          & 67.36$\pm$3.69          & 66.79$\pm$3.52          \\
                                                  & Adagrad                                       & 32.97$\pm$0.63         & 41.61$\pm$0.72          & 50.84$\pm$0.83          & 51.79$\pm$1.53          \\
                                                  & Adam                      & 55.85$\pm$2.59          & 73.27$\pm$0.68          & 78.78$\pm$0.62          & 80.12$\pm$0.43          \\
                                                  & Tom                                             & \textbf{62.53$\pm$2.05}          & \textbf{75.31$\pm$1.57}          & \textbf{79.77$\pm$0.35}          & \textbf{80.93$\pm$0.51}      \\ \hline
\end{tabular}
\caption{Test Accuracy of Resnet-18 for CINIC-10 dataset. The best accuracy among different optimizers are highlighted in bold.}
\label{tab:resnet18_cinic10_tab}
\end{table}
\begin{table}[!htbp]
\centering
\begin{tabular}{|c|c|cccc|}
\hline
\multicolumn{1}{|l|}{\multirow{3}{*}{Batch size}} & \multicolumn{1}{l|}{\multirow{3}{*}{Optimizer}} & \multicolumn{4}{c|}{CINIC-10}                                     \\ \cline{3-6} 
\multicolumn{1}{|l|}{}                            & \multicolumn{1}{l|}{}                           & \multicolumn{4}{c|}{Epochs}                                       \\ \cline{3-6} 
\multicolumn{1}{|l|}{}                            & \multicolumn{1}{l|}{}                           & 10             & 30             & 75             & 100            \\ \hline
\multirow{4}{*}{512}                              & RMSProp                                      & 61.27$\pm$4.59          & 76.64$\pm$1.88          & 80.36$\pm$0.48          & 81.16$\pm$0.36          \\
                                                  & Adagrad                                      & 49.03$\pm$1.36          & 60.74$\pm$1.14          & 67.34$\pm$0.83          &  67.04$\pm$1.31          \\
                                                  & Adam                                            & 71.54$\pm$2.00 & 79.70$\pm$0.75 & 82.47$\pm$0.36 & 82.97$\pm$0.17 \\
                                                  & Tom                                             & \textbf{72.46$\pm$1.36} & \textbf{80.75$\pm$0.67} & \textbf{82.63$\pm$0.49} & \textbf{83.13$\pm$0.22} \\ \hline
\multirow{4}{*}{1024}                             & RMSProp                                       & 49.10$\pm$3.92          & 70.28$\pm$3.74          & 78.35$\pm$1.25          & 79.78$\pm$1.00          \\
                                                  & Adagrad                                         & 42.32$\pm$1.87          & 53.43$\pm$2.70          & 61.85$\pm$1.41          & 64.83$\pm$1.33          \\
                                                  & Adam                                           & 69.41$\pm$1.40          & 79.38$\pm$0.95          & 82.22$\pm$0.26          & 82.43$\pm$0.17          \\
                                                  & Tom                                              & \textbf{70.37$\pm$1.70}          & \textbf{79.69$\pm$0.84}         & \textbf{82.56$\pm$0.25}          & \textbf{83.00$\pm$0.15}          \\ \hline
\multirow{4}{*}{4096}                             & RMSProp                                         & 29.84$\pm$4.23          & 44.86$\pm$3.40          & 60.57$\pm$8.01          & 66.39$\pm$7.01          \\
                                                  & Adagrad                     & 32.56$\pm$0.61         & 40.45$\pm$2.01          & 47.43$\pm$1.98          & 51.56$\pm$1.46          \\
                                                  & Adam                           & 51.64$\pm$2.83          & 70.75$\pm$2.26          & 78.04$\pm$1.42          & 80.54$\pm$0.46          \\
                                                  & Tom                                & \textbf{53.29$\pm$5.19}          & \textbf{71.11$\pm$1.46}          & \textbf{79.47$\pm$1.49}          & \textbf{80.88$\pm$0.30}      \\ \hline
\end{tabular}
\caption{Test Accuracy of Resnet-34 for CINIC-10 dataset. The best accuracy among different optimizers are highlighted in bold.}
\label{tab:resnet34_cinic10_tab}
\end{table}


\begin{table}[!htbp]
\centering
\begin{tabular}{|c|c|cccc|}
\hline
\multirow{2}{*}{Dataset}            & \multirow{2}{*}{Optimizer} & \multicolumn{4}{c|}{Epochs}    \\ \cline{3-6} 
                                    &                            & 50    & 100   & 150   & 200    \\ \hline
\multirow{4}{*}{Boston Housing}     & RMSProp                    & \textbf{0.7281$\pm$0.06} & \textbf{0.4621$\pm$0.07} & \textbf{0.3159$\pm$0.06} & \textbf{0.2311$\pm$0.04}  \\
                                    & Adagrad                    & 1.0970$\pm$0.07 & 1.0699$\pm$0.08 & 1.0481$\pm$0.07 & 1.0285$\pm$0.06 \\
                                    & Adam                       & 0.8138$\pm$0.07 & 0.5041$\pm$0.06 & 0.3374$\pm$0.05 & 0.2524$\pm$0.04  \\
                                    & Tom                        & 0.7937$\pm$0.06 & 0.4767$\pm$0.06 & 0.3174$\pm$0.05 & 0.2418$\pm$0.04  \\ \hline
\multirow{4}{*}{California Housing} & RMSProp                    & \textbf{0.8773$\pm$0.02} & \textbf{0.6841$\pm$0.05} & 0.5114$\pm$0.04 & 0.4107$\pm$0.02  \\
                                    & Adagrad                    & 1.0327$\pm$0.02 & 1.0241$\pm$0.01 & 1.0170$\pm$0.01 & 1.0109$\pm$0.02  \\
                                    & Adam                       & 0.9166$\pm$0.02 & 0.7009$\pm$0.05 & 0.4971$\pm$0.04 & 0.4007$\pm$0.02  \\
                                    & Tom                        & 0.9075$\pm$0.02 & 0.6711$\pm$0.05 & \textbf{0.4738$\pm$0.04} & \textbf{0.3864$\pm$0.02}  \\ \hline
\multirow{4}{*}{Diabetes dataset}      & RMSProp                    & \textbf{0.8642$\pm$0.06} & 0.6519$\pm$0.04 & 0.5688$\pm$0.02 & 0.5763$\pm$0.01  \\
                                    & Adagrad                    & 1.1088$\pm$0.05 & 1.10871$\pm$0.05 & 1.0707$\pm$0.05 & 1.0568$\pm$0.05  \\
                                    & Adam                       & 0.9175$\pm$0.06 & 0.6736$\pm$0.05 & 0.5696$\pm$0.02 & 0.5680$\pm$0.01  \\
                                    & Tom                        & 0.9047$\pm$0.06 & \textbf{0.6509$\pm$0.04} & \textbf{0.5655$\pm$0.01} & \textbf{0.5658$\pm$0.02}  \\ \hline
\end{tabular}
\caption{Test mean squared error for regression datasets. The best mean squared error among different optimizers are highlighted in bold.}
\label{tab:regression_tab}
\end{table}

\newpage
\section{Conclusion}
We have proposed \emph{Tom}, a novel variant of Adam optimizer that accounts for the trend that is observed for the gradients during the process of optimization. \emph{Tom} maintains a running history of the rate of change of gradients between two successive time steps, and helps in boosting convergence. With the extensive evaluation for the task of classification and regression, \emph{Tom} performs consistently better than Adam due to the addition of the trend component. Furthermore \emph{Tom} can be inculcated to any Adam-like optimizer to include the trend component and hence boost convergence. We validate the purpose of \emph{Tom} with intuitive argumentation and its applications on real-world datasets with minimal tuning of hyperparameters.   
\bibliographystyle{IEEEtran}
\bibliography{new_references}

\end{document}